\documentclass[10pt,journal,compsoc]{IEEEtran}
\usepackage[table,dvipsnames]{xcolor}
\usepackage{float}
\usepackage{amssymb}
\PassOptionsToPackage{hyphens}{url}\usepackage{hyperref}
\usepackage{tabularx}
\usepackage{multirow}
\usepackage{amsmath}
\newtheorem{definition}{Definition}
\usepackage{makecell}
\usepackage{bm}
\usepackage{acro}
\usepackage{hyperref}
\usepackage{threeparttable}
\usepackage[ruled,linesnumbered, vlined]{algorithm2e}
\usepackage{adjustbox}
\usepackage{booktabs}
\usepackage{wrapfig}
\usepackage{pifont}%
\DeclareAcronym{aucroc}{
    short = AUC-ROC ,
    long = area under the Receiver Operating Characteristic curve
}
\DeclareAcronym{aucpr}{
    short = AUC-PR ,
    long = area under the precision-recall curve
}
\DeclareAcronym{amr}{
    short = AMR ,
    long = adjusted mean rank
}
\DeclareAcronym{amrr}{
    short = AMRR ,
    long = adjusted mean reciprocal rank
}
\DeclareAcronym{bcel}{
    short = BCEL ,
    long = binary cross entropy loss
}
\DeclareAcronym{bns}{
    short = BNS ,
    long = Bernoulli negative sampling
}
\DeclareAcronym{cel}{
    short = CEL,
    long = cross entropy loss
}
\DeclareAcronym{cp}{
    short = CP ,
    long = canonical polyadic ,
}
\DeclareAcronym{cnn}{
    short = CNN,
    long = convolutional neural network,
}
\DeclareAcronym{cwa}{
	short = CWA ,
	long  = closed world assumption ,
}
\DeclareAcronym{gcn}{
	short = GCN ,
	long  = graph convolutional network ,
}
\DeclareAcronym{hpo}{
	short = HPO ,
	long  = hyper-parameter optimization ,
}
\DeclareAcronym{kg}{
	short = KG ,
	long  = knowledge graph ,
}
\DeclareAcronym{kgem}{
	short = KGEM ,
	long  = knowledge graph embedding model ,
}
\DeclareAcronym{kl}{
	short = KL ,
	long  = Kullback-Leibler ,
}
\DeclareAcronym{lcwa}{
	short = LCWA ,
	long  = local closed world assumption ,
}
\DeclareAcronym{slcwa}{
	short = sLCWA ,
	long  = stochastic local closed world assumption ,
}
\DeclareAcronym{mr}{
    short = MR ,
    long = mean rank
}
\DeclareAcronym{mrl}{
    short = MRL ,
    long = margin ranking loss
}
\DeclareAcronym{fmr}{
    short = FMR ,
    long = filtered mean rank
}
\DeclareAcronym{mrr}{
    short = MRR ,
    long = mean reciprocal rank
}
\DeclareAcronym{nssal}{
    short = NSSAL ,
    long = self-adversarial negative sampling loss
}
\DeclareAcronym{ntn}{
    short = NTN ,
    long = Neural Tensor Network ,
}
\DeclareAcronym{owa}{
	short = OWA ,
	long  = open world assumption ,
}
\DeclareAcronym{rgcn}{
    short = R-GCN ,
    long = Relational Graph Convolutional Network
}
\DeclareAcronym{se}{
    short = SE ,
    long = Structured Embedding
}
\DeclareAcronym{smbo}{
	short = SMBO ,
	long  = Sequential Model-Based Global Optimization
}
\DeclareAcronym{smm}{
	short = SMM ,
	long  = semantic matching model ,
}
\DeclareAcronym{spl}{
	short = SPL ,
	long  = Softplus loss ,
}
\DeclareAcronym{tdm}{
	short = TDM ,
	long  = translational distance model ,
}
\DeclareAcronym{tpe}{
	short = TPE ,
	long  = tree-structured parzen estimator ,
}
\DeclareAcronym{um}{
	short = UM ,
	long  = Unstructured Model ,
	cite = bordes2014semantic ,
}
\DeclareAcronym{umls}{
	short = UMLS ,
	long  = Unified Medical Language System ,
	cite = mccray2003upper
}
\DeclareAcronym{uns}{
	short = UNS ,
	long  = uniform negative sampling ,
}
\DeclareAcronym{yago}{
	short = YAGO ,
	long  = Yet Another Great Ontology ,
	cite = rebele2016yago
}

\begin{document}

\title{Bringing Light Into the Dark:\\A Large-scale Evaluation of Knowledge Graph Embedding Models under a Unified Framework}

\author{Mehdi~Ali,
        Max~Berrendorf$^\dagger$,
        Charles~Tapley~Hoyt$^\dagger$,
        Laurent~Vermue$^\dagger$,
        Mikhail~Galkin,\\
        Sahand~Sharifzadeh,
        Asja~Fischer,
        Volker~Tresp,
        and~Jens~Lehmann

\thanks{\textdagger Equal contribution.}
\thanks{Mehdi Ali is affiliated with Smart Data Analytics (University of Bonn), Germany, \& Fraunhofer IAIS, Sankt Augustin and Dresden, Germany.}%
\thanks{Max Berrendorf is affiliated with Ludwig-Maximilians-Universit\"at M\"unchen, Munich, Germany.}
\thanks{Charles Tapley Hoyt is affiliated with Laboratory of Systems Pharmacology, Harvard Medical School, Boston, USA.}
\thanks{Laurent Vermue is affiliated with the Technical University of Denmark, Kongens Lyngby, Denmark.}
\thanks{Mikhail Galkin is affiliated with Mila \& McGill University, Montreal, Canada}
\thanks{Sahand Sharifzadeh is affiliated with Ludwig-Maximilians-Universit\"at M\"unchen, Munich, Germany.}
\thanks{Asja Fischer is affiliated with the Ruhr University Bochum, Germany.}
\thanks{Volker Tresp is affiliated with Ludwig-Maximilians-Universit\"at M\"unchen \& Siemens AG, Munich, Germany.}
\thanks{Jens Lehmann is affiliated with Smart Data Analytics (University of Bonn), Bonn, Germany, \& Fraunhofer IAIS, Sankt Augustin and Dresden Germany.}
}
\ifCLASSOPTIONpeerreview
\markboth{KGEM Benchmarking}
{KGEM Benchmarking}
\else
\markboth{Ali \MakeLowercase{\textit{et al.}}}
{Ali \MakeLowercase{\textit{et al.}}: KGEM Benchmarking}
\fi

\maketitle

\begin{abstract}

The heterogeneity in recently published knowledge graph embedding models' implementations, training, and evaluation has made fair and thorough comparisons difficult.
To assess the reproducibility of previously published results, we re-implemented and evaluated 21 models in the PyKEEN software package.
In this paper, we outline which results could be reproduced with their reported hyper-parameters, which could only be reproduced with alternate hyper-parameters, and which could not be reproduced at all,  as well as provide insight as to why this might be the case.

We then performed a large-scale benchmarking on four datasets with several thousands of experiments and 24,804 GPU hours of computation time.
We present insights gained as to best practices, best configurations for each model, and where improvements could be made over previously published best configurations.
Our results highlight that the combination of model architecture, training approach, loss function, and the explicit modeling of inverse relations is crucial for a model's performance and is not only determined by its architecture.
We provide evidence that several architectures can obtain results competitive to the state of the art when configured carefully.
We have made all code, experimental configurations, results, and analyses available at \url{https://github.com/pykeen/pykeen} and \url{https://github.com/pykeen/benchmarking}.

\end{abstract}


\IEEEpeerreviewmaketitle

\section{Introduction}

\IEEEPARstart{A}{s} the usage of \acp{kg} becomes more widespread, their inherent incompleteness can pose a liability for typical downstream tasks that they support, e.g.,\ question answering, dialogue systems, and recommendation systems~\cite{wang2017knowledge}.
\Acp{kgem} present an avenue for predicting missing links.
However, the following two major challenges remain in their application.

First, the reproduction of  previously reported results turned out to be a major challenge --- there are even examples of different results reported for the same combinations of \acp{kgem} and datasets~\cite{akrami2018re}.
In some cases, the lack of availability of source code for \acp{kgem} or the usage of different frameworks and programming languages inevitably introduces variability.
In other cases, the lack of a precise specification of hyper-parameters introduces variability.

Second, the verification of the novelty of previously reported results remains difficult.
It is often difficult to attribute the incremental improvements in performance reported with each new state of the art model to the model's architecture itself or instead to the training approach, hyper-parameter values, or specific prepossessing steps, e.g.,\ the explicit modeling of inverse relations.
It has been shown that baseline models can achieve competitive performance to more sophisticated ones when optimized appropriately~\cite{kadlec2017knowledge,akrami2018re}.
Additionally, the variety of implementations and interpretations of common evaluation metrics for link prediction makes a fair comparison to previous results difficult~\cite{sun2019re}.

This paper makes two major contributions towards addressing these challenges:
\begin{enumerate}
    \item We performed a reproducibility study in which we tried to replicate reported experimental results in the original papers (when sufficient information was provided).

    \item We performed an extensive benchmark study on 21 \acp{kgem} over four benchmark datasets in which we evaluated the models based on different hyper-parameter values, training approaches (i.e.\ training under the \textit{\acl{lcwa}} and \textit{\acl{slcwa}}), loss functions, optimizers, and the explicit modeling of inverse relations. 
\end{enumerate}

Previous studies have already investigated important aspects for a subset of models: Kadlec \textit{et al.}~\cite{kadlec2017knowledge} showed that a fine-tuned baseline (DistMult~\cite{Yang2014}) can outperform more sophisticated models on FB15K.
Akrami \textit{et al.}~\cite{akrami2018re,akrami2020realistic} examined the effect of removing faulty triples from KGs on the model's performance.
Mohamed \textit{et al.}~\cite{mohamed2019loss} studied the influence of loss functions on the models' performances for a set of \acp{kgem}.
Concurrent to the work on this paper, Rufinelli \textit{et al.}~\cite{ruffinelli2020you} performed a benchmarking study in which they investigated ﬁve knowledge graph embedding models.
After describing their benchmarking~\cite{ruffinelli2020you}, they called for a larger study that extends the search space and incorporates more sophisticated models.
Our study answers this call and realizes a fair benchmarking by completely re-implementing \acp{kgem}, training pipelines, loss functions, and evaluation metrics in a unified, open-source framework. 
Inspired by their findings, we have also included the \ac{cel} function, which has been previously used by Kadlec \textit{et al.}~\cite{kadlec2017knowledge}.
Our benchmarking can be considered as a superset of many previous benchmarkings --- to the best of our knowledge, there exists no study of comparable breadth or depth.
A further interesting study with a different focus is the work of Rossi \textit{et al.}~\cite{rossi2020knowledge} in which they investigated the effect of the structural properties of \acp{kg} on models' performances, instead of focusing on the combinations of different model architectures, training approaches, and loss functions.

This article is structured as follows:
in Section~\ref{sec:notation}, we introduce our notation of \ac{kg} and the link prediction task and introduce an exemplary \ac{kg} to which we refer in examples throughout this paper.
In Section~\ref{sec:kges}, we present our definition of a \ac{kgem} and review the \acp{kgem} that we investigated in our studies.
In Section~\ref{sec:evaluation_of_models}, we describe and discuss established evaluation metrics as well as a recently proposed one~\cite{berrendorf2020interpretable}.
In Section~\ref{sec:benchmark_datasets}, we introduce the benchmark datasets on which we conducted our experiments.
In Section~\ref{reproducibility_study} and Section~\ref{benchmarking_study}, we present our respective reproducibility and benchmarking studies.
In Section~\ref{relational_pattern_analysis}, we investigate how well the investigated \acp{kgem} can model symmetry, anti-symmetry, and composition patterns.
Finally, we provide a discussion and an outlook for our future work in Section~\ref{sec:discussion_future_work}.

\section{Knowledge Graphs}\label{sec:notation}

\begin{figure}[!t]
\centering
\includegraphics[width=3.5in]{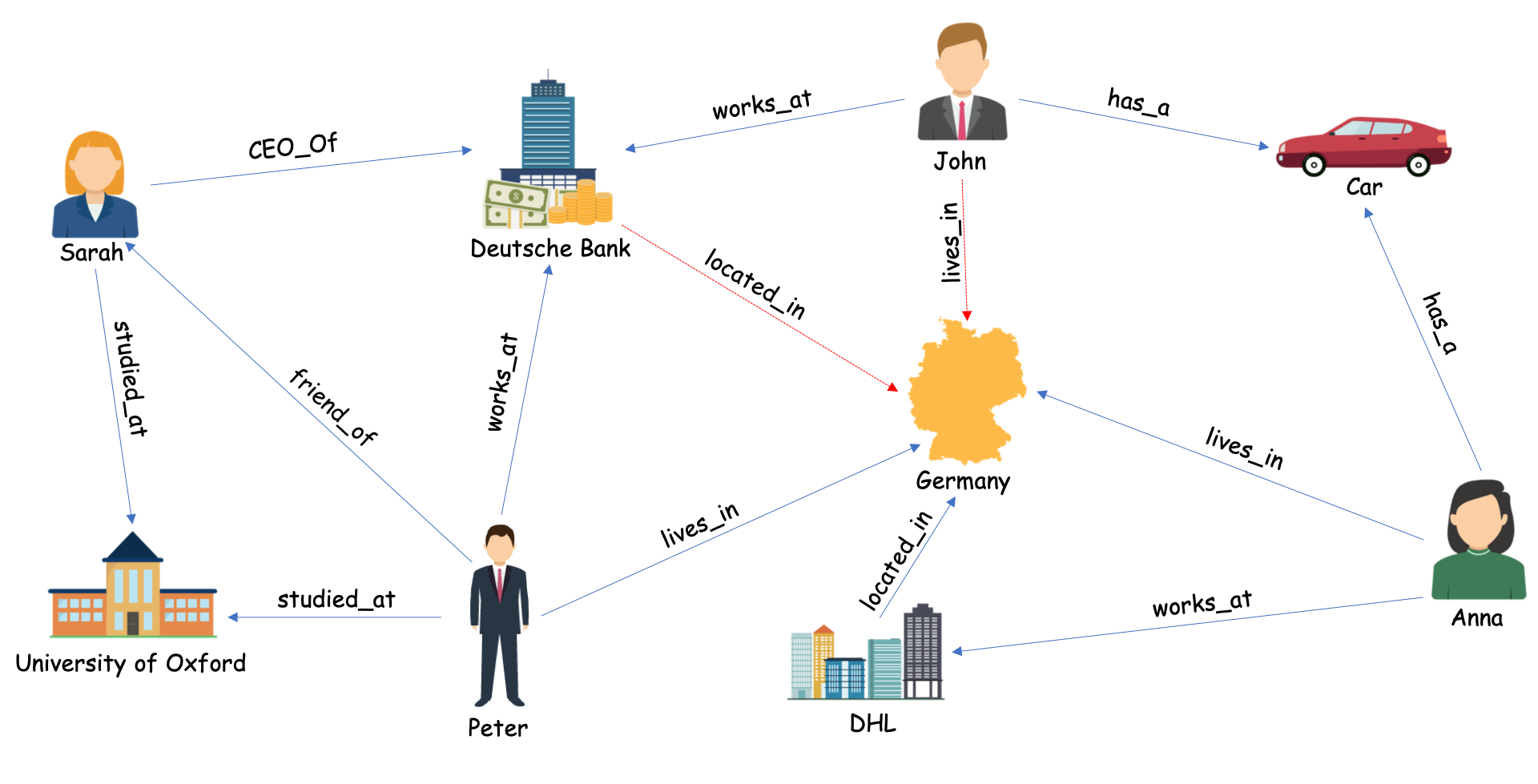}
\caption{Exemplary \ac{kg}: nodes represent entities and edges their respective relations.}
\label{fig:exemplary_kg}
\end{figure}

For a given set of entities $\mathcal{E}$ and set of relations $\mathcal{R}$, we consider a \acl{kg} \begin{math}\mathcal{K} \subseteq \mathbb{K} = \mathcal{E} \times \mathcal{R} \times \mathcal{E}\end{math} as a directed, multi-relational graph that comprises triples \begin{math}(h,r,t) \in \mathcal{K}\end{math} in which
$h,t \in \mathcal{E}$ represent a triples' respective head and tail entities and \begin{math}r \in \mathcal{R}\end{math} represents its relationship.
Figure~\ref{fig:exemplary_kg} depicts an exemplary \ac{kg}. %
The direction of a relationship indicates the roles of the entities, i.e.,\ head or tail entity.
For instance, in the triple \textit{(Sarah, CEO\_Of, Deutsche\_Bank)}, \textit{Sarah} is the head and \textit{Deutsche\_Bank} is the tail entity.
\acp{kg} usually contain only true triples corresponding to available knowledge.

In contrast to triples in a \ac{kg}, there are different philosophies, or \textit{assumptions}, for the consideration of triples \textit{not} contained in a \ac{kg}~\cite{nickel2015review,kotnis2017analysis}.
Under the \ac{cwa}, all triples that are not part of a \ac{kg} are considered as false.
Based on the example in Figure~\ref{fig:exemplary_kg}, the triple \textit{(Sarah, lives\_in, Germany)} is a false fact under the \ac{cwa} since it is not part of the \ac{kg}.
Under the \ac{owa}, it is considered unknown as to whether triples that are not part of the \ac{kg} are true or false.
The construction of \acp{kg} under the principles of the semantic web (and RDF) rely on the \ac{owa} as well as most of the relevant works to this paper~\cite{galarraga2013amie,nickel2015review}.

Because \acp{kg} are usually incomplete and noisy, several approaches have been developed to predict new links.
In particular, the task of link prediction is defined as predicting the tail/head entities for $(h,r)$/$(r,t)$ pairs.
For instance, given queries of the form \textit{(Sarah, studied\_at, ?)} or \textit{(?, CEO\_of, Deutsche Bank)}, the task is the correctly detect the entities that answer the query, i.e.\ \textit{(Sarah, studied\_at, \textbf{University of Oxford})} and \textit{(\textbf{Sarah}, CEO\_of, Deutsche Bank)}.
While classical approaches have relied on domain-specific rules to derive missing links, they usually require a large number of user-defined rules in order to generalize~\cite{nickel2015review}.
Alternatively, machine learning approaches learn to predict new links based on the set of existing ones.
It has been shown that especially relational-machine learning methods are successful in predicting missing links and identifying incorrect ones, and recently \aclp{kgem} have gained significant attention~\cite{nickel2015review}.

\section{\Aclp{kgem}}\label{sec:kges}

\begin{figure}[!t]
\centering
\includegraphics[width=2.5in]{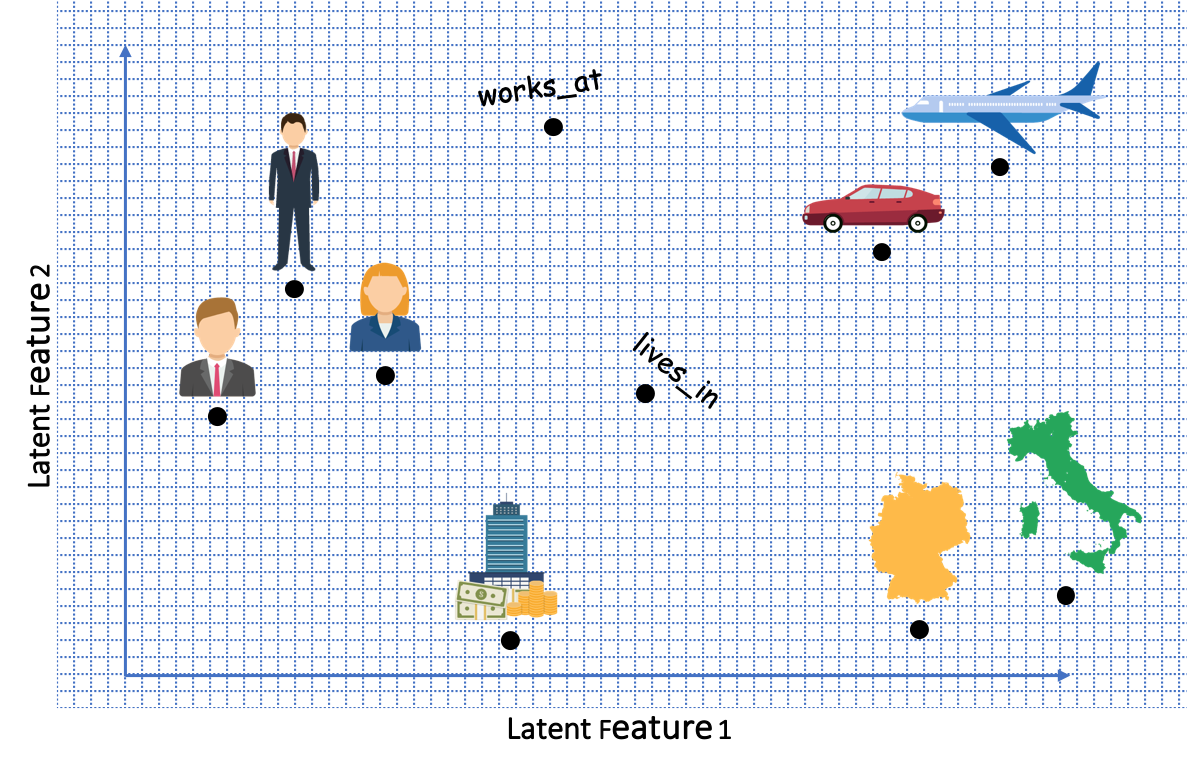}
\caption{An example embedding of the entities and relations from the knowledge graph portrayed by Figure~\ref{fig:exemplary_kges}.}
\label{fig:exemplary_kges}
\end{figure}

\Acfp{kgem} learn latent vector representations of the entities \begin{math}e \in \mathcal{E}\end{math} and relations \begin{math}r \in \mathcal{R}\end{math} in a \ac{kg} that best preserve its structural properties~\cite{wang2017knowledge,nickel2015review, ji2021survey}. 
Besides for link prediction, they have been used for tasks such as entity disambiguation, and clustering as well as for downstream tasks such as question answering, recommendation systems, and relation extraction~\cite{wang2017knowledge}.
Figure~\ref{fig:exemplary_kges} shows an embedding of the entities and relations in \begin{math}\mathbb{R}^2\end{math} from the \ac{kg} from Figure~\ref{fig:exemplary_kg}.

Here, we define a \ac{kgem} as four components: an \textit{interaction model}, a \textit{training approach}, a \textit{loss function}, and its usage of \emph{explicit inverse relations}.
This abstraction enables investigation of the effect of each component individually and in combination on each \acp{kgem}' performance.
Each are described in detail in their following respective subsections~\ref{interaction_model}, ~\ref{training_assumption}, ~\ref{loss_functions}, and \ref{sec:inverse_relations}.
We focus on \textit{shallow} embedding approaches~\cite{DBLP:journals/debu/HamiltonYL17} in this work, i.e., matrix lookups represent the entity and relation encoders.
Recently, several graph neural network (GNN)-based approaches for learning representations of \acp{kg}  have been developed. GNNs encode entities and relations by neighbor aggregation.
We refer interested readers to ~\cite{ji2021survey, DBLP:journals/debu/HamiltonYL17}.
Furthermore, learning representation for temporal \acp{kg} has gained increased interest.
Because learning representation for temporal \acp{kg} is a distinct line of research with its own benchmarking datasets, we do not discuss temporal  \acp{kgem} in this work. Instead, we refer interested readers to~\cite{DBLP:journals/jmlr/KazemiGJKSFP20}.

In this paper, we use a boldface lower-case letter \begin{math}\textbf{x}\end{math} to denote a vector, \begin{math}\|\textbf{x}\|_p\end{math} to represent its \begin{math}l_p\end{math} norm, a boldface upper-case letter \begin{math}\textbf{X}\end{math} to denote a matrix, and a fraktur-font upper-case letter \begin{math}\mathfrak{X}\end{math} to represent a three-mode tensor.
Furthermore, we use \begin{math}\odot \end{math} to denote the Hadamard product \begin{math}\odot: \mathbb{R}^d \times \mathbb{R}^d \rightarrow \mathbb{R}^d\end{math}:

\begin{equation}
    [\textbf{a} \odot \textbf{b}]_i = \textbf{a}_i \cdot \textbf{b}_i
\end{equation}

Finally, we use \begin{math}\overline{x}\end{math} to denote the conjugate of a complex number \begin{math}x \in \mathbb{C} \end{math}.

\subsection{Interaction Models}\label{interaction_model}

An interaction model \begin{math}f:\mathcal{E} \times \mathcal{R} \times \mathcal{E} \rightarrow \mathbb{R}\end{math} computes a real-valued score representing the plausibility of a triple \begin{math}(h,r,t) \in \mathbb{K}\end{math} given the embeddings for the entities and relations.
In general, a larger score indicates a higher plausibility.
The interpretation of the score value is model-dependent, and usually, it cannot be directly interpreted as a probability.
We follow~\cite{wang2017knowledge,ji2021survey} and categorize interaction models into \textit{translational distance} based and \textit{semantic matching} based interaction models. 
Translational distance interaction models compute the plausibility of triples based on a distance function, e.g., Euclidean distance between (projected) entities, and semantic similarity matching models exploit the similarity of the latent features usually induced by inner a product formulation.

\subsubsection{Translational Distance Interaction Models}

\textbf{\acl{um}} The \acl{um} (\ac{um})~\cite{bordes2014semantic} scores a triple by computing the distance between the head and tail entity
\begin{equation}\label{unstructured_model_scoring_func}
    f(h, t) = - \|\textbf{h}  - \textbf{t}\|_{2}^{2} \enspace,
\end{equation}
where \begin{math}\textbf{h}, \textbf{t} \in \mathbb{R}^d\end{math} are the embeddings of head and tail entity, respectively.
A small distance between these embeddings indicates a plausible triple.
In the \ac{um}, relations are not considered, and therefore, it cannot distinguish between different relationship types.
However, the model can be beneficial for learning embeddings for \acp{kg} that contain only a single relationship type or only equivalent relationship types, e.g.\ \textit{GrandmotherOf} and \textit{GrandmaOf}.
Moreover, it may serve as a baseline to interpret the performance of relation-aware models.

\textbf{\acl{se}} \acl{se} (\ac{se})~\cite{Bordes2011} models each relation by two matrices \begin{math}\textbf{M}_{r}^{h}, \textbf{M}_{r}^{t} \in \mathbb{R}^{d \times d}\end{math} that perform relation-specific projections of the head and tail embeddings: 
\begin{equation}
    f(h, r, t) = - \|\textbf{M}_{r}^{h} \textbf{h}  - \textbf{M}_{r}^{t} \textbf{t}\|_{1} \enspace.
\end{equation}
As before, $\mathbf{h}, \mathbf{t} \in \mathbb{R}^d$ are the embeddings of head and tail entity, respectively.
By employing different projections for the embeddings of the head and tail entities, \ac{se} explicitly distinguishes between the subject- and object-role of an entity.

\textbf{TransE}\label{TransE} TransE~\cite{Bordes2013} models relations as a translation of head to tail embeddings, i.e.\ \begin{math}\textbf{h} + \textbf{r} \approx \textbf{t}\end{math}.
Thus, the interaction model is defined as: 
\begin{equation}
    f(h, r, t) = - \|\textbf{h} + \textbf{r} - \textbf{t}\|_{p} \enspace,
\end{equation}
with $p \in \{1, 2\}$ is a hyper-parameter.
A major advantage of TransE is its computational efficiency which enables its usage for large scale \acp{kg}.
However, it inherently cannot model 1-N, N-1, and N-M relations: assume \begin{math}(h,r,t_{1}), (h,r,t_{2}) \in \mathcal{K}\end{math}, then the model adapts the embeddings in order to ensure \begin{math}\textbf{h} + \textbf{r} \approx \textbf{t}_1\end{math} and \begin{math}\textbf{h} + \textbf{r} \approx \textbf{t}_2\end{math} which results in \begin{math}\textbf{t}_1 \approx \textbf{t}_2\end{math}.

\textbf{TransH} TransH~\cite{Wang2014} is an extension of TransE that specifically addresses the limitations of TransE in modeling 1-N, N-1, and N-M relations.
In TransH, each relation is represented by a hyperplane, or more specifically a normal vector of this hyperplane \begin{math}\textbf{w}_{r} \in \mathbb{R}^d\end{math}, %
and a vector \begin{math}\textbf{d}_{r} \in \mathbb{R}^d\end{math} that lies in the hyperplane.
To compute the plausibility of a triple \begin{math}(h,r,t)\in \mathbb{K}\end{math}, the head embedding \textbf{h} $\in \mathbb{R}^d$ and the tail embedding \textbf{t} $\in \mathbb{R}^d$ are first projected onto the relation-specific hyperplane: 
$\mathbf{h}_{r} = \mathbf{h} - \mathbf{w}_{r}^\top \mathbf{h} \mathbf{w}_r$ and $\textbf{t}_{r} = \textbf{t} - \textbf{w}_{r}^\top \textbf{t} \textbf{w}_r$.
Then, the projected embeddings are used to compute the score for the triple \begin{math}(h,r,t)\end{math}:
\begin{equation} \label{eq:TransH_score_fct}
    f(h, r, t) = -\|\textbf{h}_{r} + \textbf{d}_r - \textbf{t}_{r}\|_{2}^2 \enspace .
\end{equation}

\textbf{TransR} TransR~\cite{Lin2015} is an extension of TransH that explicitly considers entities and relations as different objects and therefore represents them in different vector spaces.
For a triple \begin{math}(h,r,t) \in \mathbb{K}\end{math}, the entity embeddings,  \begin{math}\textbf{h}\end{math}, \begin{math}\textbf{t}  \in \mathbb{R}^d\end{math}, are first projected into the relation space by means of a relation-specific projection matrix \begin{math}\textbf{M}_{r} \in \mathbb{R}^{k \times d}\end{math}:
$\textbf{h}_{r} = \textbf{M}_{r}\textbf{h}$ and $\textbf{t}_{r} = \textbf{M}_{r}\textbf{t}$.
Finally, the score of the triple \begin{math}(h,r,t)\end{math} is computed:
\begin{equation} \label{eq:TransR_score_fct}
    f(h,r,t) = -\|\textbf{h}_{r} + \textbf{r} - \textbf{t}_{r}\|_{2}^2 \enspace 
\end{equation}
where \begin{math}\textbf{r} \in \mathbb{R}^k\end{math}.

\textbf{TransD} TransD~\cite{Ji2015} is an extension of TransR that, like TransR, considers entities and relations 
as objects living in different vector spaces.
However,
instead of performing the same relation-specific projection for all entity embeddings, entity-relation-specific projection matrices
\begin{math}\textbf{M}_{r,h}, \textbf{M}_{t,h}  \in \mathbb{R}^{k \times d}\end{math} are constructed.
To do so, all head entities, tail entities, and relations
are represented by two vectors, \begin{math}\textbf{h}, \textbf{h}_p, \textbf{t}, \textbf{t}_p \in \mathbb{R}^d\end{math} and \begin{math}\textbf{r}, \textbf{r}_p \in \mathbb{R}^k\end{math}, respectively.
The first set of embeddings is used for calculating the entity-relation-specific projection matrices: $\textbf{M}_{r,h} = \textbf{r}_{p} \textbf{h}_{p}^{T} + \tilde{\textbf{I}}$ and $\textbf{M}_{r,t} = \textbf{r}_{p} \textbf{t}_{p}^{T} + \tilde{\textbf{I}}$,
where \begin{math}\tilde{\textbf{I}} \in \mathbb{R}^{k \times d}\end{math} is a \begin{math}k \times d\end{math} matrix with ones on the diagonal and zeros elsewhere.
Next, $\textbf{h}$ and $\textbf{t}$ are projected into the relation space by means of the constructed projection matrices: $\textbf{h}_{r} = \textbf{M}_{r,h} \textbf{h}$ and $\textbf{t}_{r} = \textbf{M}_{r,t} \textbf{t}$.
Finally, the plausibility score for \begin{math}(h,r,t) \in \mathbb{K}\end{math} is given by:
\begin{equation} \label{eq:TransD_score_fct}
    f(h,r,t) = -\|\textbf{h}_{r} + \textbf{r} - \textbf{t}_{r}\|_{2}^2 \enspace.
\end{equation}

\textbf{RotatE} RotatE~\cite{Sun2019} models relations as rotations from head to tail entities in the complex space: $\textbf{t}= \textbf{h} \odot \textbf{r}$, 
where \begin{math}\textbf{h},\textbf{r},\textbf{t} \in \mathbb{C}^{d}\end{math} and \begin{math}|r_i| = 1, \end{math} that is the complex elements of \begin{math}\textbf{r}\end{math} are restricted to have a modulus of one.
Because of the latter, $r_i$ can be represented as \begin{math}e^{{i\theta_{r,i}}}\end{math}, which corresponds to a counterclockwise rotation by \begin{math}\theta_{r,i}\end{math} radians.
The interaction model is then defined as:%
\begin{equation}
f(h,r,t) = -\|\textbf{h} \odot \textbf{r} - \textbf{t}\| \enspace,
\end{equation}

which allows to model \textit{symmetry}, \textit{antisymmetry}, \textit{inversion}, and \textit{composition}~\cite{Sun2019}.

\textbf{MuRE} MuRE~\cite{DBLP:conf/nips/BalazevicAH19} is the Euclidean counterpart of MuRP, a hyperbolic interaction model that is capable of effectively modeling hierarchies in \ac{kg}.
Its interaction model involves a distance function:

\begin{equation}
    f(h,r,t) = - \|\mathbf{R}\mathbf{h} -\mathbf{t}+\mathbf{r}\|_{2}^{2} + \mathbf{b_h} + \mathbf{b_t}
\end{equation}

where the head entity is transformed by the diagonal matrix $\mathbf{R} \in \mathbf{R}^{d \times d}$ and the tail entity by the relation r.  $\mathbf{b_h}$ and $\mathbf{b_t}$ represent scalar offsets.

\textbf{KG2E} KG2E~\cite{he2015learning} aims to explicitly model (un)certainties in entities and relations (e.g.\ influenced by the number of triples observed for these entities and relations).
Therefore, entities and relations are represented by probability distributions, in particular by multi-variate Gaussian distributions \begin{math}\bm{\mathcal{N}}_i(\bm{\mu}_i,\bm{\Sigma}_i)\end{math} where the mean \begin{math}\bm{\mu}_i \in \mathbb{R}^d\end{math} denotes the position in the vector space and the diagonal variance \begin{math}\bm{\Sigma}_i \in \mathbb{R}^{d \times d}\end{math} %
models the uncertainty.
Inspired by the TransE model, relations are modeled as transformations from head to tail entities: \begin{math}\bm{\mathcal{H}} - \bm{\mathcal{T}} \approx \bm{\mathcal{R}}\end{math} where \begin{math}\bm{\mathcal{H}} \sim \bm{\mathcal{N}}_h(\bm{\mu}_h,\bm{\Sigma}_h)\end{math}, \begin{math}\bm{\mathcal{H}} \sim \bm{\mathcal{N}}_t(\bm{\mu}_t,\bm{\Sigma}_t)\end{math}, \begin{math}\bm{\mathcal{R}} \sim \bm{\mathcal{P}_r} =  \bm{\mathcal{N}}_r(\bm{\mu}_r,\bm{\Sigma}_r)\end{math} and \begin{math}\bm{\mathcal{H}} - \bm{\mathcal{T}} \sim \bm{\mathcal{P}_e =  \bm{\mathcal{N}}_{h-t}(\bm{\mu}_h - \bm{\mu}_t,\bm{\Sigma}_h + \bm{\Sigma}_t)}\end{math} (since head and tail entities are considered to be independent with regards to the relations).
The interaction model measures the similarity between \begin{math}\bm{\mathcal{P}_e}\end{math} and \begin{math}\bm{\mathcal{P}_r}\end{math} by means of the \ac{kl} divergence:
\begin{equation}
    \begin{split}
        f(h,r,t) = \mathcal{D_{KL}}(\bm{\mathcal{P}}_e, \bm{\mathcal{P}}_r) \\
        = \frac{1}{2}\Big\{tr(\bm{\Sigma}_{r}^{-1}\bm{\Sigma}_{e}) + (\bm{\mu}_{r} - \bm{\mu}_{e})^{T} \bm{\Sigma}_{r}^{-1} (\bm{\mu}_{r} - \bm{\mu}_{e}) \\
        - log(\frac{det(\bm{\Sigma}_e)}{det(\bm{\Sigma}_r)}) - d\Big\} \enspace.
    \end{split} 
\end{equation}
Besides the asymmetric KL divergence, the authors propose a symmetric variant which uses the expected likelihood.

\subsubsection{Semantic Matching Interaction Models}

\textbf{RESCAL} RESCAL~\cite{Nickel2011} is a bilinear model that models entities as vectors and relations as matrices.
The relation matrices \begin{math}\textbf{W}_{r} \in \mathbb{R}^{d \times d}\end{math} %
contain weights $w_{i,j}$ that 
capture the amount of interaction between the $i$-th latent factor of \begin{math}\textbf{h} \in \mathbb{R}^{d}\end{math} and the $j$-th latent factor of \begin{math}\textbf{t} \in \mathbb{R}^{d}\end{math}~\cite{nickel2015review,Nickel2011}.
Thus, the plausibility score of $(h,r,t) \in \mathbb{K}$ is given by:
\begin{equation} \label{eq:RESCAL_score_fct}
    f(h,r,t) = \textbf{h}^{T} \textbf{W}_{r} \textbf{t} = \sum_{i=1}^{d}\sum_{j=1}^{d} w_{ij}^{(r)} h_{i} t_{j}
\end{equation}

\textbf{DistMult} DistMult~\cite{Yang2014} is a simplification of RESCAL where the relation matrices \begin{math}\textbf{W}_{r} \in \mathbb{R}^{d \times d}\end{math} are restricted to diagonal matrices:
\begin{equation} \label{eq:DistMult_score_fct}
    f(h,r,t) = \textbf{h}^{T} \textbf{W}_r \textbf{t} = \sum_{i=1}^{d}\textbf{h}_i \cdot diag(\textbf{W}_r)_i \cdot \textbf{t}_i \enspace.
\end{equation}
Because of its restriction to diagonal matrices DistMult is computational more efficient than RESCAL, but at the same time less expressive.
For instance, it is not able to model anti-symmetric relations, since \begin{math}f(h,r, t) = f(t,r,h)\end{math}.

\textbf{ComplEx} ComplEx~\cite{Trouillon2016} is an extension of DistMult that uses complex valued representations for the entities and relations.
Entities and relations are represented as vectors \begin{math}\textbf{h}, \textbf{r}, \textbf{t} \in \mathbb{C}^d\end{math}, and the plausibility score is computed using the Hadamard product:
\begin{equation}
    \begin{aligned}
        f(h,r,t) 
        =&  Re(\mathbf{h}\odot\mathbf{r}\odot\mathbf{t}) %
    \end{aligned}
\end{equation}

where $Re(\textbf{x})$ denotes the real component of the complex valued vector $\mathbf{x}$.
Because the Hadamard product is not commutative in the complex space, ComplEx can model anti-symmetric relations in contrast to DistMult.

\textbf{QuatE} QuatE~\cite{DBLP:conf/nips/0007TYL19} learns hypercomplex valued representations (quaternion embeddings) for entities and relations, i.e., $\mathbf{e_i}, \mathbf{r_j} \in \mathbb{H}^d$. Hypercomplex representations extend complex representations by representing each number with one real and three imaginary components. In QuatE, relations are modelled as rotations in the hypercomplex space. More precisely, the relation is used to rotate the head entity: $\mathbf{h_r} = \mathbf{h} \otimes \mathbf{r}$, where in this context $\otimes$ represents the Hamilton product. The final score is obtained by computing the inner product between the rotated head and the the tail entity:

\begin{equation}
    f(h,r,t) = \mathbf{h_r} \cdot \mathbf{t}
\end{equation}

In contrast to ComplEx, QuatE is capable of modeling \textit{composition} patterns.

\textbf{SimplE} SimplE~\cite{kazemi2018simple} is an extension of \textbf{\ac{cp}}~\cite{kazemi2018simple}, one of the early tensor factorization approaches.
In \ac{cp}, each entity \begin{math}e \in \mathcal{E}\end{math} is represented by two vectors \begin{math}\textbf{h}_{e}, \textbf{t}_{e} \in \mathbb{R}^d\end{math} and each relation by a single vector \begin{math}\textbf{r} \in \mathbb{R}^d\end{math}.
Depending whether an entity participates in a triple as the head or tail entity, either \begin{math}\textbf{h}_{e}\end{math} or \begin{math}\textbf{t}_{e}\end{math} is used.
Both entity representations are learned independently, i.e.\ observing a triple \begin{math}(e_{1},r,e_{2})\end{math}, the method only updates \begin{math}\textbf{h}_{e_{1}}\end{math} and \begin{math}\textbf{t}_{e_{2}}\end{math}.
In contrast to \ac{cp}, SimplE introduces for each relation \begin{math}r\end{math} the inverse relation \begin{math}r'\end{math}, and formulates  %
the interaction model based on both:%
\begin{equation}
f(h,r,t) = \frac{1}{2}\left(\left\langle\textbf{h}_{e_{i}}, \textbf{r}, \textbf{t}_{e_{j}}\right\rangle + \left\langle\textbf{h}_{e_{j}}, \textbf{r}', \textbf{t}_{e_{i}}\right\rangle\right) \enspace.
\end{equation}
Therefore, for each triple \begin{math}(e_{1},r,e_{2}) \in \mathbb{K}\end{math}, both \begin{math}\textbf{h}_{e_{1}}\end{math} and \begin{math}\textbf{t}_{e_{2}}\end{math} as well as
\begin{math}\textbf{h}_{e_{2}}\end{math} and \begin{math}\textbf{t}_{e_{1}}\end{math}
are updated~\cite{kazemi2018simple}.

\textbf{TuckER} TuckER~\cite{balavzevic2019tucker} is a linear model that is based on the tensor factorization method Tucker~\cite{tucker1964extension} in which a three-mode tensor $\mathfrak{X} \in \mathbb{R}^{I \times J \times K}$ is decomposed into a set of factor matrices \begin{math}\textbf{A} \in \mathbb{R}^{I \times P}\end{math}, \begin{math}\textbf{B} \in \mathbb{R}^{J \times Q}\end{math}, and \begin{math}\textbf{C} \in \mathbb{R}^{K \times R}\end{math} and a core tensor \begin{math}\mathfrak{Z} \in \mathbb{R}^{P \times Q \times R}\end{math} (of lower rank): $\mathfrak{X} \approx \mathfrak{Z} \times_1 \textbf{A} \times_2 \textbf{B} \times_3 \textbf{C}$,
where \begin{math}\times_n\end{math} is the tensor product, with \begin{math}n\end{math} denoting along which mode the tensor product is computed.
In TuckER, a \ac{kg} is considered as a binary tensor which is factorized using the Tucker factorization where \begin{math}\textbf{E} = \textbf{A} = \textbf{C} \in \mathbb{R}^{n_{e} \times d_e}\end{math} denotes the entity embedding matrix, \begin{math}\textbf{R} = \textbf{B} \in \mathbb{R}^{n_{r} \times d_r}\end{math} represents the relation embedding matrix, and \begin{math}\mathfrak{W} = \mathfrak{Z} \in \mathbb{R}^{d_e \times d_r \times d_e}\end{math} is the \textit{core tensor} that indicates the extent of interaction between the different factors.
The interaction model is defined as:
\begin{equation}
    f(h,r,t) = \mathfrak{W} \times_1 \textbf{h} \times_2 \textbf{r} \times_3 \textbf{t} \enspace, 
\end{equation}
where $\textbf{h},\textbf{t}$ correspond to rows of $\textbf{E}$ and $\textbf{r}$ to a row of $\textbf{R}$.

\textbf{ProjE} ProjE~\cite{shi2017proje} is a neural network-based approach with a \textit{combination} and a \textit{projection} layer.
The interaction model first combines \begin{math}h\end{math} and \begin{math}r\end{math} by a combination operator~\cite{shi2017proje}: $\textbf{h} \otimes \textbf{r} = \textbf{D}_e \textbf{h} + \textbf{D}_r \textbf{r} + \textbf{b}_c$,
where \begin{math}\textbf{D}_e, \textbf{D}_r \in \mathbb{R}^{k \times k}\end{math} are diagonal matrices which are used as shared parameters among all entities and relations, and \begin{math}\textbf{b}_c \in \mathbb{R}^{k}\end{math} represents the candidate bias vector shared across all entities.
Next, the score for the triple $(h,r,t) \in \mathbb{K}$ is computed:
\begin{equation}
    f(h, r, t) = g(\textbf{t} \ z(\textbf{h} \otimes \textbf{r}) + \textbf{b}_p) \enspace,
\end{equation}
where \begin{math}g\end{math} and \begin{math}z\end{math} are activation functions, and \begin{math}\textbf{b}_p\end{math} represents the shared projection bias vector.

\textbf{HolE} Holographic embeddings (HolE)~\cite{Nickel2016} make use of the circular correlation operator to compute interactions between latent features of entities and relations:
\begin{equation}
    f(h,r,t) = \sigma(\textbf{r}^{T}(\textbf{h} \star \textbf{t})) \enspace.
\end{equation}
where the circular correlation \begin{math}\star: \mathbb{R}^d \times \mathbb{R}^d \rightarrow \mathbb{R}^d\end{math} is defined as $[\textbf{a} \star \textbf{b}]_i = \sum_{k=0}^{d-1} \textbf{a}_{k} * \textbf{b}_{(i+k)\ mod \ d}$.
By using the correlation operator each component \begin{math}[\textbf{h} \star \textbf{t}]_i\end{math} represents a sum over a fixed partition over pairwise interactions.
This enables the model to put semantic similar interactions into the same partition and share weights through \begin{math}\textbf{r}\end{math}.
Similarly irrelevant interactions of features could also be placed into the same partition which could be assigned a small weight in \begin{math}\textbf{r}\end{math}.

\textbf{ERMLP} ERMLP~\cite{Dong2014} is a multi-layer perceptron based approach that uses a single hidden layer and represents entities and relations as vectors.
In the input-layer, for each triple the embeddings of head, relation, and tail are concatenated and passed to the hidden layer.
The output-layer consists of a single neuron that computes the plausibility score of the triple:
\begin{equation}
    f(h,r,t) = \textbf{w}^{T} g(\textbf{W} [\textbf{h}; \textbf{r}; \textbf{t}]),
\end{equation}
where \begin{math}\textbf{W} \in \mathbb{R}^{k \times 3d}\end{math} represents the weight matrix of the hidden layer, \begin{math}\textbf{w} \in \mathbb{R}^{k}\end{math}, the weights of the output layer, and \begin{math}g\end{math} denotes an activation function such as the hyperbolic tangent.

\textbf{\Acl{ntn}} The \acf{ntn}~\cite{socher2013reasoning} uses a bilinear tensor layer instead of a standard linear neural network layer:
\begin{equation}
    f(h,r,t) = \textbf{u}_{r}^{T} \cdot \tanh(\textbf{h} \mathfrak{W}_{r} \textbf{t} + \textbf{V}_r [\textbf{h};\textbf{t}] + \textbf{b}_r) \enspace,
\end{equation}
where \begin{math}\mathfrak{W}_r \in \mathbb{R}^{d \times d \times k}\end{math} is the relation specific tensor, and the weight matrix \begin{math}\textbf{V}_r \in \mathbb{R}^{k \times 2d}\end{math}, %
the bias vector \begin{math}\textbf{b}_r, \end{math} and the weight vector \begin{math}\textbf{u}_r \in \mathbb{R}^k\end{math} %
are the standard parameters of a neural network, which are also relation specific.
The result of the tensor product \begin{math}\textbf{h} \mathfrak{W}_{r} \textbf{t}\end{math} is a vector \begin{math}\textbf{x} \in \mathbb{R}^k\end{math} where each entry \begin{math}x_i\end{math} is computed based on the slice \begin{math}i\end{math} of the tensor \begin{math}\mathfrak{W}_{r}\end{math}: \begin{math}\textbf{x}_i = \textbf{h}\mathfrak{W}_{r}^{i} \textbf{t}\end{math}~\cite{socher2013reasoning}.
As indicated by the interaction model, \ac{ntn} defines for each relation a separate neural network which makes the model very expressive, but at the same time computationally expensive.

\textbf{ConvKB}\label{ConvKB} ConvKB~\cite{nguyen2017novel} uses a \ac{cnn} whose feature maps capture global interactions of the input.
Each triple \begin{math}(h,r,t) \in \mathbb{K}\end{math} is represented as a input matrix \begin{math}\textbf{A} = [\textbf{h}; \textbf{r}; \textbf{t}] \in \mathbb{R}^{d \times 3}\end{math} in which the columns represent the embeddings for \begin{math}h,r \end{math} and \begin{math}t\end{math}.
In the convolution layer, a set of convolutional filters \begin{math}\bm{\omega}_i \in \mathbb{R}^{1 \times 3}, i=1, \dots, \tau,\end{math} are applied on the input in order to compute for each dimension global interactions of the embedded triple.
Each \begin{math}\bm{\omega}_i \end{math} is applied on every row of \begin{math}\textbf{A}\end{math} creating a feature map \begin{math}\textbf{v}_i = [v_{i,1},...,v_{i,d}] \in \mathbb{R}^d\end{math}:%
\begin{equation}
    \textbf{v}_i = g(\bm{\omega}_j \textbf{A} + \textbf{b}) \enspace,
\end{equation}

where \begin{math}\textbf{b} \in \mathbb{R}\end{math} denotes a bias term and \begin{math}g\end{math} an activation function which is employed element-wise.
Based on the resulting feature maps $\textbf{v}_1, \dots, \textbf{v}_{\tau}$, the plausibility score of a triple %
is given by:
\begin{equation}
    f(h,r,t) = [\textbf{v}_i; \ldots ;\textbf{v}_\tau] \cdot \textbf{w}\enspace,
\end{equation}
where \begin{math}[\textbf{v}_i; \ldots ;\textbf{v}_\tau] \in \mathbb{R}^{\tau d \times 1}\end{math} and \begin{math}\textbf{w} \in \mathbb{R}^{\tau d \times 1} \end{math} is a shared weight vector. %
ConvKB may be seen as a restriction of ER-MLP with a certain weight sharing pattern in the first layer.

\textbf{ConvE}\label{ConvE} ConvE~\cite{Dettmers2018} is a \ac{cnn}-based approach.
For each triple $(h,r,t)$, the input to ConvE is a matrix $\mathbf{A} \in \mathbb{R}^{2 \times d}$ where the first row of $\textbf{A}$ represents $\mathbf{h} \in \mathbb{R}^d$ and the second row represents $\mathbf{r} \in \mathbb{R}^d$.
$\mathbf{A}$ is reshaped to a matrix $\mathbf{B} \in \mathbb{R}^{m \times n}$ where the first $m/2$ half rows represent $\mathbf{h}$ and the remaining $m/2$ half rows represent $\mathbf{r}$.
In the convolution layer, a set of \textit{2-dimensional} convolutional filters $\Omega = \{\bm{\omega}_i \ | \ \bm{\omega}_i \in \mathbb{R}^{r \times c}\}$ are applied on $\mathbf{B}$ that capture interactions between $\mathbf{h}$ and $\mathbf{r}$.
The resulting feature maps are reshaped and concatenated in order to create a feature vector $\mathbf{v} \in \mathbb{R}^{|\Omega|rc}$.
In the next step, $\mathbf{v}$ is mapped into the entity space using a linear transformation $\mathbf{W} \in \mathbb{R}^{|\Omega|rc \times d}$, that is $\mathbf{e}_{h,r} = \mathbf{v}^{T} \textbf{W}$.
The score for the triple $(h,r,t) \in \mathbb{K}$ is then given by:
\begin{equation}
    f(h,r,t) = \textbf{e}_{h,r} \mathbf{t} \enspace.
\end{equation}
Since the interaction model can be decomposed into $f(h,r,t) = \left\langle f'(\mathbf{h}, \mathbf{r}), \mathbf{t} \right\rangle$, the model is particularly designed to 1-N scoring, i.e.\ efficient computation of scores for $(h,r,t)$ for fixed $h,r$ and many different $t$.

\subsection{Training Approaches}\label{training_assumption}

Because most \acp{kg} contain only positive examples, we require training approaches involving techniques such as negative sampling to avoid over-generalization to true facts.
Here, we describe two common training approaches found in the literature: the \acf{lcwa} and the \acf{slcwa}.
It should be noted that the \ac{lcwa} and the \ac{slcwa} do not affect the evaluation. 

\begin{figure}
    \centering
    \includegraphics[width=\linewidth]{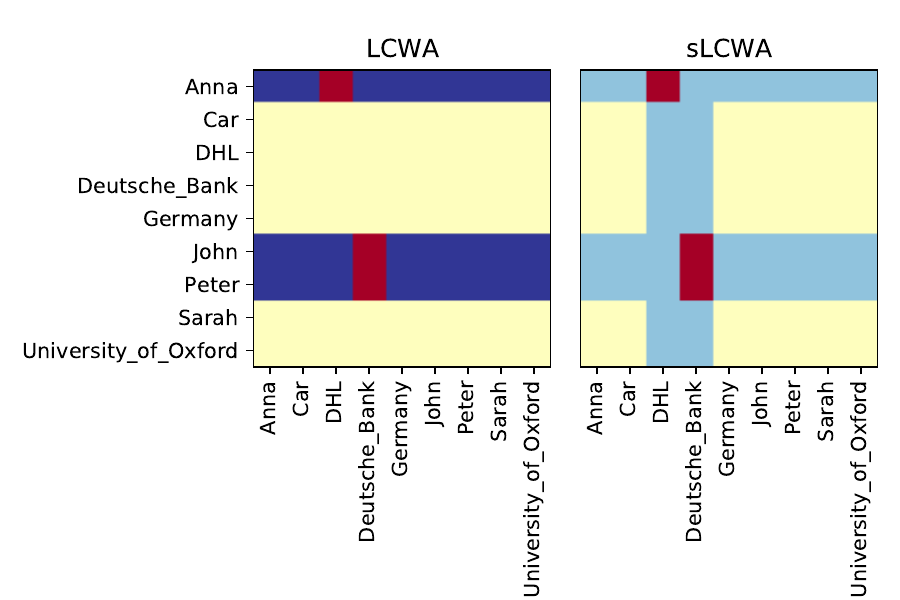}
    \caption{Visualization of different training approaches for the relation \texttt{works\_at} in the \ac{kg} in Figure~\ref{fig:exemplary_kg}. Red color indicates positive examples, i.e. true triples present in the \ac{kg}. Dark blue color denotes triples used as negative examples in \ac{lcwa}. Light blue color sampling candidates for negative examples in \ac{slcwa}. Yellow color indicates triples that are not considered.}
    \label{fig:assumptions}
\end{figure}

\subsubsection{\Acl{lcwa}}\label{ta_lcwa}

The \ac{lcwa} was introduced by~\cite{Dong2014} and used in subsequent works as an approach to generate negative examples during training~\cite{Dettmers2018,balavzevic2019tucker}. In this setting, for any triple \begin{math}(h,r,t) \in \mathcal{K}\end{math} that has been observed, a set $\mathcal{T}^-(h,r)$ of negative examples is created by considering all triples \begin{math}(h, r, t_i) \notin \mathcal{K}\end{math} as false. Therefore, for our exemplary \ac{kg} (Figure~\ref{fig:exemplary_kg}) for the pair \textit{(Peter, works\_at)}, the triple \textit{(Peter, works\_at, DHL)} is a false fact since for this pair only the triple \textit{(Peter, works\_at, Deutsche Bank)} is part of the \ac{kg}.
Similarly, we can construct $\mathcal{H}^-(r,t)$ based on all triples \begin{math}(h_i, r, t) \notin \mathcal{K}\end{math}, or $\mathcal{R}^-(h,t)$ based on the triples \begin{math}(h, r_i, t) \notin \mathcal{K}\end{math}. Constructing $\mathcal{R}^-(h,t)$ is a popular choice in visual relation detection domain~\cite{zhang2017visual,sharifzadeh2019improving}.
However, most of the works in knowledge graph modeling construct only $\mathcal{T}^-(h, r)$ as the set of negative examples, and in the context of this work refer to $\mathcal{T}^-(h, r)$ as the set of negatives examples when speaking about \ac{lcwa}.

\subsubsection{\Acl{slcwa}}\label{ta_owa}
Under the \acf{slcwa}, instead of considering all possible triples $(h,r,t_i) \notin \mathcal{K}$, $(h_i,r,t) \notin \mathcal{K}$ or $(h,r_i,t) \notin \mathcal{K}$ as false, we randomly take samples of these sets.

Two common approaches for generating negative samples are \ac{uns}~\cite{Bordes2013} and \ac{bns}~\cite{Wang2014} in which negative triples are created by corrupting a positive triple \begin{math}(h,r,t) \in \mathcal{K}\end{math} by replacing either \begin{math}h\end{math} or \begin{math}t\end{math}.
We denote with $\mathcal{N}$ %
the set of all potential negative triples:

\begin{eqnarray}\label{eq:negative_triples}
    \mathcal{T}(h, r) &=& \{(h, r, t') \mid t' \in \mathcal{E} \land t' \neq t\}\\
    \mathcal{H}(r, t) &=& \{(h', r, t) \mid h' \in \mathcal{E} \land h' \neq h\}\\
    \mathcal{N} &=& \bigcup_{(h,r,t) \in \mathcal{K}} \mathcal{T}(h, r) \cup \mathcal{H}(r, t)
     \enspace.
\end{eqnarray}

Theoretically, we would need to exclude all positive triples from this set of candidates for negative triples, i.e., $\mathcal{N}^- = \mathcal{N} \setminus \mathcal{K}$.
In practice, however, since usually $|\mathcal{N}| \gg |\mathcal{K}|$, the likelihood of generating a false negative is rather low.
Therefore, the additional filter step is often omitted to lower computational cost.
It should be taken into account that a corrupted triple that is \textit{not part }of the \ac{kg} can represent a true fact.

\Ac{uns} and \ac{bns} differ in the way they define sample weights for \begin{math}(h',r,t)\end{math} or \begin{math}(h,r,t')\end{math}:

\textbf{\Acl{uns}} With \acf{uns}~\cite{Bordes2013}, the first step is to randomly (uniformly) determine whether \begin{math}h\end{math} or \begin{math}t\end{math} shall be corrupted for a positive triple \begin{math}(h,r,t) \in \mathcal{K}\end{math}.
Afterwards, an entity \begin{math}e \in \mathcal{E}\end{math} is uniformly sampled and selected as the corrupted head/tail entity.

\textbf{\Acl{bns}} With \acf{bns}~\cite{Wang2014}, the probability of corrupting \begin{math}h\end{math} or \begin{math}t\end{math} in \begin{math}(h,r,t) \in \mathcal{K}\end{math} is determined by the property of the relation \begin{math}r\end{math}: if the relation is a \textit{one-to-many} relation (e.g.\ \textit{motherOf}), \ac{bns} assigns a higher probability to replace \begin{math}h\end{math}, and if it is a \textit{many-to-one} relation (e.g.\ \textit{bornIn}) it assigns a higher probability to replace \begin{math}t\end{math}.
More precisely, for each relation \begin{math}r \in \mathcal{R}\end{math} the average number of tails per head (\textit{tph}) and heads per tail (\textit{hpt}) are first computed.
These statistics are then used to define a Bernoulli distribution with parameter \begin{math}\frac{tph}{tph + hpt}\end{math}.
For a triple \begin{math}(h,r,t) \in \mathcal{K}\end{math} the head is corrupted with probability \begin{math}\frac{tph}{tph + hpt}\end{math} and the tail with probability \begin{math}\frac{hpt}{tph + hpt}\end{math}.
The described approach reduces the chance of creating corrupted triples that represent true facts~\cite{Wang2014}.

\subsection{Loss Functions}\label{loss_functions}

The loss function can have a significant influence on the performance of \acp{kgem}~\cite{mohamed2019loss}.
In the following, we describe \textit{pointwise}, \textit{pairwise}, and \textit{setwise} loss functions that have been frequently be used within \acp{kgem}.
For additional discussion and a slightly different categorization we refer to the work of Mohamed \textit{et al.}~\cite{mohamed2019loss}.

\subsubsection{\textbf{Pointwise Loss Functions}}\label{pwlf}

Let $f$ denote the interaction model of a \ac{kgem}.
With $t_i$, we denote a triple (i.e. $t_i \in \mathbb{K}$), and with  $l_i \in \{0,1\}$ or  $\hat{l}_i \in \{-1, 1\}$ its corresponding label, where 1 corresponds to the label of the positive triples, and 0 / -1 to the label of the negative triples.
Pointwise loss functions compute an independent loss term for each triple-label pair, i.e. for a batch $B = \{(t_i, l_i)\}_{i=1}^{|B|}$, the loss is given as
\begin{equation}
    \mathcal{L} = \frac{1}{|B|} \sum \limits_{(t_i, l_i) \in B} L(t_i, l_i)
\end{equation}
In the following, we describe four different pointwise losses:
The \emph{square error loss}, \emph{\acf{bcel}}, \emph{pointwise hinge loss}, and \emph{logistic loss}.

\textbf{Square Error Loss} The square error loss function computes the squared difference between the predicted scores and the labels $l_i \in \{0, 1\}$~\cite{mohamed2019loss}:
\begin{equation}
    L(t_i, l_i) = \frac{1}{2} (f(t_{i}) - l_{i})^2
\end{equation}
The squared error loss strongly penalizes predictions that deviate considerably from the labels, and is usually used for regression problems.
For simple models it often permits more efficient optimization algorithms involving analytical solutions of sub-problems, e.g.\ the Alternating Least Squares algorithm used by~\cite{Nickel2011}.

\textbf{\Acl{bcel}} The \acl{bcel} is defined as~\cite{Dettmers2018}:
\begin{equation}
\begin{aligned}
L(t_i, l_i) = 
&-(l_{i} \cdot \log(\sigma(f(t_{i}))) \\
&+ (1 - l_{i}) \cdot \log(1 - \sigma(f(t_{i})))),
\end{aligned}
\end{equation}
where $l_i \in \{0, 1\}$ and $\sigma$ represents the logistic sigmoid function.
Thus, the problem is framed as a binary classification problem of triples, where the model's outputs are regarded as logits. 
The loss is not well-suited for translational distance models because these models produce a negative distance as score and cannot produce positive model outputs.
ConvE and TuckER were originally trained in a multi-class setting using the \acl{bcel} where each $(h,r)$-pair has been classified against $e \in \mathcal{E}$ simultaneously, i.e.,\ if $|\mathcal{E}|=n$, the label vector for each $(h,r)$-pair has $n$ entries indicating whether the triple $(h,r,e_i)$ is (not) part of the \ac{kg}, and along each dimension of the label vector a binary classification is performed.
It should be noted that there exist different implementation variants of the \acl{bcel} that address numerical stability. ConvE and TuckER employed a numerically unstable variant, and in the context of this work, we refer to this variant when referring to the \acl{bcel}.

\textbf{Pointwise Logistic Loss/\Acl{spl}} An alternative, but equivalent formulation of the \acl{bcel} is the pointwise logistic loss (or \acf{spl}):
\begin{equation}
L(t_i, l_i) = \log(1 + \exp(-\hat{l}_{i} \cdot f(t_{i}))  
\end{equation}
where $\hat{l}_{i} \in \{-1, 1\}$~\cite{mohamed2019loss}.
It has been used to train ComplEx, ConvKB, and SimplE.
We consider both variants separately because both have been used in different model implementations, and their implementation details might yield different results (e.g., to numerical stability).

\textbf{Pointwise Hinge Loss} The pointwise hinge loss sets the score of positive examples larger than a margin parameter $\lambda$ while reducing the scores of negative examples to values below $-\lambda$:
\begin{equation}
    L(t_i, l_i) = \max(0,\lambda - \hat{l}_{i} \cdot f(t_{i}))  
\end{equation}
where $\hat{l}_{i} \in \{-1, 1\}$.
The loss penalizes scores of positive examples which are smaller than $\lambda$, but does not impose any restriction on values $>\lambda$.
Similarly, negative scores larger than $-\lambda$ contribute to the loss, whereas all values smaller than $-\lambda$ do not have any loss contribution~\cite{mohamed2019loss}.
Thereby, the model is not encouraged to further optimize triples which are already predicted well enough (according to the margin parameter $\lambda$).

\subsubsection{\textbf{Pairwise Loss Functions}}
Next, we describe widely applied pairwise loss functions that are used within \acp{kgem}, namely the \textit{pairwise hinge loss} and the \textit{pairwise logistic loss}.
They both compare the scores of a positive triple $t^+$ and a negative triple $t^-$.
The negative triple in a pair is usually obtained by corrupting the positive one.
Thus, the pairs often share common head or tail entities and relations.
For a batch of pairs $B = \{(t^{+}_{i}, t^{-}_{i})\}_{i=1}^{|B|}$, the loss is given as
\begin{equation}
    \mathcal{L} = \frac{1}{|B|} \sum \limits_{(t^{+}_{i}, t^{-}_{i}) \in B} L(f(t^{-}_{i}) - f(t^{+}_{i})) \enspace.
\end{equation}
Hence, the loss function evaluates the difference in scores $\Delta = f(t^{-}_{i}) - f(t^{+}_{i})$ between a positive and a negative triple, rather than their absolute scores.
This is in accordance to the \ac{owa} assumption, where we do not assume to have negative labels, but just "less positive" ones.

\textbf{Pairwise Hinge Loss/\Acl{mrl}} The pairwise hinge loss or \acf{mrl} is given by
\begin{equation}
L(\Delta) = \max(0, \lambda + \Delta) \enspace.
\end{equation}

\textbf{Pairwise Logistic Loss} The pairwise logistic loss is defined as~\cite{mohamed2019loss}:
\begin{equation}
L(\Delta) = \log(1 + \exp(\Delta)) \enspace.
\end{equation}
Thus, it can be seen as a soft-margin formulation of the pairwise hinge loss with a margin of zero.

\subsubsection{\textbf{Setwise Loss Functions}}
Setwise loss functions neither compare individual scores, or pairs of them, but rather more than two triples' scores.
Here, we describe the \ac{nssal} and the \acf{cel} as examples of such loss functions that have been applied within \acp{kgem}~\cite{Sun2019,mohamed2019loss}.

\textbf{\Acl{nssal}} The \Acf{nssal} addresses the limitation that many negative examples are trivial and do not provide helpful information.
The authors of~\cite{Sun2019} propose to overcome this limitation by sampling negative samples according to the scores predicted by the interaction model~\cite{Sun2019}: 

\begin{equation}\label{eq:nnsal_dist}
    p((h_i', r, t_i')| (h_i,r_i,t_i)) = \frac{
        \exp(\alpha f(h_i', r, t_i'))
    }{
        \sum_{j=1}^{n}\exp(\alpha f(h_j',r, t_j'))
    }\enspace,
\end{equation}

where $(h_i, r_i, t_i) \in \mathcal{K}$ denotes a true triple, $\{(h_i', r, t_i')\}_{i=1}^K$ it's set of negative samples generated, and $\alpha \in \mathbb{R}$ a temperature parameter.
Because sampling from this distribution may be computationally expensive, the probabilities obtained by Equation~\ref{eq:nnsal_dist} are used to weight the generated negative examples in the loss function~\cite{Sun2019}.

\begin{equation}
    \begin{aligned}
    \mathcal{L} =& -\log(\sigma(\gamma + f(h, r, t)))\\ &-\sum_{i=1}^{K} p((h',r,t')) \cdot \log(\sigma(-(\gamma + f(h_i', r, t_i')))) \enspace.
    \end{aligned}
\end{equation}
Thus, negative samples for which the model predicts a high score relative to other samples are weighted stronger.

\textbf{\Acl{cel}} The \acf{cel} has been successfully applied together with 1-N scoring, i.e., predicting for each $(h,r)$-pair simultaneously a score for each possible tail entity, and framing the problem as a multi-class classification problem~\cite{kadlec2017knowledge,ruffinelli2020you}.
To apply the \ac{cel}, first, the labels are normalized in order to form a proper probability distribution. Second, the predicted scores for the tail entities of $(h,r)$-pair are normalized by a softmax:
\begin{equation}
     p(t \mid h, r) = \frac{\exp(f(h, r, t))}{\sum\limits_{t' \in \mathcal{E}}\exp(f(h, r, t'))} \enspace.
\end{equation}
Finally, the cross entropy between the distribution of the normalized scores and the normalized label distribution is computed:
\begin{equation}
    \mathcal{L} = - \sum\limits_{t' \in \mathcal{E}} \mathbb{I}[(h,r,t') \in \mathcal{K}] \cdot \log(p(t\mid h,r)) \enspace,
\end{equation}
where $\mathbb{I}$ denotes the indicator function. Note that this loss differs from the multi-class binary cross entropy as it applies a softmax normalization implying that this is a \textit{single-label} multi-class problem.

\subsection{Explicitly Modeling Inverse Relations}\label{sec:inverse_relations}

Inverse relations introduced by \cite{kazemi2018simple} and \cite{lacroix2018canonical} are explicitly modeled by extending the set of relations $\mathcal{R}$ by a set of inverse relations \begin{math}r_{inv} \in \mathcal{R}_{inv}\end{math} with $\mathcal{R}_{inv} \cap \mathcal{R} = \emptyset$.
This is achieved by training an inverse triple \begin{math}(t,r_{inv},h) \end{math} for each triple \begin{math}(h,r,t) \in \mathcal{K}\end{math}.
Equipping a \ac{kgem} with inverse relations implicitly doubles the relation embedding space of any model that has relation embeddings.
The goal is to alter the scoring function, such that the task of predicting the head entities for $(r,t)$ pairs becomes the task of predicting tail entities for $(t, r_{inv})$ pairs. The explicit training of the implicitly known inverse relations can lead to better model performance \cite{lacroix2018canonical} and can for some models increase the computational efficiency \cite{Dettmers2018}.

\section{Evaluation Metrics for \acp{kgem}}\label{sec:evaluation_of_models}

\Acp{kgem} are usually evaluated based on link prediction, which is on \ac{kg} defined as predicting the tail/head entities for $(h,r)$/$(r,t)$ pairs.
For instance, given queries of the form \textit{(Sarah, studied\_at, ?)} or \textit{(?, CEO\_of, Deutsche Bank)} the capability of a link predictor to predict the correct entities that answer the query, i.e.\ \textit{(Sarah, studied\_at, \textbf{University of Oxford})} and \textit{(\textbf{Sarah}, CEO\_of, Deutsche Bank)} is measured.

However, given the fact that usually true negative examples are not available, both the training and the test set contain only true facts.
For this reason, the evaluation procedure is defined as a ranking task in which the capability of the model to differentiate corrupted triples from known true triples is assessed~\cite{Bordes2013}.
For each test triple $t^{+} = (h,r,t) \in \mathcal{K}_{test}$ two sets of corrupted triples are constructed:
\begin{enumerate}
    \item $\mathcal{H}(r, t) = \{(h', r, t) \mid h' \in \mathcal{E} - \{h\}$ which contains all the triples where the head entity has been corrupted, and
    \item $\mathcal{T}(h, r) = \{(h, r, t') \mid t' \in \mathcal{E} - \{t\}\}$ that contains all the triples with corrupted tail entity.
\end{enumerate}
For each $t^+$ and its corresponding corrupted triples, the scores are computed and the entities sorted accordingly.
Next, the rank of every $t^+$ among its corrupted triples is determined, i.e.\ the position in the score-sorted list.

Among the corrupted triples in $\mathcal{H}(r, t)$ / $\mathcal{T}(h, r)$, there might be true triples that are part of the \ac{kg}.
If these false negatives are ranked higher than the current test triple $t^{+}$, the results might get distorted.
Therefore, the \emph{filtered} evaluation setting has been proposed \cite{Bordes2013}, in which the corrupted triples are filtered to exclude known true facts from the train and test set.
Thus, the rank does not decrease when ranking another true entity higher.

Moreover, we want to draw attention to the fact that the metrics can be further be distorted by \emph{unknown false negatives}, i.e., true triples that are contained in the set of corrupted triples but are not part of the \ac{kg} (and therefore cannot be filtered out).
Therefore, it is essential to investigate the predicted scores of a \ac{kgem} and not solely rely on the computed metrics.

Based upon these individual ranks, the following measures are frequently used to summarize the overall performance:

\textbf{\textbf{\Acl{mr}}} The \acf{mr} represents the average rank of the test triples, i.e.
\begin{equation}
    \text{MR} = \frac{1}{|\mathcal{K}_{test}|} \sum \limits_{t \in \mathcal{K}_{test}} rank(t)
\end{equation}

Smaller values indicate better performance.

\textbf{\textbf{\Acl{amr}}} Because the interpretation of the \ac{mr} depends on the number of available candidate triples, comparing \acp{mr} across different datasets (or inclusion of inverse triples) is difficult.
This is sometimes further exacerbated in the filtered setting because the number of candidates varies.
Therefore, with fewer candidates available, it becomes easier to achieve low ranks.
The \acf{amr}~\cite{berrendorf2020interpretable} compensates for this problem by comparing the mean rank against the expected mean rank under a model with random scores:
\begin{equation}
    \text{AMR} = \frac{MR}{\frac{1}{2}\sum \limits_{t \in \mathcal{K}_{test}} (\xi(t)+1)}
\end{equation}
where $\xi(t)$ denotes the number of candidate triples against which the true triple $t \in \mathcal{K}_{test}$ is ranked.
In the unfiltered setting we have $\xi(t) = |\mathcal{E}| - 1$ for all $t \in \mathcal{K}_{test}$.
Thereby, the measure also adjusts for chance, as a random scoring achieves an expected \acl{amr} of $1$.
The \ac{amr} has a fixed value range from 0 to 1, where smaller values (AMR $\ll$ 1) indicate better performance.

\textbf{\Acl{mrr}} The \acf{mrr} is defined as:
\begin{equation}
\text{MRR} = \frac{1}{|\mathcal{K}_{test}|} \sum\limits_{t \in \mathcal{K}_{test}} \frac{1}{rank(t)}
\end{equation}
where $\mathcal{K}_{test}$ is a set of test triples, i.e.\ the \ac{mrr} is the mean over reciprocal individual ranks.
However, the \ac{mrr} is flawed since the reciprocal rank is an ordinal scale and not an interval scale, i.e. computing the arithmetic mean is statistically incorrect~\cite{fuhr2018,stevens1946theory}.
Still, it is often used for early stopping since it is a smooth measure with stronger weight on small ranks, and less affected by outlier individual ranks than the mean rank.
The \ac{mrr} has a fixed value range from 0 to 1, where larger values indicate better performance.

\textbf{\textbf{Hits@K}} Hits@K denotes the ratio of the test triples that have been ranked among the top \textit{k} triples, i.e.,

\begin{equation}
    \text{Hits@k} = \frac{|\{t \in \mathcal{K}_{test} \mid rank(t) \leq k\}|}{|\mathcal{K}_{test}|}
\end{equation}

Larger values indicate better performance.

\textbf{Additional Metrics} Further metrics that might be relevant are the \ac{aucroc} and the \ac{aucpr}~\cite{nickel2015review}.
However, these metrics require the number of true positives, false positives, true negatives, and false negatives, which in most cases cannot be computed since the \acp{kg} are usually incomplete.

\section{Existing Benchmark Datasets}\label{sec:benchmark_datasets}

In this section, we describe the benchmark datasets that have been established to evaluate \acp{kgem}.
A summary is also given in Table~\ref{tab:existing_datasets}.

\textbf{FB15K} Freebase is a large cross-domain \ac{kg} consisting of around 1.2 billion triples and more than 80 million entities.
Bordes \textit{et al.}~\cite{Bordes2013} extracted a subset of Freebase, which is used as a benchmark dataset and named it FB15K.
It contains 14,951 entities, 1,345 relations, as well as more than half a million triples describing facts about movies, actors, awards, sports, and sports teams~\cite{Dettmers2018}.

\textbf{FB15K-237} FB15K has a test-leakage, i.e.\ a major part of the test triples (\begin{math}\sim\end{math}81\%) are inverses of triples contained in the training set: for most of the test triples of the form $(h, r, t)$, there exists a triple $(h, r', t)$ or $(t, r', h)$ in the training set.
Therefore, Toutanova and Chen~\cite{toutanova2015observed} constructed FB15K-237 in which inverse relations were removed~\cite{toutanova2015observed}.
FB15K-237 contains 14,541 entities and 237 relations.

\textbf{WN18} WordNet\footnote{https://wordnet.princeton.edu/} is a lexical knowledge base in which entities represent terms and are called \textit{synsets}.
Relations in WordNet represent conceptual-semantic and lexical relationships (e.g.\ hyponym).
Bordes \textit{et al.}~\cite{bordes2014semantic} extracted a subset of WordNet named WN18 that is frequently used to evaluate \acp{kgem}.
It contains 40,943 synsets and 18 relations.

\textbf{WN18RR} Similarly to FB15K, WN18 also has a test-leakage (of approximately 94\%)~\cite{toutanova2015observed}.
For instance, for most of the test triples of the form \textit{(h, hyponym, t)}, there exists a triple \textit{(t, hypernym, o)} in the training set.
Dettmers \textit{et al.}~\cite{Dettmers2018} have shown that a simple rule-based system can obtain results competitive to the state of the art results on WN18.
For this reason, they constructed WN18RR by removing inverse relations similarly to the procedure applied to FB15K.
WN18RR contains 40,943 entities and 11 relations.

\textbf{Kinships} The Kinships~\cite{denham1973detection} dataset describes relationships between members of the Australian tribe \textit{Alyawarra} and consists of 10,686 triples.
It contains 104 entities representing members of the tribe and 26 relationship types that represent kinship terms such as \textit{Adiadya} or \textit{Umbaidya}~\cite{bordes2014semantic}.

\textbf{Nations} The Nations~\cite{rummel1976dimensionality} dataset contains data about countries and their relationships with other countries.
Exemplary relations are \textit{economic\_aid} and \textit{accusation}~\cite{bordes2014semantic}.

\textbf{\acl{umls}} The \acf{umls} is an ontology that describes relationships between high-level concepts in the biomedical domain.
Examples of contained concepts are \textit{Cell}, \textit{Tissue}, and \textit{Disease}, and exemplary relations are \textit{part\_of} and \textit{exhibits}~\cite{bordes2014semantic,mccray2003upper}.

\textbf{YAGO3-10} \ac{yago} is a \ac{kg} containing facts that have been extracted from Wikipedia and aligned with WordNet in order to exploit the large amount of information contained in Wikipedia and the taxonomic information included in WordNet.
It contains general facts about public figures, geographical entities, movies, and further entities, and it has a taxonomy for those concepts.
YAGO3-10 is a subset of YAGO3~\cite{mahdisoltani2013yago3} (which is an extension of \ac{yago}) that contains entities associated with at least ten different relations.
In total, YAGO3-10 has 123,182 entities and 37 relations, and most of the triples describe attributes of persons such as citizenship, gender, and profession~\cite{Dettmers2018}.

\begin{table}[t]
\centering
\caption{Existing Benchmark Datasets.}\label{tab:existing_datasets}
\begin{tabular}{lrrr}
\toprule
Dataset & Triples & Entities & Relations \\
\midrule
FB15K & 592,213 & 14.951 & 1,345\\
FB15K-237 & 272,115 & 14,541 & 237\\
WN18 & 151,442 & 40,943 & 18\\
WN18RR & 93,003 & 40,943 & 11\\
Kinships & 10,686 & 104 & 26\\
Nations & 11,191 & 14 & 56\\
\ac{umls} & 893,025 & 135 & 49\\
YAGO3-10 & 1,079,40 & 132,182 & 37\\
\bottomrule
\end{tabular}
\end{table}

\section{Reproducibility Studies}\label{reproducibility_study}

The goal of the reproducibility studies was to investigate whether it is possible to replicate experiments based on the information provided in each model's accompanying paper.
If specific information was missing, such as the number of training epochs, we tried to find this information in the accompanying source code if it was accessible.
For our study, we focused on the two most frequently used benchmark datasets, FB15K and WN18, as well as their respective subsets FB15K-237 and WN18RR.
Table~\ref{fig:overview_model_dataset} (Appendix \pageref{fig:overview_model_dataset}) illustrates for which models results were reported (in the accompanying publications) for the considered datasets.
A checkmark denotes that results were reported, and green background indicates that the entire experimental setup for the corresponding dataset was described.
Results have not been reported for every model for every dataset because some of the benchmark datasets were created after the models were published.
Therefore, these models have been excluded from our reproducibility study.

\textbf{Experimental Setup} For each \ac{kgem}, we applied identical training and evaluation settings as described in their concomitant papers.
We ran each experiment four times with random seeds to measure the variance in the obtained results.
We evaluated the models based on the ranking metrics \ac{mr}, \ac{amr}, \ac{mrr}, and Hits@K.
As discussed in~\cite{sun2019re,berrendorf2020interpretable}, the exact computation of ranks differs across different codebases, and can lead to significant differences~\cite{sun2019re}.
We follow the nomenclature of Berrendorf \textit{et al.}~\cite{berrendorf2020interpretable}, and report scores based on the optimistic, pessimistic, and realistic rank definitions.

Tables~\ref{tab:rps_fb15k}-\ref{tab:rps_wn18rr} (Appendix \pageref{tab:rps_fb15k}-\pageref{tab:rps_wn18rr}) represent the results for FB15K, FB15K-237, WN18, and WN18RR where experiments highlighted in black were reproducible, in blue soft-reproducible experiments (i.e., could be reproduced by a margin $\leq 5\%$), and experiments highlighted in orange could not be reproduced.
In the following, we discuss the observations that we made during our experiments.

\subsection{Reproductions Requiring Alternate Hyper-Parameters}\label{different_values}

One of the observations we made is that for some experiments, results could only be reproduced with a different set of hyper-parameter values.
For instance, the results for TransE could only be reproduced by adapting the batch size and the number of training epochs.
We trained TransE on WN18 for 4000 epochs compared to a reported number of 1000 epochs in order to obtain comparable results.
Furthermore, for RotatE on FB15K and WN18, we received better results when adapting the learning rate.
The reason for these differences might be explained by the implementation details of the underlying frameworks which have been used to train the models.
Authors of early \acp{kgem} often implemented their training algorithms themselves or used frameworks that were popular at the respective time but are not used anymore.
Therefore, differences between the former and current frameworks may require an adaption of the hyper-parameter values.
Even within the same framework, bug fixes or optimizations of the framework can lead to different results based on the used version.
Our benchmarking study highlights that with adapted settings, results can be reproduced and even improved.

\subsection{Unreported Hyper-parameters Impedes Reproduction}

Some experiments did not report the full experimental setup impeding the reproduction of results.
For example, the embeddings in the ConvKB experiments have been pre-trained based on TransE.
However, the batch size for training TransE has not been reported, which can significantly affect the results, as previously discussed.
Furthermore, we obtained a high deviation for the reported results for HolE on FB15K.
The apparent reason is that we could not find the hyper-parameter setting for FB15K, such that we used the same setting as for WN18, which we found in the accompanying implementation.

\subsection{Two Perspectives: Publication versus Implementation}

While preparing our experiments, we observed that for some experiments, essential aspects, which are part of the released source code, have not been discussed in the paper.
For instance, in the publication describing ConvE, it is not mentioned that inverse triples have been added to the \acp{kg} in a pre-processing step.
This step seems to be essential to reproduce the results.
A second example is SimplE, for which the predicted scores have been clamped to the range of $[-20,20]$.
This step was not mentioned in the publication, but it can have a significant effect when the model is evaluated based on an optimistic ranking approach, which is the case for SimplE.

\subsection{Lack of Official Implementations Impedes Reproduction}

During our experiments, we observed that for DistMult and TransD, we were able to reproduce the results on WN18, but not on FB15K.
A reason might be differences in the implementation details of the frameworks used to train and evaluate the models.
For example, the initialization of the embeddings or the normalization of the loss values could have an impact on the performance.
Since there exists no official implementation (see Table~\ref{fig:overview_model_dataset} in Appendix \pageref{fig:overview_model_dataset}) for DistMult and TransD, it is not possible to check the above-mentioned aspects.
Furthermore, we were not able to reproduce the results for TransH for which also no official implementation is available.
There exist reference implementations\footnote{https://github.com/thunlp/OpenKE}, which slightly differ from the model initially proposed.

\subsection{Reproducibility is Dependent on The Ranking Approach}

As discussed in \cite{sun2019re,berrendorf2020interpretable}, the ranking metrics have been implemented differently by various authors.
In our experiments, we report results based on three common implementations of the ranking metrics: i.) realistic, ii.) optimistic and iii.) pessimistic ranking (Section~\ref{sec:evaluation_of_models}).
If a model predicts the same score for many triples, there will be a large discrepancy between the three ranking approaches.
We could observe such a discrepancy for SimplE for which the results on FB15K (Table~\ref{tab:rps_fb15k} in Appendix \pageref{tab:rps_fb15k}) and WN18 (Table~\ref{tab:rps_wn18} in Appendix \pageref{tab:rps_wn18}) were almost 0\% based on the realistic ranking approach, but were much higher based on the optimistic ranking approach.
Similar observations for other \ac{kgem} have been made in \cite{sun2019re}.

\section{Benchmarking}\label{benchmarking_study}

In our benchmarking studies, we evaluated a large set of different combinations of interaction models, training approaches, loss functions, and the effect of explicitly modeling inverse relations.
Additionally, we evaluated how well the interaction models can model symmetry, anti-symmetry and composition patterns (Appendix~\ref{relation_patterns_definition}).
In particular, we investigated 21 interaction models, two training approaches, and five loss functions on four datasets.
We refer to a speciﬁc combination of interaction model, training approach, loss function, and whether inverse relations are explicitly modeled as a \emph{configuration}, e.g., RotatE + \ac{lcwa} + \ac{spl} + inverse relations.
We do \emph{not} refer to different hyper-parameter values such as batch size or learning rate when we use the term configuration.
For each configuration, we used random search to perform the hyper-parameter optimizations over all other hyper-parameters and applied early stopping on the validation set.
Each \acl{hpo} experiment lasted for a maximum of 24 hours or 100 iterations, in which new hyper-parameters have been sampled in each iteration.
Overall, we performed individual hyper-parameter optimizations for more than 1,000 configurations. 
We retrain the model with the best hyper-parameter setting and report evaluation results on the test set.

Before presenting our results, we provide an overview of the experimental setup, comprising the investigated interaction models, training approaches, loss functions, negative samplers, and datasets.
We used the \ac{slcwa} and \ac{lcwa} as training approaches.
For the \ac{slcwa} we applied a \textit{1:k}-Scoring as usually done throughout the literature~\cite{Bordes2013,Trouillon2016},  where $k$ denotes the number of negative examples for each positive.
For the \ac{lcwa}, we applied a \textit{1:N}-Scoring, i.e., we sample each batch against all negatives examples as typically done for training with the \ac{lcwa}~\cite{Dettmers2018}.
Table~\ref{tab:hpo_params} (Appendix \pageref{tab:hpo_params}) shows the hyper-parameter ranges for the \ac{slcwa} and the \Ac{lcwa} assumptions.

\textbf{Datasets} We performed experiments on the following four datasets: WN18RR, FB15K-237, Kinships and YAGO3-10.
We selected WN18RR and FB15K-237 since they are widely applied benchmarking datasets.
We chose Kinships and YAGO3-10 to investigate the performance of \acp{kgem} on a small and a larger dataset.

\textbf{Interaction Models} We investigated all interaction models described in Section~\ref{interaction_model}.
Because of our vast experimental setup and the size of YAGO3-10, we restricted the number of interaction models on YAGO3-10 as otherwise, the computational effort would be prohibitive.
Based on their variety of model types as described in Section~\ref{interaction_model}, we selected the following interaction models: ComplEx, ConvKB, DistMult, ERMLP, HolE, MuRE, QuatE, RESCAL, RotatE, \ac{se}, TransD, and TransE.

\textbf{Training Approaches} We trained the interaction models based on the \ac{slcwa} (Section~\ref{ta_owa}) and the \ac{lcwa} (Section~\ref{ta_lcwa}) training approaches.
Due of the extent of our benchmarking study and the fact that YAGO3-10 contains more than 132,000 entities, which makes the training based on the \ac{lcwa} with 1-n scoring expensive, we restricted the training approach to the \ac{slcwa} for YAGO3-10.

\textbf{Loss Functions} We investigated \ac{mrl}, \ac{bcel}, \ac{spl}, \ac{nssal}, and \ac{cel} since they represent the variety of types described in Section~\ref{loss_functions} and because they have been previously shown to yield good results.
\ac{mrl} has not been historically used in the 1-N scoring setting likely due to the fact that in 1-N scoring, the number of positive and negative scores in each batch is not known in advance and dynamic.
Thus, the number of possible pairs varies as well ranging from $N-1$ to $(N/2)^2$ for each $(h, r)$ combination.
The accompanying variance in memory requirements for each batch thus poses practical challenges. Therefore, we did not use the \ac{mrl} in combination with the 1-N scoring setting.

\textbf{Negative Sampler} When using the \ac{slcwa}, we generated negative samples with \ac{uns}. When training with the \ac{lcwa} and 1-N scoring, no explicit negative sampling was required.

\textbf{Early Stopping} We evaluated each model every 50 epochs and performed early stopping with a patience of 100 epochs on all datasets except for YAGO3-10.
There, considering the larger number of triples seen in each epoch we evaluated each model every 10 epochs and performed early stopping with a patience of 50 epochs.

Below, we describe the results of our benchmarking study.
In the four following subsections, we summarize the results for each dataset (i.e.,\ Kinships, WN18RR, FB15K-237, YAGO3-10) along with a discussion of the effect of the models' individual components (i.e., training approaches, loss functions, the explicit modeling of inverse relations) and optimizers on the performance.
Finally, we compare the model complexity versus performance.
In the appendix, we provide further results.
In particular, we provide for each model the results of all tested combinations of interaction model, training approach, and loss function.

\subsection{Results on the Kinships Dataset}

\begin{figure*}[tb]
\centering
\includegraphics[width=1.0\linewidth]{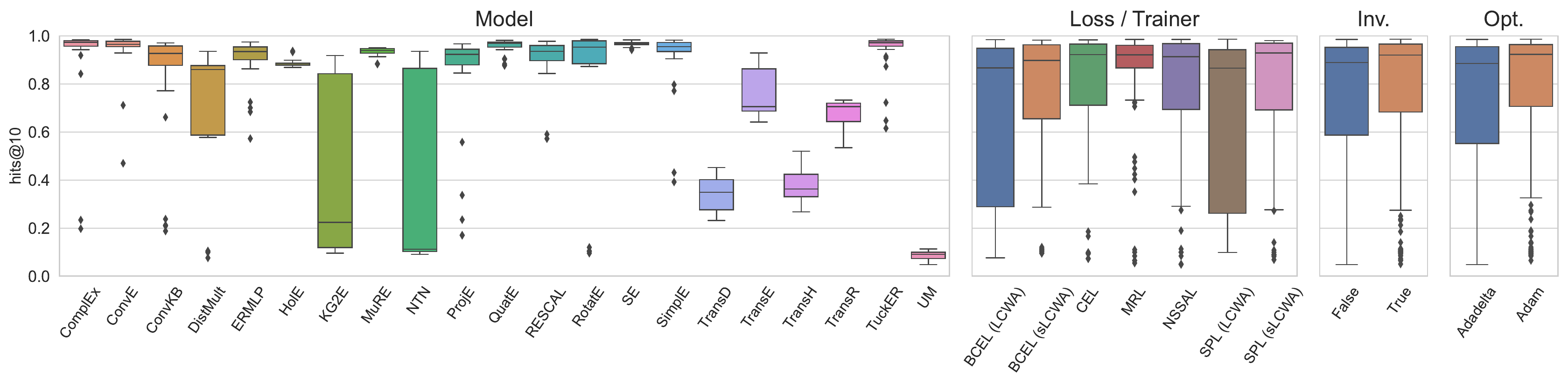}
\caption{Overall hits@10 results for Kinships where box-plots summarize the best results across different configurations, i.e., combinations of interaction models, training approaches, loss functions, and the explicit usage of inverse relations.}
\label{fig:summary_kinhsips}
\end{figure*}

Investigating the model performances on Kinhsips is interesting because it is a comparatively small \ac{kg} and thus permits for each configuration a large number of \ac{hpo} iterations for all interaction models.
Figure~\ref{fig:summary_kinhsips} provides a general overview of the results, i.e., performance of the interaction models, loss functions, training approach, the effect of modeling inverse relations, and the effect of the optimizers.
Overall, it can be observed that for most interaction models, several well-performing configurations can be determined.
However, some interaction models heavily depend on specific configurations such as KG2E and QuatE.
Although link prediction on Kinships seems to be relatively easy, there are several translational distance-based interaction models that perform relatively poor (i.e., TransD, TransE, TransH, TransR, and \ac{um}).
The poor performance of \ac{um} is not surprising considering that it omits the multi-relational information of the data.
Finally, the results illustrate that Adam outperforms Adadelta (in many cases with high margin).
Therefore, we decided to progress only with Adam as optimizer for the remaining datasets in order to reduce the computational costs.

\begin{figure}[ht]
\centering
\includegraphics[width=0.5\textwidth]{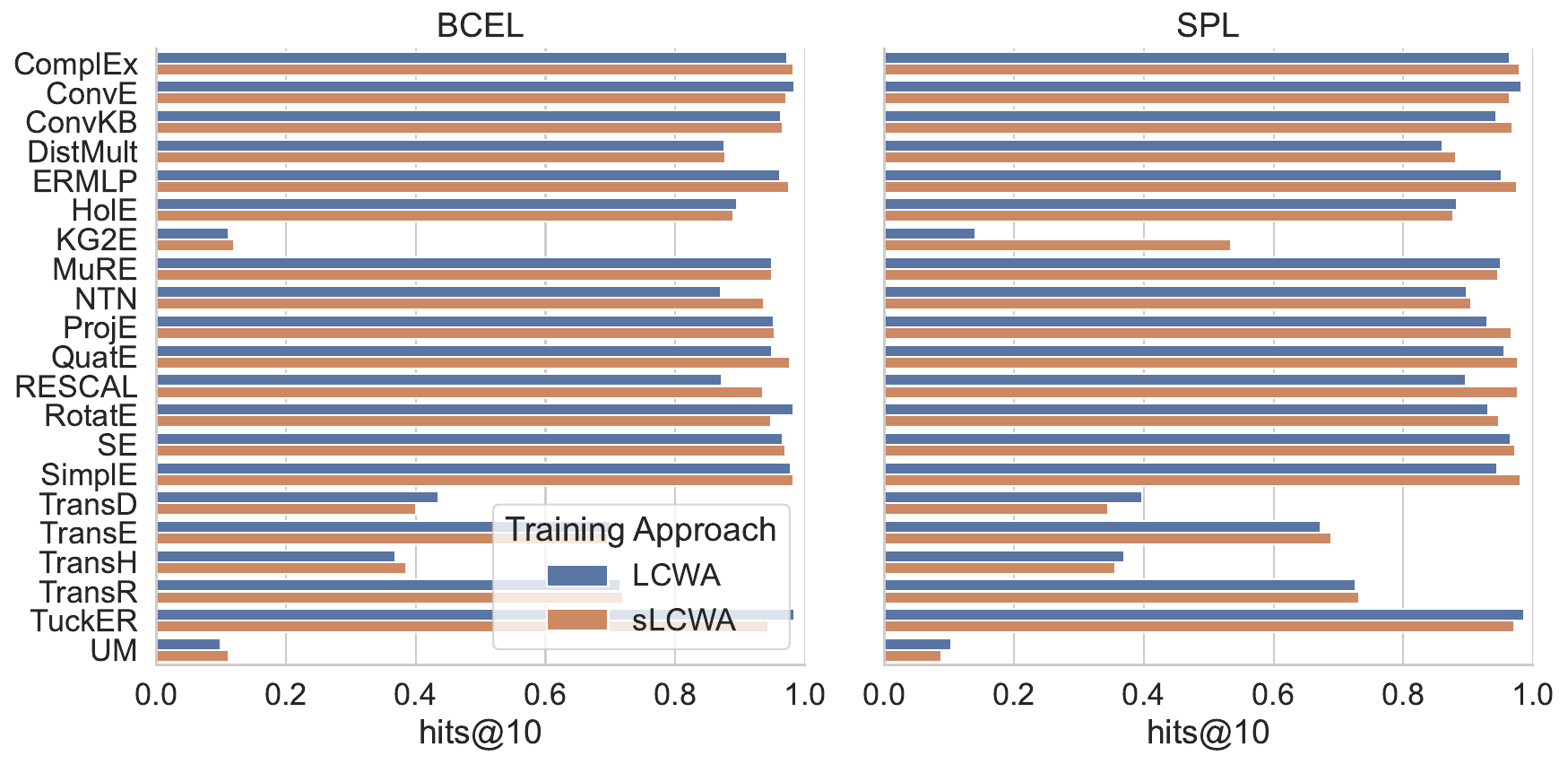}
\caption{Impact of training approach on the performance for a fixed interaction model and loss function for the Kinships dataset based on Adam.}
\label{fig:kinships_ta_model_loss_fct}
\end{figure}

\textbf{Impact of the Training approach}  Figure~\ref{fig:kinships_ta_model_loss_fct} depicts the effect of the training approaches.
We focus only on the \ac{bcel} and the \ac{spl} (which is equivalent to \ac{bcel}, but numerical more stable, see Section \ref{pwlf}) since they have been trained with both training approaches.
It can be observed that some interaction models such as MuRE perform equally well on both training approaches on Kinships whereas others such as RESCAL benefit from one of the training approaches (in this case from the \ac{slcwa}).

\textbf{Impact of the Loss Function}

Figure~\ref{fig:summary_kinhsips} highlights that
selecting the appropriate loss function is crucial also for relatively small dataset such as Kinships.
Although all five loss functions achieve high performance, all except the \ac{mrl} exhibit high variance.
Comparing an interaction model that has been trained with the \ac{mrl} with an interaction model that has been trained with a different loss function can lead to misleading conclusions since finding a suitable configuration for the loss functions except for the \ac{mrl} is more difficult.

\textbf{Impact of Explicitly Modeling Inverse Relations}

\begin{figure}[ht]
\centering
\includegraphics[width=0.4\textwidth]{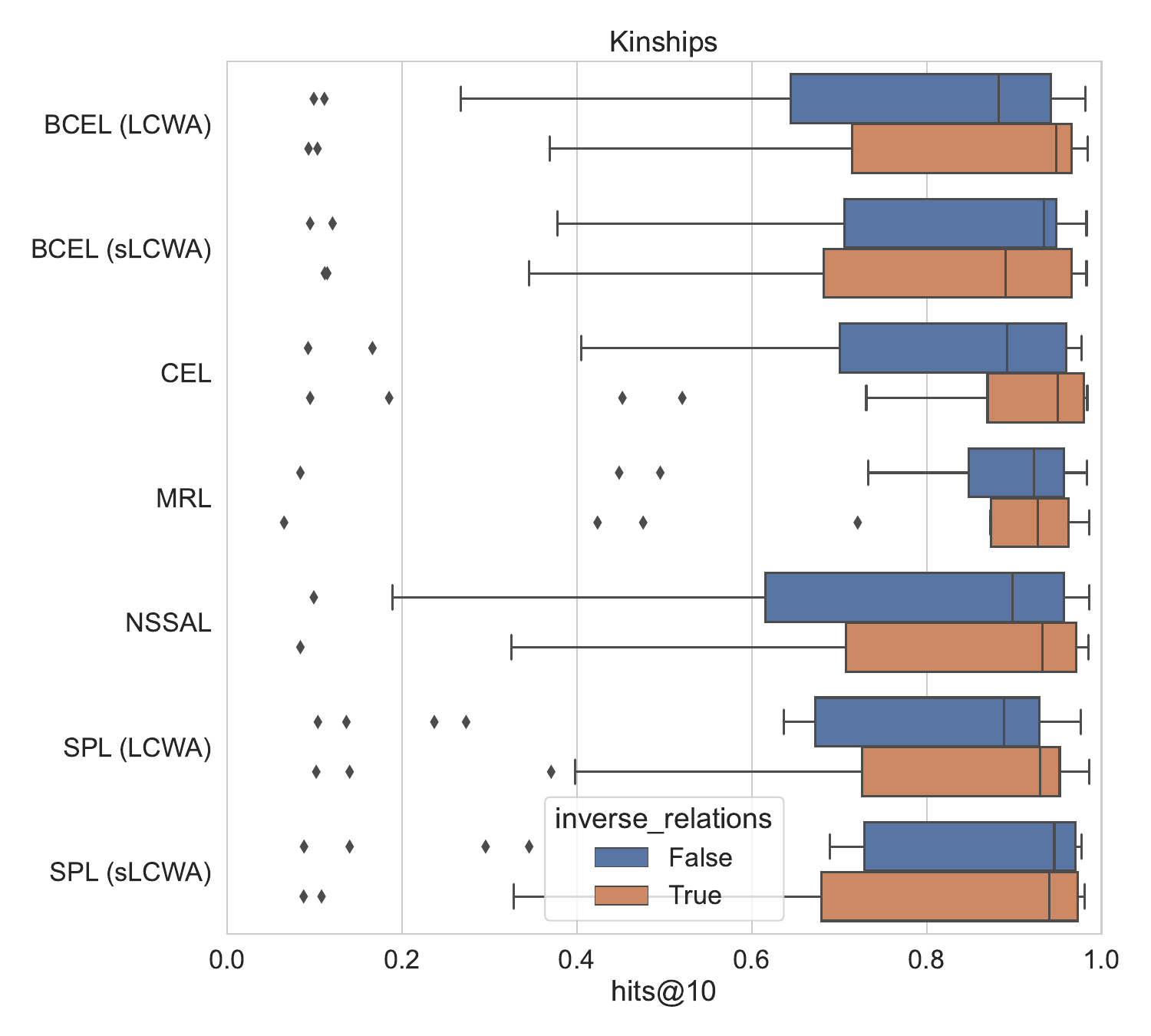}
\caption{Impact of explicitly modeling inverse relations on the performance for a fixed loss function for the Kinships dataset.}
\label{fig:kinhsips_inverse_loss}
\end{figure}

Figures~\ref{fig:summary_kinhsips} and~\ref{fig:kinhsips_inverse_loss} present the effect of explicitly modeling inverse relations.
Overall, explicitly modeling inverse relations results in less variance across the investigated configurations (Figure~\ref{fig:summary_kinhsips}).
Further investigating the effect of modeling of inverse relations on the different loss functions and training approaches (Figure~\ref{fig:kinhsips_inverse_loss}), it can be observed that in general, the \ac{lcwa} benefits from explicit usage of inverse relations in terms of robustness.
This is to be expected since, in the \ac{lcwa}, the model only learns to perform tail predictions, and without explicitly modeling inverse relations, the model might have difficulties in correctly predicting head entities.
However, when explicitly modeling inverse relations, the head predictions are obtained by predicting the tail entities of the corresponding inverse triples (see Section~\ref{sec:inverse_relations})

Interestingly, \ac{mrl} and \ac{nssal}-based configurations, which are both only trained with the \ac{slcwa} (i.e., the model already learns to perform head and tail predictions) are more robust when trained with inverse relations.
Therefore, depending on the dataset, it might be helpful to employ inverse relation for these loss functions even though they might be trained with \ac{slcwa}.

\textbf{Model Complexity versus Performance} Figure~\ref{fig:model_size_performance} (Appendix \pageref{fig:model_size_performance}) plots the model size against the obtained performance.
The results highlight that there is no strong correlation between model size and performance, i.e.,\ models with a small number of parameters can perform equally well as large models on the Kinships data set.
The skyline comprises small \ac{um} models, some intermediate HolE and ProjE models, and larger RotatE and TuckER models.
A full list is provided in Table~\ref{tab:skyline_kinships_model_bytes} in Appendix~\pageref{tab:skyline_kinships_model_bytes}.

\subsection{Results on the WN18RR Dataset}\label{benchmarking_results_wn18rr}

\begin{figure*}[tb]
\centering
\includegraphics[width=1.0\linewidth]{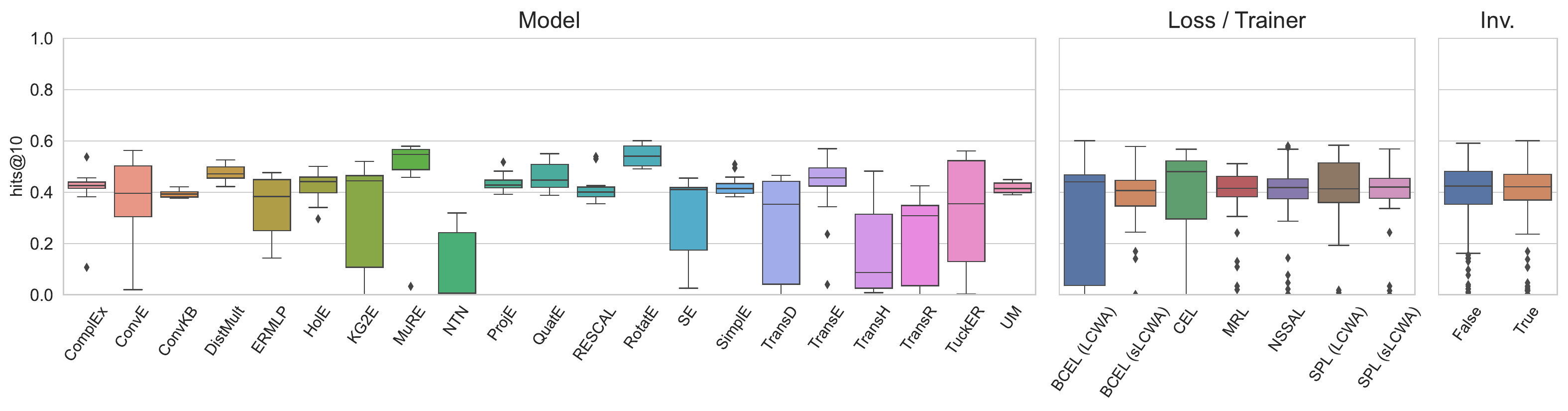}
\caption{Overall hits@10 results for WN18RR where box-plots summarize the results across different combinations of interaction models, training approaches, loss functions, and the explicit usage of inverse relations.}
\label{fig:summary_wn18rr}
\end{figure*}

Figure~\ref{fig:summary_wn18rr} depicts the overall results over WN18RR.
A detailed overview of all configurations can be found in Figure~\ref{fig:all_configs_wn18rr_adam} in Appendix~\pageref{fig:all_configs_wn18rr_adam}.
The results highlight that there are several combinations of interaction models, loss functions, and training approaches that obtain hits@10 results that are competitive with state-of-the-art results\footnote{https://paperswithcode.com/sota/link-prediction-on-wn18rr}.
In particular, ComplEx (53.74\%), ConvE (56.33\% compared to 52.00\% in the original paper~\cite{Dettmers2018} ), DistMult (52.62\%), MuRE (57.90\% compared to 55.50\% in the original paper~\cite{DBLP:conf/nips/BalazevicAH19}), KG2E (52.30\%),  ProjE (51,73\%), TransE (56.98\%), RESCAL (53.92\%), RotatE (60.09\% compared to 56.61\% in the original paper~\cite{Sun2019}), SimplE (50.89\%), and TuckER (56.09\% compared to 52.6\% in the original paper~\cite{balavzevic2019tucker}) obtained high performance. 
Especially the result obtained by TransE is impressive since with a suitable configuration, it beats most of the published state-of-the-art results.
The results highlight that determining an appropriate combination of interaction model, loss function, training approach, and the decision to explicitly modeling inverse relation is fundamental since many interaction models such as ConvE and KG2E reveal a high variance across different configurations. 
The results for ComplEx and RESCAL further underpin this observation. They reveal competitive results with very specialized configurations that represent outliers.
Another interesting observation is the performance of \ac{um}, which does not model relations, but can still compete with some of the other interaction models on WN18RR.
This observation might indicate that the relational patterns in WN18RR are not too diverse across relations.

\begin{figure}[ht]
\centering
\includegraphics[width=0.5\textwidth]{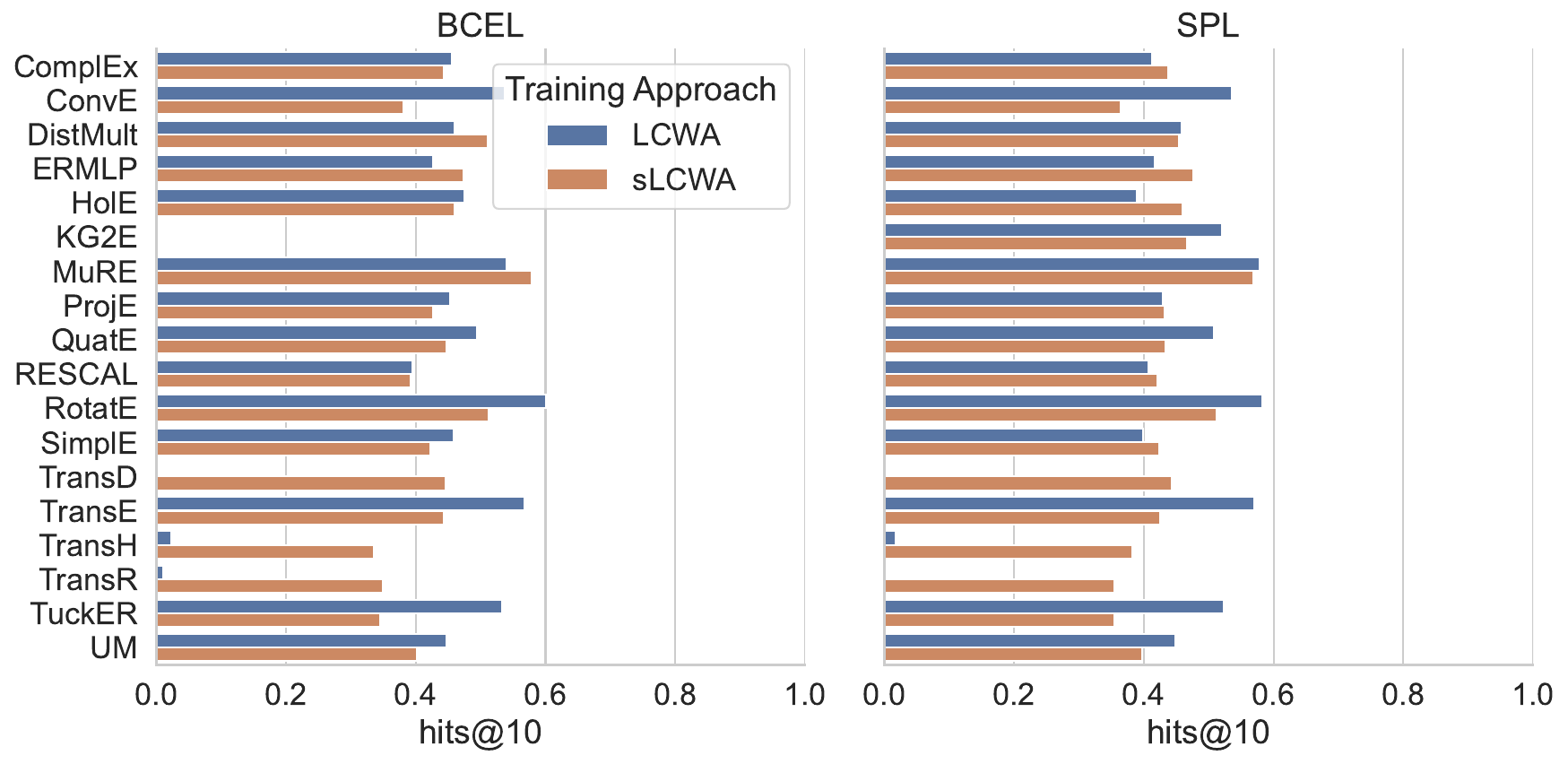}
\caption{Impact of training approach on the performance for a fixed interaction model and loss function for the WN18RR dataset.}
\label{fig:wn18rr_ta_model_loss_fct}
\end{figure}

\textbf{Impact of the Training Approach} Figures~\ref{fig:summary_wn18rr} and~\ref{fig:wn18rr_ta_model_loss_fct} depict the impact of the training approach.
Again, we focus only on \ac{bcel} and \ac{spl} since they have been trained under both the \ac{slcwa} and \ac{lcwa}.
The figures highlight that for both realizations of the \acl{bcel}, the \ac{lcwa} achieves higher maximum performance, but at the same time, it reveals a larger variance on both loss functions.
Consequently, it may be more difficult to find configurations that obtain high performance.
The overall lower variance of \ac{spl} can be explained by the fact that it is numerically more stable than the \ac{bcel}.

Figure~\ref{fig:wn18rr_ta_model_loss_fct} shows the impact of the training approaches for fixed interaction models and used loss functions.
The results indicate that for some combinations of interaction models and loss functions, the training approach's choice has a significant impact on the results.
For instance, ConvE, RotatE, TransE and TuckER reveal stronger performance when trained with the \ac{lcwa} whereas TransH suffer under the \ac{lcwa}.

\textbf{Impact of the Loss Function} Figure~\ref{fig:summary_wn18rr} depicts the performance of the different loss functions.
State-of-the-art results for WN18RR are currently between 50\% and 60\%, and for each loss function, at least 50\% could be achieved~(Figure \ref{fig:all_configs_wn18rr_adam} in Appendix \pageref{fig:all_configs_wn18rr_adam}).
However, the \ac{mrl} is comparably less competitive than the other loss functions.
This observation is especially important considering that early \acp{kgem} have often been trained with the \ac{mrl}.
The results highlight that there is a trade-off between highest performance and robustness, i.e.,\ \ac{spl} and \ac{bcel} achieve the highest performance (when trained under the \ac{lcwa}), but also have high variance across different configurations (especially \ac{bcel} + \ac{lcwa}).

\begin{figure*}[!t]
\centering
\includegraphics[width=0.8\textwidth]{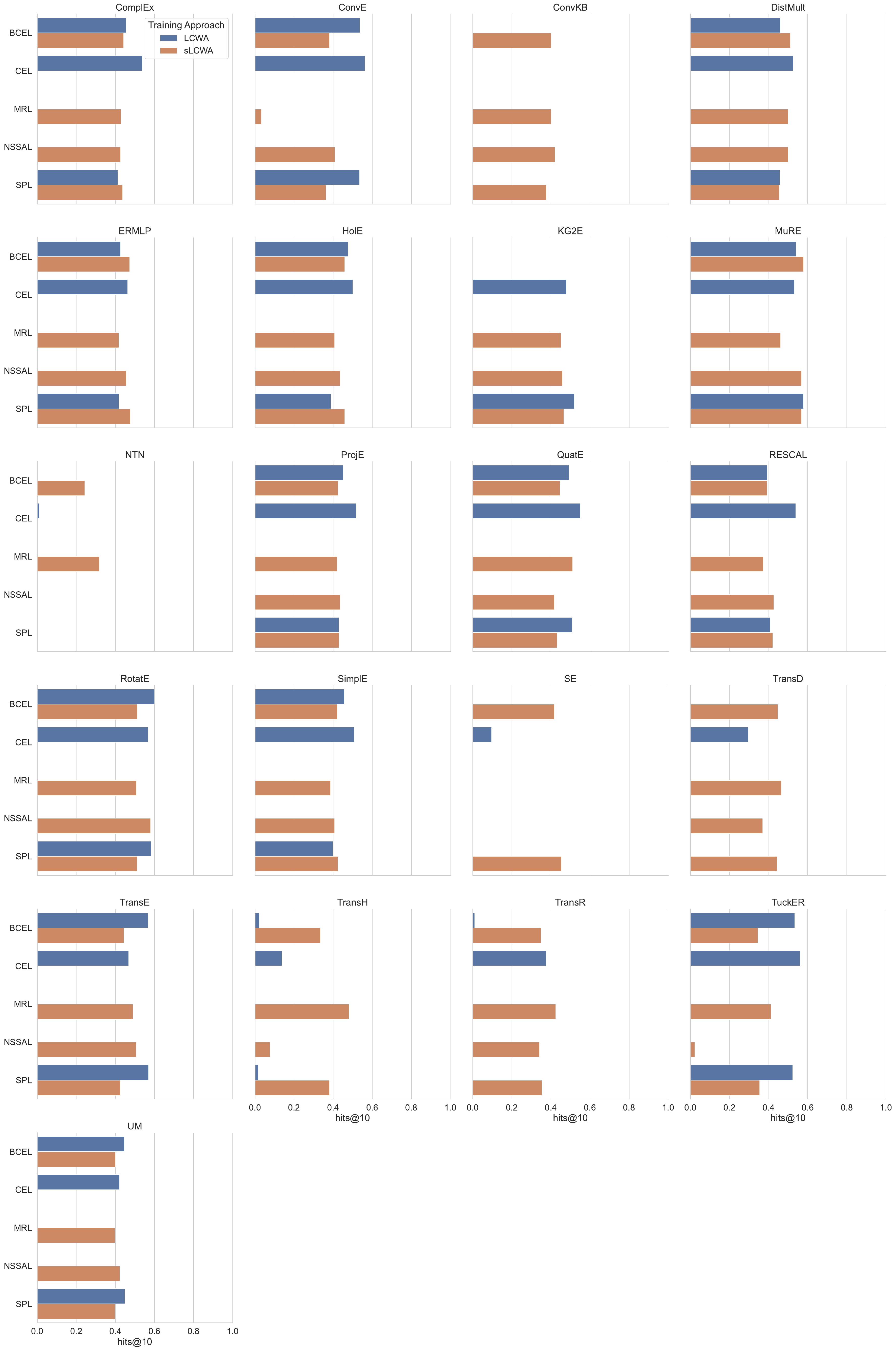}
\caption{Impact of the training approach on the performance for a fixed interaction model and loss function for the WN18RR dataset (results represent for each setting the best-performing configuration). \textbf{\ac{bcel}} refers to the binary cross entropy loss, \textbf{\ac{cel}} to the cross entropy loss, \textbf{\ac{mrl}} to the margin ranking loss, \textbf{\ac{nssal}} refers to the negative sampling self-adversarial loss, \textbf{\ac{spl}} to the softplus loss, \textbf{\ac{lcwa}} to the local closed world assumption training approach and \textbf{\ac{slcwa}} to the stochastic local closed world assumption training approach.}
\label{fig:wn18rr_loss_fct_each_model}
\end{figure*}

Figure~\ref{fig:wn18rr_loss_fct_each_model} (Appendix \pageref{fig:wn18rr_loss_fct_each_model}) reveals that some interaction
models can obtain a further performance boost when configured with specific loss functions. 
For instance, the performance of ComplEx, ProjE and RESCAL can be increased by a significant margin when composed together with the \ac{cel}.

\begin{figure}[ht]
\centering
\includegraphics[width=0.4\textwidth]{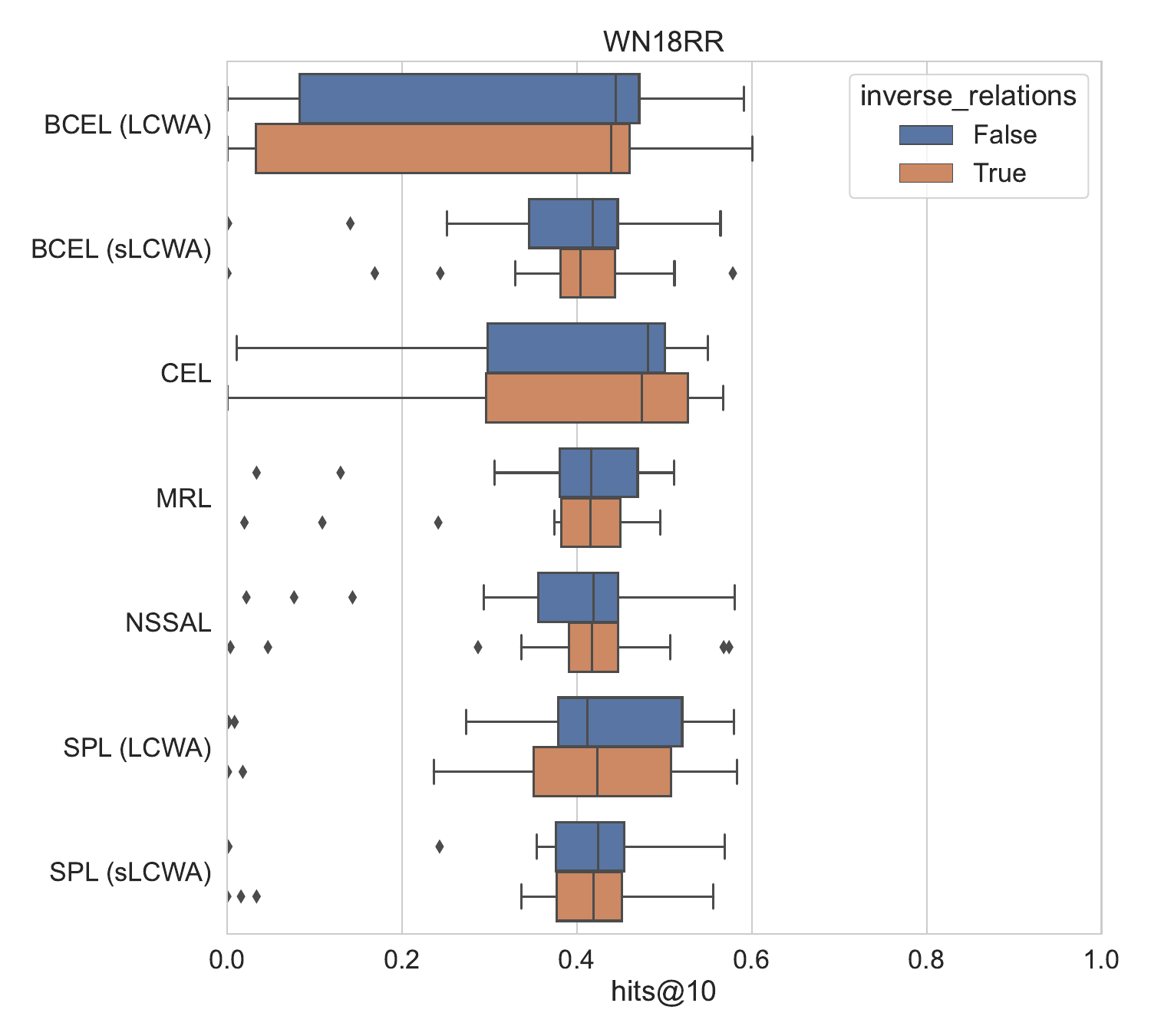}
\caption{Impact of explicitly modeling inverse relations on the performance for a fixed loss function for the WN18RR dataset.}
\label{fig:wn18rr_inverse_loss}
\end{figure}

\textbf{Impact of Explicitly Modeling Inverse Relations} Figure~\ref{fig:wn18rr_inverse_loss} illustrates that it is easier to find a strong performing \ac{slcwa}-configurations when trained without inverse relations.
Surprising is that for \ac{lcwa} based configurations, the interaction models are still competitive when trained without inverse relations.
This observation is surprising because \acp{kgem} that are configured with the \ac{lcwa} and without inverse relations are not explicitly trained to predict the head entities of triples.

\textbf{Model Complexity vs. Performance} Figure~\ref{fig:model_size_performance} (Appendix \pageref{fig:model_size_performance}) highlights that there is no significant correlation between model size and performance.
Instead, the results show that with an appropriate configuration, the model complexity can be significantly reduced (Table~\ref{tab:skyline_wn18rr_model_bytes} in Appendix \pageref{tab:skyline_wn18rr_model_bytes}). 
For instance, for RotatE, several high-performing configurations have been found (Figure~\ref{fig:all_configs_wn18rr_adam} in the Appendix \pageref{fig:all_configs_wn18rr_adam}), and the second-best configuration achieved a hits@10 value of 58.33\% while trained with an embedding dimension of \emph{64} (in the complex space).
This is especially interesting considering that RotatE originally obtained a performance of 57.1\% hits@10~\cite{Sun2019} with an embedding dimension of 500 (in the complex space) using the \ac{slcwa} as training approach and the NSSAL as loss function \footnote{\url{https://github.com/DeepGraphLearning/KnowledgeGraphEmbedding}}.
By changing the training approach and the loss function, the embedding dimension could be reduced significantly while getting at the same time an improvement in the hits@10 score.

\subsection{Results on the FB15K-237 Dataset}

\begin{figure*}[tb]
\centering
\includegraphics[width=1.0\linewidth]{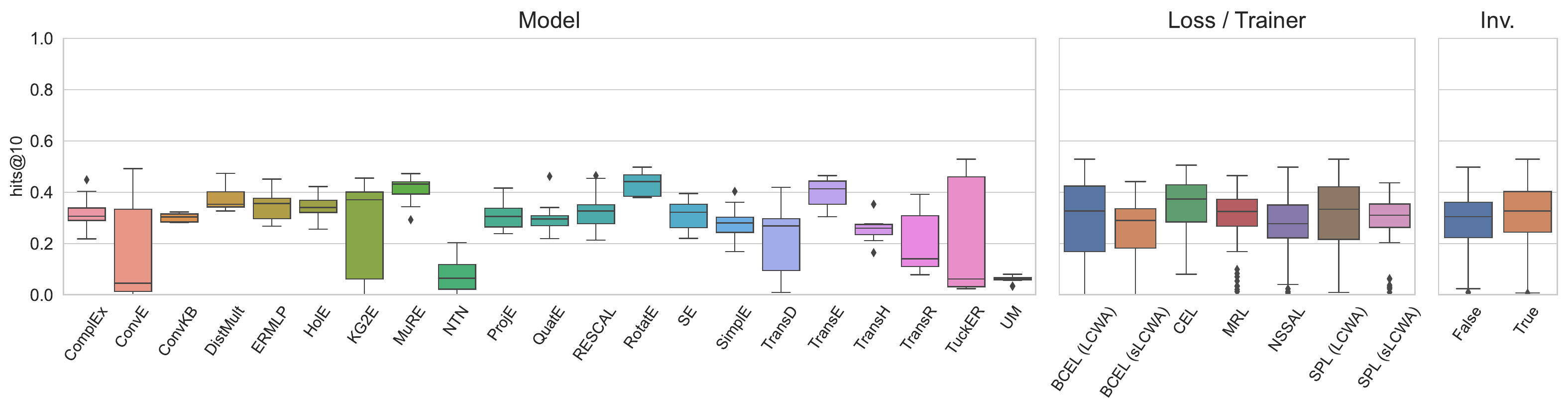}
\caption{Overall hits@10 results for FB15K-237 where box-plots summarize the results across different combinations of interaction models, training approaches, loss functions, and the explicit usage of inverse relations.}
\label{fig:summary_fb15k237}
\end{figure*}

Figure~\ref{fig:summary_fb15k237} provides an overall overview of the results obtained on FB15K-237.
For the results for each individual configuration, we refer to Figure~\ref{fig:all_configs_fb15k237_adam} in Appendix \pageref{fig:all_configs_fb15k237_adam}.
We can observe that TuckER outperforms the other interaction models followed by RotatE. 
DistMult again obtains surprisingly good results (Table ~\ref{fig:all_configs_fb15k237_adam} in Appendix~\pageref{fig:all_configs_fb15k237_adam}) considering that the interaction model enforces symmetric relations. 
The results illustrate again that choosing a suitable composition is essential for the performance of an interaction model.
For instance, TuckER and QuatE perform well only with dedicated compositions.
A further example is DistMult, which again obtains surprisingly good results (Table ~\ref{fig:all_configs_fb15k237_adam} in Appendix \pageref{fig:all_configs_fb15k237_adam}) considering that the interaction model enforces symmetric relations. 
DistMult, however, achieves a strong performance only when composed with the \ac{lcwa} and the \ac{cel} (Table ~\ref{best_models_fb15k237} in Appendix \pageref{best_models_fb15k237}), highlighting that a simple interaction model can obtain strong performance when composed beneficially.

\begin{figure}[ht]
\centering
\includegraphics[width=0.5\textwidth]{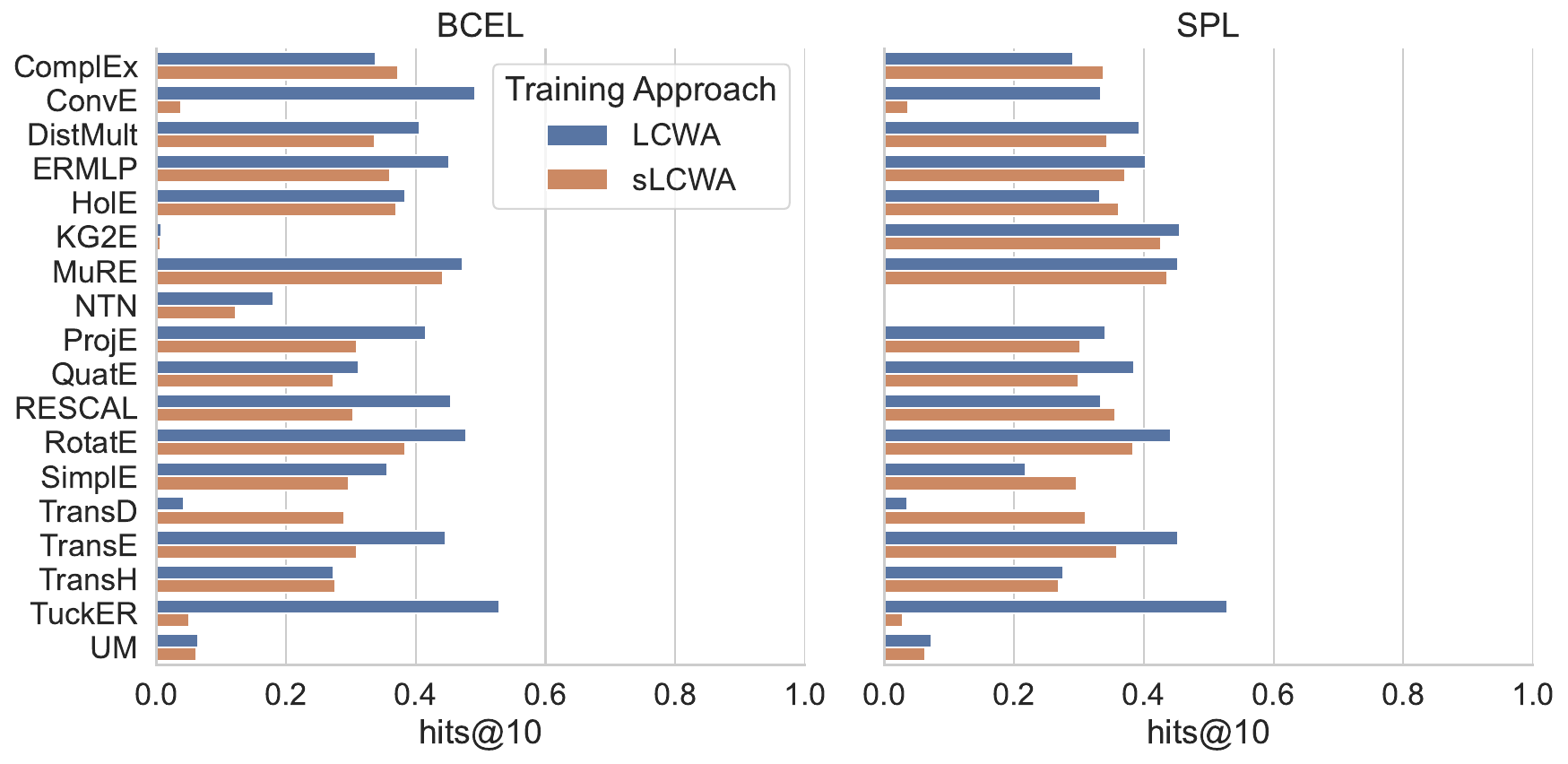}
\caption{Impact of training approach on the performance for a fixed interaction model and loss function for the FB15K-237 dataset.}
\label{fig:fb15k237_ta_model_loss_fct}
\end{figure}

\begin{figure*}[!t]
\centering
\includegraphics[width=0.8\textwidth]{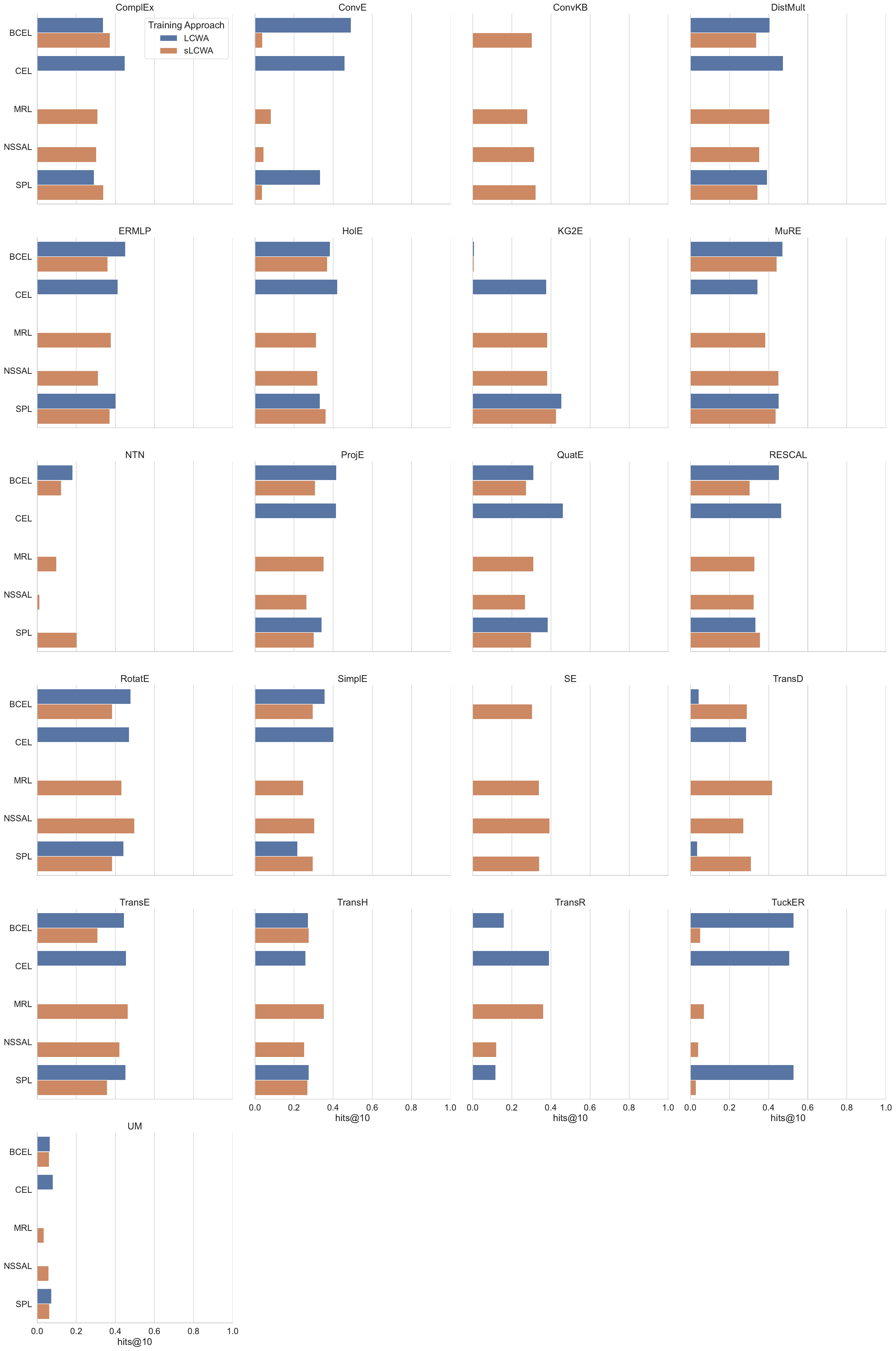}
\caption{Impact of the training approach on the performance for a fixed interaction model and loss function for the FB15K-237 dataset (results represent for each setting the best-performing configuration). \textbf{\ac{bcel}} refers to the binary cross entropy loss, \textbf{\ac{cel}} to the cross entropy loss, \textbf{\ac{mrl}} to the margin ranking loss, \textbf{\ac{nssal}} refers to the negative sampling self-adversarial loss, \textbf{\ac{spl}} to the softplus loss, \textbf{\ac{lcwa}} to the local closed world assumption training approach and \textbf{\ac{slcwa}} to the stochastic local closed world assumption training approach.}
\label{fig:fb15k237loss_fct_each_model}
\end{figure*}

\textbf{Impact of the Training Approach} Figure~\ref{fig:summary_fb15k237} shows that for both, \ac{bcel} and \ac{spl}, the \ac{lcwa} obtains significantly higher results, but they express a high variance at the same time.
Figures~\ref{fig:fb15k237_ta_model_loss_fct} and \ref{fig:fb15k237loss_fct_each_model} (Appendix \pageref{fig:fb15k237loss_fct_each_model}) illustrate that some interaction models are extremely sensitive to the choice of the training approaches.
For instance, it can be observed that RotatE, TransE, and TuckER suffer when trained together with the \ac{slcwa} for both loss functions.
Table~\ref{best_models_fb15k237} (Appendix \pageref{best_models_fb15k237}) shows that most of the interaction models obtain their best performance on FB15K-237 when trained together with the \ac{lcwa}.

\textbf{Impact of the Loss Function} Figure~\ref{fig:summary_fb15k237} illustrates that the \ac{bcel} and \ac{spl} outperform the other loss functions, but they also exhibit higher variance.
Figure~\ref{fig:fb15k237loss_fct_each_model} (Appendix \pageref{fig:fb15k237loss_fct_each_model}) expresses that some interaction models seem to be more sensitive to the usage of different loss function. For instance, ConvE and TuckER suffer from the \ac{mrl} and the \ac{nssal}, DistMult together with the \ac{cel} outperforms the other loss functions.
However, TransE performs similarly for all loss functions except the \ac{nssal}.

\begin{figure}[ht]
\centering
\includegraphics[width=0.4\textwidth]{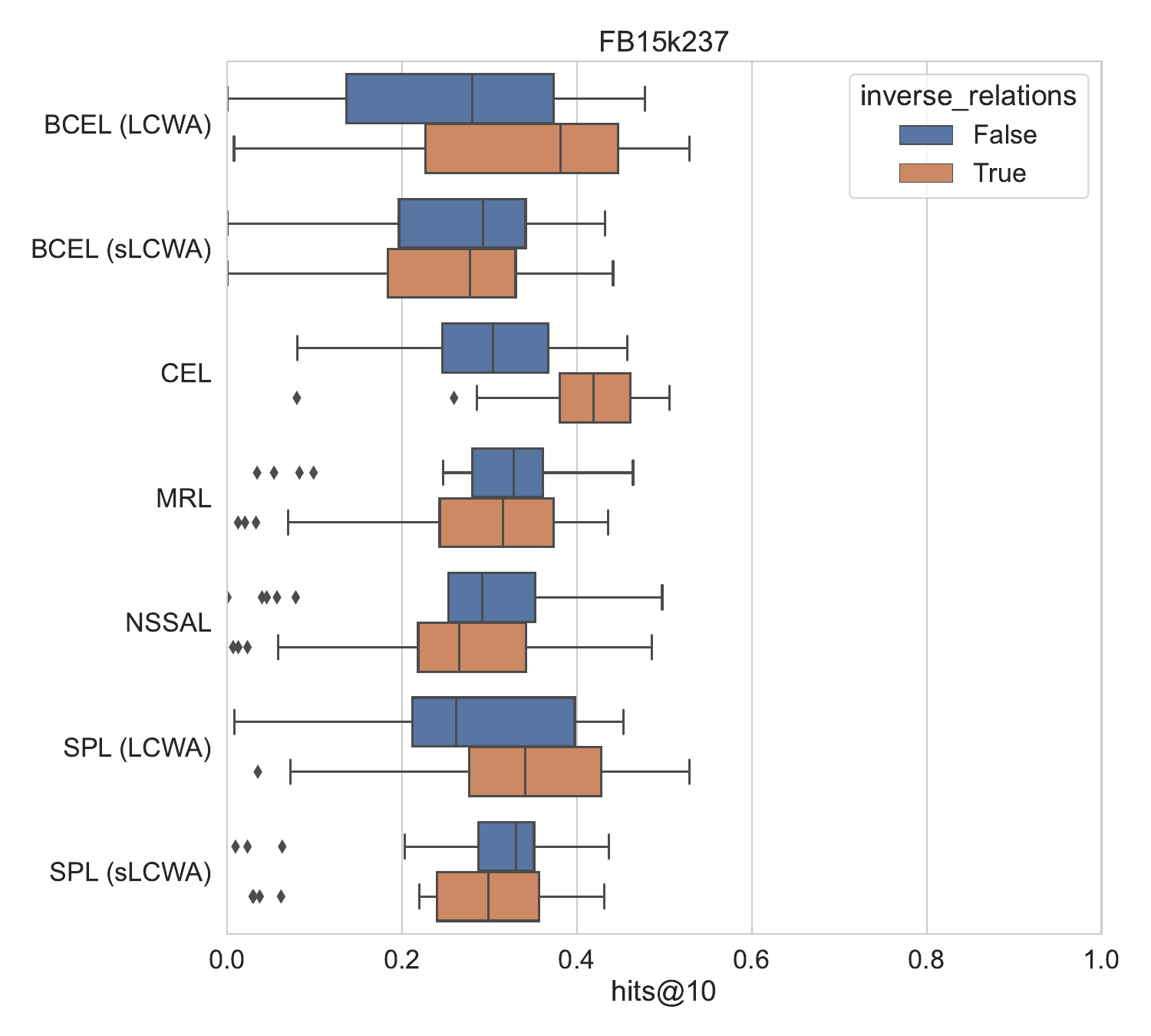}
\caption{Impact of explicitly modeling inverse relations on the performance for a fixed loss function for the FB15K-237 dataset.}
\label{fig:fb15k237_inverse_loss}
\end{figure}

\textbf{Impact of Explicitly Modeling Inverse Relations} Figure~\ref{fig:fb15k237_inverse_loss} reveals, as for the previous datasets, that in general, the usage of inverse relations is crucial for the training based on the \ac{lcwa} approach. 
Different from the results obtained for WN18RR, the \ac{lcwa} is not competitive when trained without inverse relations.

\textbf{Model Complexity vs. Performance} Figure~\ref{fig:model_size_performance} (Appendix \pageref{fig:model_size_performance}) illustrates that for FB15K-237, there is no clear correlation between model size and performance. Tiny models can already obtain similar performance as larger models.
The skyline comprises an intermediate \ac{um}, TransE and DistMult models, and a larger TuckER model.
A full list is provided in Table~\ref{tab:skyline_fb15k237_model_bytes} (Appendix \pageref{tab:skyline_fb15k237_model_bytes}).

\subsection{Results on the YAGO3-10 Dataset}

\begin{figure*}[tb]
\centering
\includegraphics[width=0.8\linewidth]{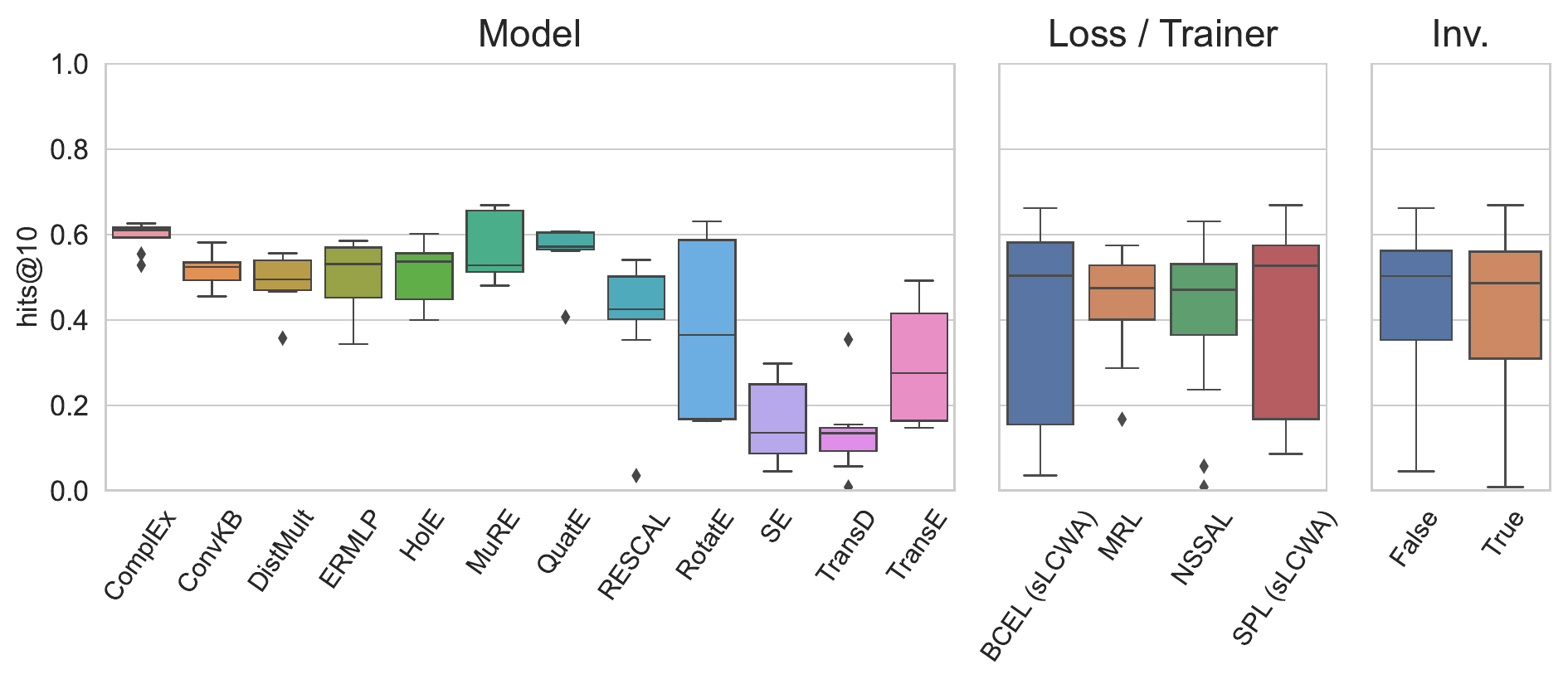}
\caption{Overall hits@10 results for YAGO3-10 where box-plots summarize the results across different combinations of interaction models, training approaches, loss functions, and the explicit usage of inverse relations. In contrast, to the previous datasets, the models have only been trained based on the \acl{slcwa}.}
\label{fig:summary_yago310}
\end{figure*}

YAGO3-10 is the largest benchmark dataset in our study.
Therefore, it is of interest to investigate how the different interaction models perform on a larger \ac{kg}.
As mentioned in the introduction of this chapter, we reduced the experimental setup for YAGO3-10 in order to reduce the computational complexity of our entire study.
Figure~\ref{fig:summary_yago310} depicts the overall results obtained for YAGO3-10. 
Detailed results for all configurations are illustrated in Figure~\ref{fig:all_configs_yago310_adam} in Appendix \pageref{fig:all_configs_yago310_adam}.

The results highlight the previous observation that the performance of many \acp{kgem} heavily depends on the choice of its components and is dataset-specific.
For instance, MuRE, the best-performing interaction model, and especially RotatE, which is among the top-performing interaction models, exhibit high variance across their configurations.
TransE, which was among the top-performing interaction models on WN18RR, performed poorly on YAGO3-10.
One might conclude that TransE performs better on smaller \acp{kg}, but the results obtained on Kinships do not support this assumption.
It should be taken into account that some interaction models might benefit from being trained with the \ac{lcwa} on YAGO3-10 as observed for TransE on WN18RR.
Therefore, TransE might perform much better when trained with the \ac{lcwa} approach.
Remarkably, ComplEx and QuatE seem to be robust for all \ac{slcwa} configurations.
With regards to the loss functions, all loss functions except \ac{mrl} obtain comparable results.
Though, the \ac{mrl} is more robust than other loss functions.

\textbf{Impact of the Loss Function} 
Figure~\ref{fig:summary_yago310} shows again that the choice of the loss functions has an import impact on the models' performance: the \acl{mrl} and the \acl{nssal} are less competitive than the \acl{bcel}/\acl{spl}.
Figure~\ref{fig:all_configs_yago310_adam} (Appendix A 14) highlights that some interaction models are susceptible to the choice of the loss function. 
For instance, RotatE and TransE suffer when trained with \ac{bcel} and \ac{spl} whereas ERMLP suffers when trained with the \ac{mrl}.

\begin{figure}[ht]
\centering
\includegraphics[width=0.4\textwidth]{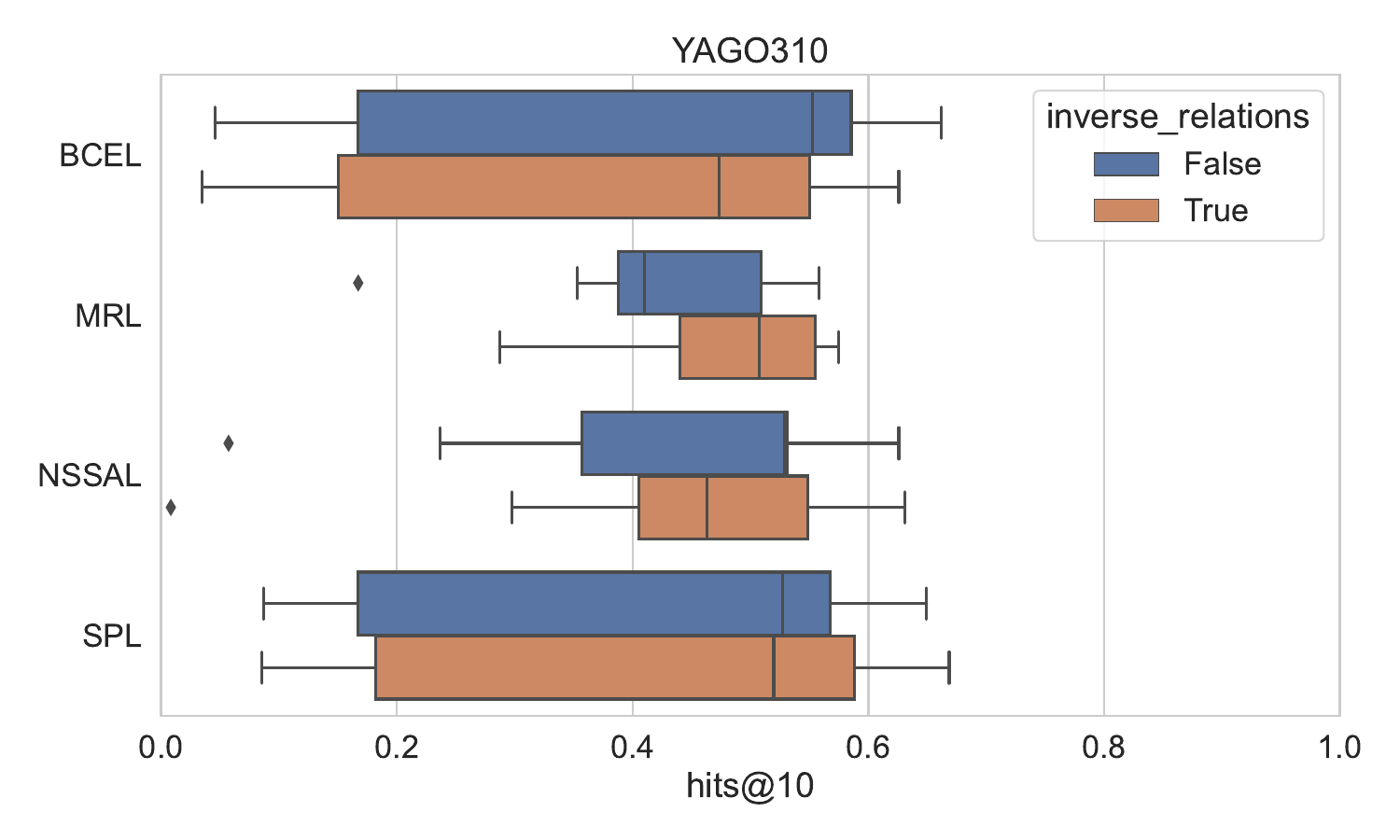}
\caption{Impact of explicitly modeling inverse relations on the performance for a fixed loss function for the YAGO3-10 dataset.}
\label{fig:yago310_inverse_triples_ta}
\end{figure}

\textbf{Impact of Explicitly Modeling Inverse Relations} Figure~\ref{fig:yago310_inverse_triples_ta} shows the effect of explicitly modeling inverse relations for fixed loss functions (it should be noted that the results are obtained based only on the \ac{slcwa} training approach).
In contrast to the results observed for WN18RR and FB15K-237, the \ac{mrl} benefits from explicitly modeling inverse relations.
Furthermore, also the \ac{spl} obtains its best performance with inverse inverse relations.

\begin{figure}[H]
\centering
\includegraphics[width=0.8\textwidth]{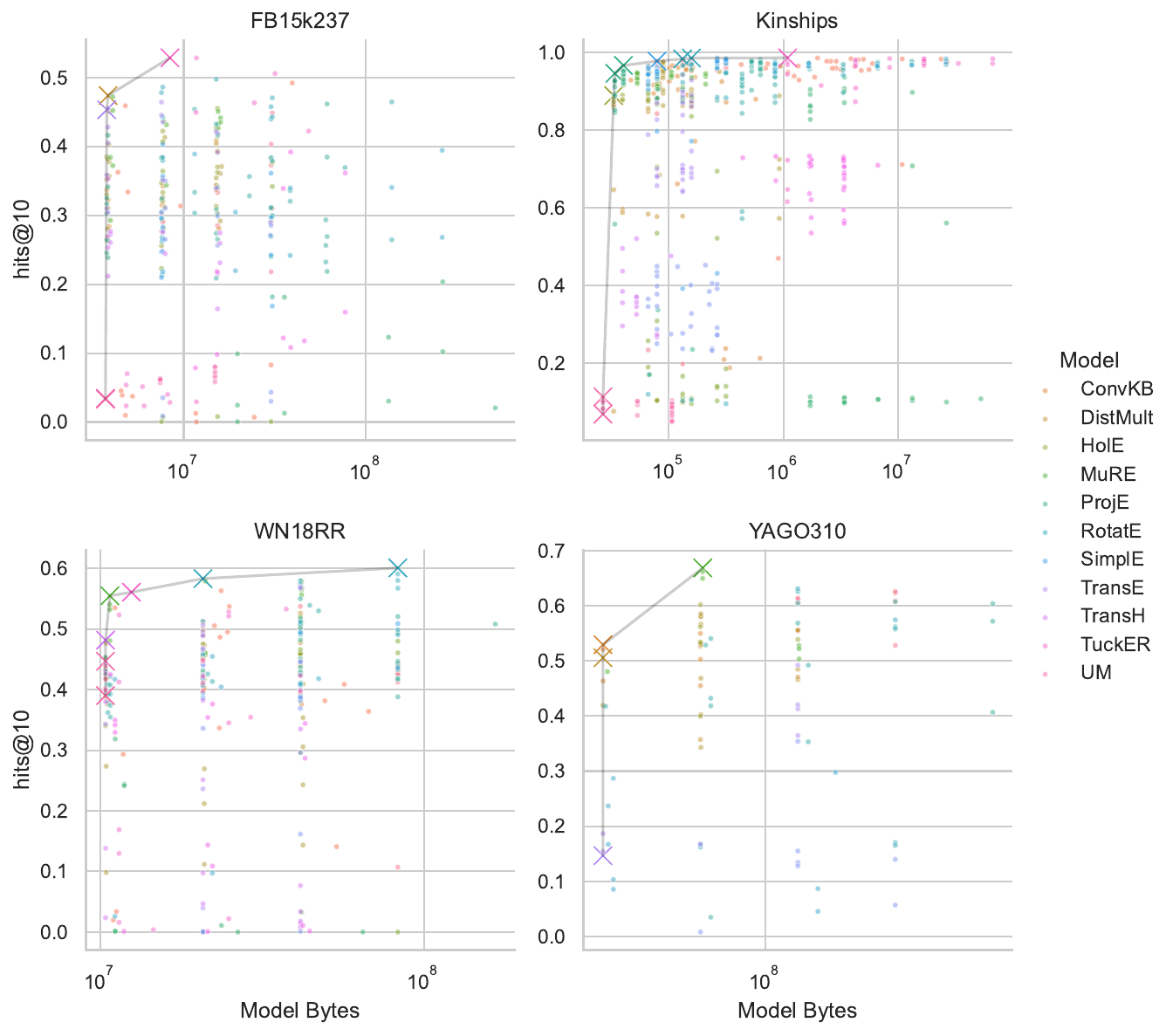}
\caption{
Scatter plots comparing model size in number of bytes and model performance in terms of Hits@10 for all trained models on each dataset.
The color indicates the model type, and the model size is shown on a logarithmic axis.
Pareto-optimal models are highlighted by cross symbols.
In general we only see a low correlation between model size and performance.
A more thorough comparison
can be found in Figures~\ref{fig:summary_kinhsips}, \ref{fig:summary_wn18rr}, \ref{fig:summary_fb15k237}, and \ref{fig:summary_yago310}.}
\label{fig:model_size_performance}
\end{figure}

\textbf{Model Complexity vs. Performance} Figure~\ref{fig:model_size_performance} (Appendix \pageref{fig:model_size_performance}) expresses that there is a low correlation between model size and performance for YAGO3-10.
However, the improvement is tiny compared to the differences in model size.
It should be taken into account that for \acp{kgem}, the model size is usually dependent on the number of entities and relations.
Therefore, dependent on the space complexity of the interaction model (Table~\ref{tab:supported_kge_models} in Appendix \pageref{tab:supported_kge_models}), the size can grow fast for large \acp{kg}.
The skyline comprises an intermediate TransE, DistMult and ConvKB model, and a larger MuRE model.
A full list is provided in Table~\ref{tab:skyline_yago310_model_bytes} (Appendix \pageref{tab:skyline_yago310_model_bytes}).

\section{Relational Pattern Analysis}\label{relational_pattern_analysis}

\Aclp{kg} exhibit relational patterns such as symmetry (e.g., the relation \textit{marriedTo}), and the performance of \acp{kgem} depend on how well these patterns can be modeled. 
Four major relational patterns that have been investigated in the literature are \textit{symmetry}, \textit{anti-symmetry}, \textit{inversion}, and \textit{composition}~\cite{Sun2019,Trouillon2016,toutanova2015observed}.
Here, we provide a large-scale performance analysis of our investigated \acp{kgem} in modeling \textit{symmetry}, \textit{anti-symmetry}, and \textit{composition} patterns for the datasets FB15k-237, WN18RR, and YAGO3-10.
First, we provide statistics about the \textit{support} and \textit{confidence} of the symmetry, anti-symmetry, inversion, and composition patterns in the FB15k-237, WN18RR and YAGO3-10 datasets.
Next, we describe our experimental setup. Finally, we present the results of our relational pattern analysis.

\subsection{Relational Patterns and their Detection}
\label{relation_patterns_definition}

Here, we formally define the relational patterns symmetry, anti-symmetry, inversion, and composition patterns according to~\cite{Sun2019}, the measures \textit{support} and \textit{confidence}, and provide an overview of the \textit{support} and \textit{confidence} of the these patterns in the FB15k-237, WN18RR and YAGO3-10 datasets.

\begin{definition}[Symmetric Relation]
  A relation $r \in \mathcal{R}$ is \textbf{symmetric}, if $(h, r, t) \in \mathcal{T} \implies (t, r, h) \in \mathcal{T}$
\end{definition}

\begin{definition}[Anti-Symmetric Relation]
  A relation $r \in \mathcal{R}$ is \textbf{anti-symmetric}, if $(h, r, t) \in \mathcal{T} \implies (t, r, h) \notin \mathcal{T}$
\end{definition}

\begin{definition}[Inverse Relation]
  A relation $r \in \mathcal{R}$ is \textbf{inverse} to $r_{inv} \in \mathcal{R}$, if $(h, r, t) \in \mathcal{T} \implies (t, r_{inv}, h) \in \mathcal{T}$.
  If there exists a $r' \in \mathcal{R}$ with $r' \neq r$ and $r'$ is inverse to $r$, then we call $r$ an inverse relation.
\end{definition}

\begin{definition}[Composite Relation]
  A relation $r \in \mathcal{R}$ is a \textbf{composition} of two relations $r_1, r_2 \in \mathcal{R}$, if $(a, r_1, b) \in \mathcal{T} \land (b, r_2, c) \in \mathcal{T} \implies (a, r, c) \in \mathcal{T}$.
  We call $r$ a composite relation, if such two relations exist.
\end{definition}

Since \acp{kg} are known to be incomplete, a false antecedent, i.e., right-hand side of a rule, may not only be caused by the relation not being of the relation type of interest, but also originate from the \ac{kg}'s incompleteness.
Thus, we detect relation types using a support and confidence threshold, defined akin to the concepts of association rule mining.

The \emph{support} of one of the aforementioned patterns $p$ for a relation $r$ indicates the number of different assignments of entities such that the precedent, i.e., the left-hand side of a rule, holds.
For most of the simple rules this is equivalent to the relation frequency, but, e.g., for composite relations, we need to consider all pairs of triples with matching the candidate relations $r_1, r_2$ and being linked by the intermediate entity $b$.

The \textit{confidence} of a relational pattern is the number of times the right-hand side holds divided by the support.
Thus, it can be interpreted as an estimate of the the conditional probability of the antecedent, given the precedent holds.

\subsection{Relation Patterns in Benchmark Datasets}

\label{relation_patterns_stats}
\begin{table}
    \centering
    \caption{Frequency of detected relation patterns across the benchmark datasets.}
    \label{tab:pattern_freq}
\begin{tabular}{lrrr}
\toprule
pattern &  anti-symmetry &  composition &  symmetry \\
dataset  &                &              &           \\
\midrule
fb15k237 &            205 &          147 &         3 \\
wn18rr   &              7 &            1 &         3 \\
yago310  &             30 &            3 &         2 \\
\bottomrule
\end{tabular} \end{table}

Table~\ref{tab:pattern_freq} shows the frequency of the detected pattern types for the three studied benchmark datasets.
Similar to related work we used a confidence threshold of 97\%~\cite{toutanova2015observed}.
Note that we did not detect a single inverse relation, since FB15k-237 and WN18RR have been explicitly preprocessed to remove such.

\subsection{Experimental Setup}\label{relation_patterns_es}
To measure the performance of the investigated \acp{kgem} in modeling symmetry, anti-symmetry, and composition patterns, we slightly adapted the standard link prediction evaluation procedure (Section~\ref{sec:evaluation_of_models}).
Instead of computing the metrics based on all test triples, we extracted for each relational pattern all test triples that contain the associated relations, aggregated the single ranks obtained of each triple in the subset, and computed the hits@10 metric for each subset.
Therefore, we can express how well a \ac{kgem} can model a specific relational pattern.

\subsection{Results}\label{relation_patterns_results}

\begin{figure}
    \centering
    \includegraphics[width=\linewidth, height=4cm, keepaspectratio]{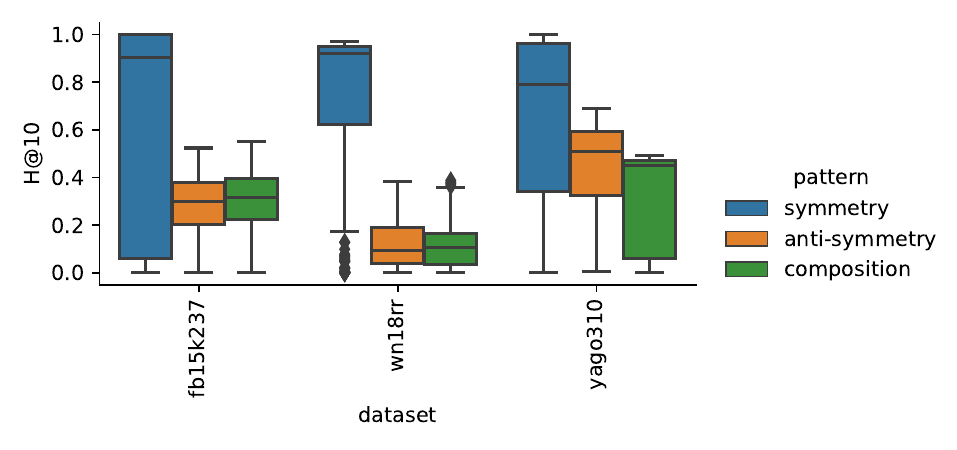}
    \caption{
    Performance Distribution of all best models per configuration in H@10.
    }
    \label{fig:pattern-results-overall}
\end{figure}

Figure~\ref{fig:pattern-results-overall} shows the overall performance on pattern types per dataset.
We show the distribution of best models' performance for each configuration in terms of H@10.
We generally observe a tendency that symmetric relations are easier to model than anti-symmetric and composite relations, which seem to be equally challenging.

\begin{figure}
    \centering
    \includegraphics[width=\linewidth]{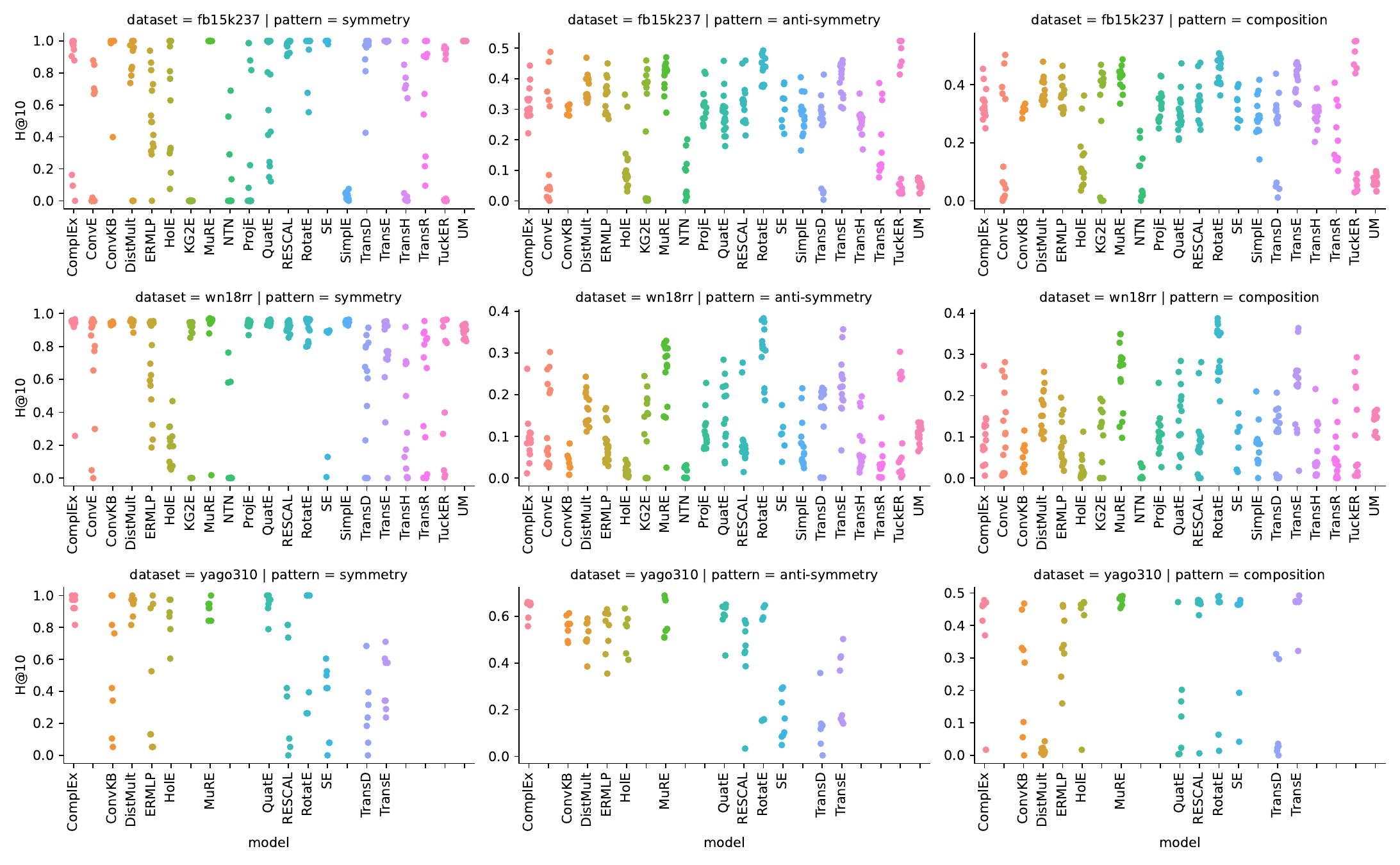}
    \caption{
    Performance of best models' for each configuration for each dataset and pattern type, grouped by interaction function.
    }
    \label{fig:pattern-results-model}
\end{figure}

Figure~\ref{fig:pattern-results-model} (Appendix A2) shows the performance of best models' for each configuration for each dataset and pattern type, grouped by interaction function.
For the most simple pattern, symmetry, almost all interaction functions can obtain strong results on WN18RR, with NTN, TransD and SE slightly falling behind.
For FB15k237, we observe similar results, except that SimplE and KG2E fail to capture this pattern (while performing still sufficiently good on other patterns).
On YAGO3-10, translation-based methods such as TransE or TransD cannot match the performance of, ComplEx, RotatE and DistMult, with ER-MLP's performance in between.

On the more difficult anti-symmetry and composition patterns, the differences are more pronounced.
Overall, RotatE and TransE obtain the best results, whereas UM and NTN cannot obtain good results.

\section{Discussion \& Future Work}\label{sec:discussion_future_work}

Table~\ref{tab:eval_stats} (Appendix \pageref{tab:eval_stats}) illustrates the extent of our studies and Table~\ref{tab:summary} (Appendix \pageref{tab:summary}) summarizes the main findings our work.
Although the re-implementation of all machine learning components into a unified, fully configurable framework was a major effort, we believe it is essential to analyze reproducibility and obtain fair results on benchmarking.
In particular, we were able to address the issue of incompatible evaluation procedures and preprocessing steps in previous publications that are not obvious. 
We highlighted that the evaluation metrics, which usually are utilized to evaluate the performance of \aclp{kgem}, are realized differently depending on the definition of the \textbf{rank}. Specifically, three major rank definitions are employed: \textit{optimistic}, \textit{realistic}, and \textit{pessimistic} ranking. 
Because the optimistic and pessimistic ranking can lead to distorted conclusions in cases where a \ac{kgem} predicts the same score for many triples, we recommend evaluating \aclp{kgem} based on the realistic ranking approach.

\begin{table*}[ht]
    \caption{Summary of main insights over all datasets. Each component (i.e., interaction model, loss function, and training approach) is considered to be among the top-ten performing configurations when they occur at least once in the top-ten performing configurations. Note that a single component is part of several configurations, and therefore, can occur multiple times in the top-ten performing configurations.}
    \begin{tabularx}{\textwidth}{p{2cm}X}
        \toprule
         \multicolumn{2}{l}{\emph{Interaction Models}} \\\midrule  
         RotatE & Among top-ten-performing interaction models across all datasets.\\
         MuRE & Among top-ten-performing interaction models on WN18RR, FB15K-237, and YAGO3-10.\\
         ConvE & Among top-ten-performing interaction models on Kinships and FB15K-237 (has not been evaluated on YAGO3-10).\\
         ComplEx & Among top-ten-performing interaction models on Kinships and YAGO3-10.\\
         TuckER & Among top-ten-performing interaction models for Kinships, and FB15K-237 (has not been evaluated on YAGO3-10).\\
         DistMult & Among top-ten-performing interaction models on FB15K-237.\\
         QuatE & Among top-ten-performing interaction models on YAGO3-10.\\
         TransE & Among top-ten-performing interaction models on WN18RR.\\
         \ac{se} & Among top-ten-performing interaction models on Kinships.\\
         \midrule  
        \multicolumn{2}{l}{\emph{Loss Functions}} \\\midrule 
         \Ac{bcel} & Among top-ten-performing loss functions across all datasets.\\
         \Ac{nssal} & Among top-ten-performing loss functions across all datasets.\\
         \Ac{spl} & Among top-ten-performing loss functions across all datasets.\\
         \Ac{cel} & Among top-ten-performing loss functions on Kinships and FB15K-237 (has not been evaluated on YAGO3-10).\\
         \Ac{mrl} & Among top-ten-performing loss functions on Kinships.\\
         \midrule 
        \multicolumn{2}{l}{\emph{Training Approaches}} \\\midrule 
         \ac{slcwa} & Among top-ten-performing training approaches across all datasets. \\ 
         \Ac{lcwa} & Among top-ten-performing training approaches  on Kinships, WN18RR and FB15K-237 (has not been evaluated on YAGO3-10). \\ 
         \midrule 
        \multicolumn{2}{l}{\emph{Explicit Modeling of Inverse Relations}} \\\midrule 
          & Is usually beneficial in combination with the \acl{lcwa}. \\ 
         \midrule 
        \multicolumn{2}{l}{\emph{Configurations}} \\\midrule 
         Performance & Appropriate combination of interaction model, training assumption, loss function, choice of explicitly modeling inverse relations is crucial for the performance, e.g., TransE can compete when with several state-of-the-art interaction models on WN18RR when appropriate configuration is selected.\\
         & There is no single best configuration that works best for all dataset. \\
         Variance & Some interaction models exhibit a high variance across different configurations, e.g., RotatE on YAGO3-10 (Figure \ref{fig:summary_yago310} on page \pageref{fig:summary_yago310}) \\ 
         Pareto-Optimal Configurations & Tables \ref{tab:skyline_fb15k237_model_bytes}-\ref{tab:skyline_yago310_model_bytes} in Appendix \pageref{tab:skyline_fb15k237_model_bytes} describe Pareto-optimal configurations. It can be seen that there are configurations that require fewer parameters while obtaining almost the same performance. In some cases, for the same interaction model, the model can be significantly compressed. %
         \\ 
         \midrule 
        \multicolumn{2}{l}{\emph{Reproducibility}} \\\midrule 
         Results & For FB15K, four out of 13, for WN18, five out of 13, for FB15K-237, two out of three, and for WN18RR, three out of five experiments can be categorized as soft-reproducible.\\ 
        Code & For four out of 15 models, no official implementation was available. \\ 
        Parameters & For six out of 15 papers, source code was available and full experimental setup was precisely described. \\ 
         \midrule 
        \multicolumn{2}{l}{\emph{General Insights}} \\\midrule 
        SOTA & For WN18RR, we achieve based on a RotatE-configuration (together with Graph Attenuated Attention Networks~\cite{wang2019knowledge}) state-of-the-art results in terms of hits@10 through our study (60.09\% Hits@10). Furthermore, we found a TransE configuration that achieves high performance beating most of the published SOTA results (56.98\% Hits@10). Based on our results, we emphasize to further investigate the hyper-parameters space for the most promising configurations for the remaining benchmarking datasets.\\ 
         Improvements & For ConvE (56.33\% compared to 52.00\%~\cite{Dettmers2018}), MuRE (57.90\% compared to 55.50\%~\cite{DBLP:conf/nips/BalazevicAH19}) and TuckER (56.09\% compared to 52.6\%~\cite{balavzevic2019tucker}), we are beating the reported results in the original papers due selecting appropriate configurations and hyper-parameters on WN18RR.\\
         \bottomrule
    \end{tabularx}
    \label{tab:summary}
\end{table*}

During our reproducibility study, we found that the reproduction of experiments is a major challenge and, in many cases, not possible with the available information in current publications.
In particular, we observed the following four main aspects:
\begin{itemize}
    \item For a set of experiments, the results can sometimes only be reproduced with a different set of hyper-parameter values.
    \item For some experiments, the entire experimental setup was not provided, impeding the reproduction of experiments.
    \item The lack of an official implementation hampers the reproduction of results.
    \item Some results are dependent on the utilized ranking approach (average, optimistic, and pessimistic ranking approach). For example, the optimistic rank may lead to incorrect conclusions about the model's performance.
\end{itemize}

Our benchmarking study shows that the term \ac{kgem} should be used with caution and should be differentiated from the actual interaction model since our results highlight that the specific \emph{combination} of the interaction model, training approach, loss function, and the usage of explicit inverse relations is often fundamental for the performance. 

No configuration performs best across all datasets. 
Depending on the dataset, several configurations can be found that achieve comparable results (Tables~\ref{best_models_fb15k237}-\ref{best_models_yago310} in Appendix \pageref{best_models_fb15k237}-\pageref{best_models_yago310}, and Figures~\ref{fig:all_configs_kinhsips_adam}-\ref{fig:all_configs_yago310_adam} in Appendix \pageref{fig:all_configs_kinhsips_adam}-\pageref{fig:all_configs_yago310_adam}).
Moreover, with an appropriate configuration, the model size can significantly be compressed (see Pareto-optimal configurations in Tables \ref{tab:skyline_fb15k237_model_bytes}-\ref{tab:skyline_yago310_model_bytes} in Appendix \pageref{tab:skyline_fb15k237_model_bytes}) that has especially a practical relevance when looking for a trade-off between required memory and performance.

The results also highlight that even interaction models such as TransE that have been considered as baselines can outperform state-of-the-art interaction models when trained with an appropriate training approach and loss function. 
This raises the question of the necessity of the vast number of available interaction models.
However, for some interaction models such as RotatE, MuRE or TuckER, we can observe a good performance across all datasets (note: TuckER has not been evaluated on YAGO3-10).
For RotatE, we even obtained the state-of-the-art results on WN18RR (similar results were obtained by Graph Attenuated Attention Networks~\cite{wang2019knowledge}), and for ConvE, MuRE, and TuckER, we obtained results superior to the originally published ones.
ComplEx proved to be a very robust interaction model across different configurations.
This can, in particular, be observed from the results obtained on YAGO3-10 (Figure~\ref{fig:summary_yago310}).

We discovered that no loss function consistently achieves the best results.
Instead it can be seen that with different loss functions, such as the \ac{bcel}, \ac{nssal}, and \ac{spl}, good results can be obtained across all datasets.
Remarkably, the \ac{mrl} is overall the worst-performing loss function. %
However, one might argue that the \ac{mrl} is the most compatible loss function with the \ac{slcwa} since it does not assume artificially generated negative examples to be actually false in contrast to the other loss functions used.
The \ac{mrl} only learns to score positive examples higher than \emph{corresponding} negative examples, but it does not ensure that a negative example is scored lower than every other positive example.
Thus, the absolute score values are not interpretable and cannot be used to compare triples without common head/tail entities. They can only be interpreted relatively, and only when comparing scores for triples with the same $(hr)$/$(rt)$.
Although loss functions such as \ac{bcel} or \ac{spl} treat generated negative triples as true negatives that actually contain also unknown positive examples, they obtain good performance. 
This might be explained by the fact that usually the set of unknown triples are dominated by false triples.
Therefore, it is likely that a major part of the generated triples are actually negative.
Consequently, the \ac{kgem} learns to distinguish better positive from negative examples.

Considering the explicit usage of inverse relations, we found out that the impact of inverse relations can be significant, especially when the interaction model is trained under the \ac{lcwa}.
This might be explained by the fact that based on the \ac{lcwa}-training, the \ac{kgem} only learns to perform \textit{one-side predictions} (i.e., it learns to either predict head or tail entities), but during the evaluation, it is asked to perform \emph{both-side predictions}.
Through the inclusion of inverse relations, the model learns to perform both-side predictions based on one side, i.e., $(*,r,t)$ can be predicted through $(t,r_{inverse},*)$.
Overall, our results indicate that further investigations on FB15K-237 and YAGO3-10 might lead to results that are competitive to the state-of-the-art.

Looking forward, it would be of great interest to re-investigate previously performed studies that analyze the relationship between the performance of \acp{kgem} and the properties of the underlying \acp{kg} to verify that their findings indeed can be attributed to the \emph{interaction model} alone, rather than the exact configuration including the loss function, the training approach and the explicit modeling of inverse relations.
Further, the effect of explicitly modeling inverse relations has not been analyzed in depth, in particular how the learned representations of a relation and its inverse are related to each other.
Ultimately, we believe our work provides an empirical foundation for such studies and a practical tool to execute them.

\section*{Acknowledgment}
We want to thank the Center for Information Services and High Performance Computing (ZIH) at TU Dresden for generous allocations of computer time and the Technical University of Denmark for providing us access to their DTU Compute GPU cluster that enabled us to conduct our studies.
This work was funded by the German Federal Ministry of Education and Research (BMBF) under Grant No. 01IS18036A and Grant No. 01IS18050D (project “MLWin”),
the Innovation Fund Denmark with the Danish Center for Big Data Analytics driven Innovation (DABAI),
and the 
Defense Advanced Research Projects Agency (DARPA) Automating Scientific Knowledge Extraction (ASKE) program under grant
HR00111990009.

\ifCLASSOPTIONcaptionsoff
  \newpage
\fi

\bibliographystyle{IEEEtran}
\bibliography{lib.bib}

\begin{IEEEbiography}[{\includegraphics[width=1in,height=1.25in,clip,keepaspectratio]{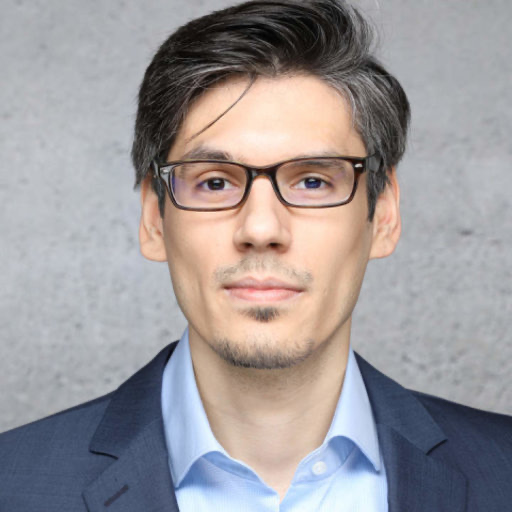}}]%
{Mehdi Ali}
Mehdi Ali received his M.Sc. degree in Computer Science with a focus on intelligent systems from the University of Bonn.
Currently, he is a Ph.D. candidate at the computer science department of the University of Bonn and a research associate at the Fraunhofer Institute IAIS.
In his Ph.D., he focuses on machine learning models for (knowledge) graphs, multi-modal models that combine graph and textual information, and reproducibility in the field of knowledge graph embedding models.

\end{IEEEbiography}

\begin{IEEEbiography}[{\includegraphics[width=1in,height=1.25in,clip,keepaspectratio]{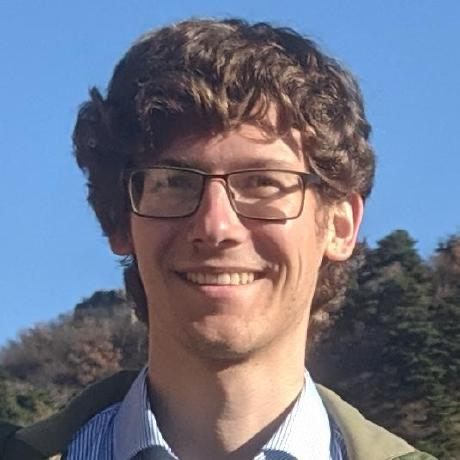}}]%
{Max Berrendorf}
Max Berrendorf received his B.Sc and M.Sc. degree in Computer Science with a minor in Mathematics from RWTH Aachen University. Currently he is pursuing a Ph.D. degree at the chair of Database Systems and Data Mining at Ludwig-Maximilians-Universität München.
In his research, he focuses on machine learning on graphs, in particular knowledge graphs, graph matching problems, and reproducibility in machine learning.
\end{IEEEbiography}

\begin{IEEEbiography}[{\includegraphics[width=1in,height=1.25in,clip,keepaspectratio]{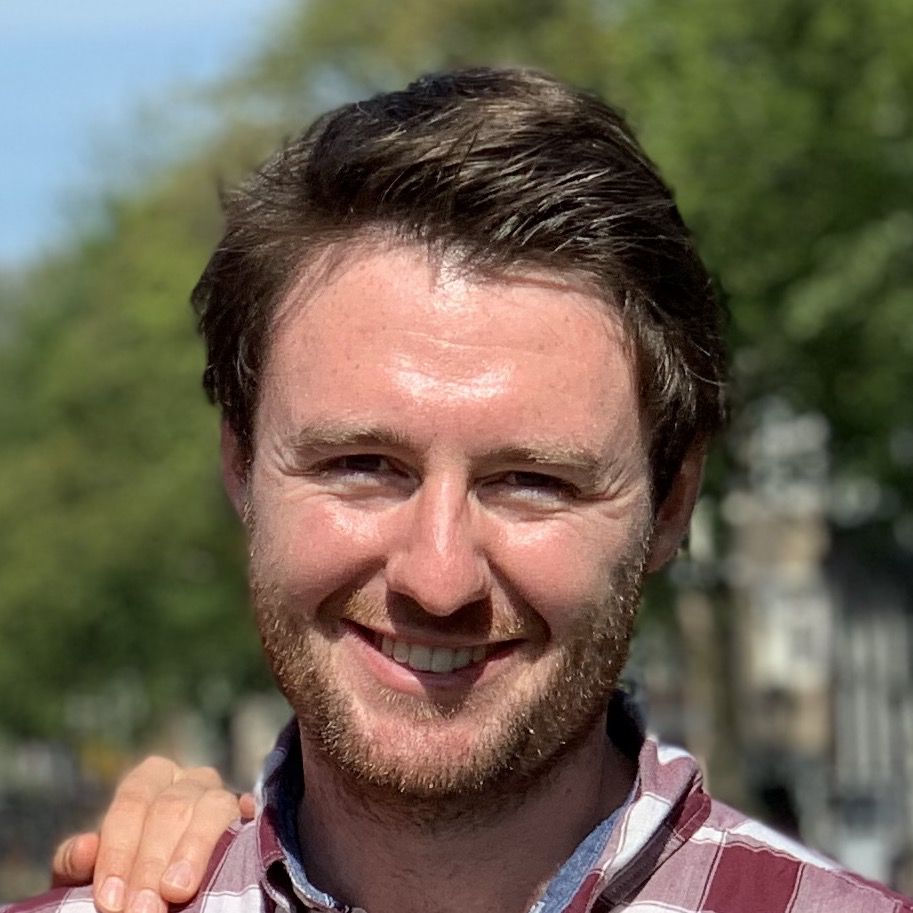}}]%
{Charles Tapley Hoyt}
Dr. Charles Tapley Hoyt completed his Ph.D. in Computational Life Sciences from the University of Bonn in 2019 and is now affiliated with the Laboratory of Systems Pharmacology at Harvard Medical School, Boston, USA.
His interests are in the biological applications of knowledge graph embedding models towards proteochemometrics, target prioritization, drug repositioning, predictive toxicology, and precision medicine.
\end{IEEEbiography}

\begin{IEEEbiography}[{\includegraphics[width=1in,height=1.25in,clip,keepaspectratio]{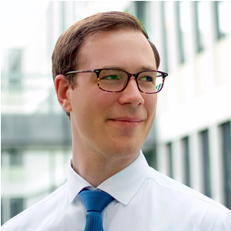}}]%
{Laurent Vermue}
Laurent Vermue received his M.Sc. degree in Industrial Engineering and Management at the Technical University of Berlin and MMSc. degree in Management Science and Engineering at the Tongji University. Currently he is a Ph.D. student at the Section for Statistics and Data Analysis and the Section for Cognitive Systems at DTU Compute, Technical University of Denmark. His research interests include machine learning, complex network modeling and open research software.
\end{IEEEbiography}

\begin{IEEEbiography}[{\includegraphics[width=1in,height=1.25in,clip,keepaspectratio]{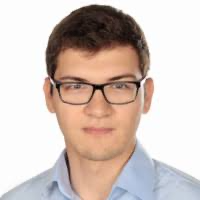}}]%
{Mikhail Galkin}
Dr. Mikhail Galkin received his Ph.D. degree in Computer Science from the University of Bonn in 2018 studying knowledge graphs, their creation, integration, and querying. Currently, he is a postdoctoral fellow at Montreal Institute for Learning Algorithms (Mila) and McGill University.
His interests include applications of knowledge graphs and graph representation learning to neural reasoning and natural language processing.

\end{IEEEbiography}

\begin{IEEEbiography}[{\includegraphics[width=1in,height=1.25in,clip,keepaspectratio]{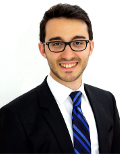}}]%
{Sahand Sharifzadeh}
Sahand Sharifzadeh received his M.Sc degree from Technical University of Munich majoring in Computer Vision and Artificial Intelligence. Currently, he is a Ph.D. candidate at Ludwig-Maximilians-Universität München.
In his research, he focuses on extracting graphs from images and text, as well as knowledge graph modeling.
He often collaborates with biologists, physicists and robotic engineers as interdisciplinary machine learning research is one of his interests.
\end{IEEEbiography}

\begin{IEEEbiography}[{\includegraphics[width=1in,height=1.25in,clip,keepaspectratio]{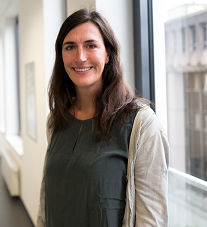}}]%
{Asja Fischer}
Asja Fischer is professor for machine learning at Ruhr University Bochum. %
Her research interests are focus on the development, analysis, and application of deep learning models and methods. %
Before becoming a professor in Bochum she was assistant professor at Bonn university, and a post-doctoral researcher at the Montreal Institute for Learning Algorithms (MILA). Between 2010 and 2015, she was employed both at the Institute for Neural Computation at the Ruhr University Bochum and the Department of Computer Science at the University of Copenhagen working on her PhD, which she defended in Copenhagen in 2014. Before, she studied Biology, Bioinformatics, Mathematics, and Cognitive Science at the Ruhr-University Bochum, the Universidade de Lisboa, and the University of Osnabrück.

\end{IEEEbiography}

\begin{IEEEbiography}[{\includegraphics[width=1in,height=1.25in,clip,keepaspectratio]{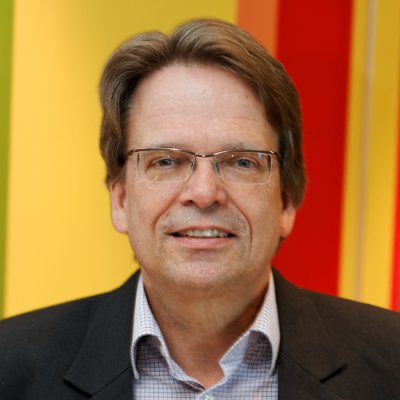}}]%
{Volker Tresp}
Volker Tresp received the Diploma degree from the University of Goettingen, Germany, in 1984 and the M.Sc. and Ph.D. degrees from Yale University, New Haven, CT, USA, in 1986 and 1989, respectively.
Since 1989, he has been the head of various research teams in machine learning at Siemens, Research and Technology, Munich, Germany. He filed more than 70 patent applications and was inventor of the year of Siemens in 1996. He has published more than 100 scientific articles and administered over 20 Ph.D. dissertations. The company Panoratio is a spin-off out of his team. His research focus in recent years has been machine learning in information networks for modeling knowledge graphs, medical decision processes, and sensor networks. He is the coordinator of one of the first nationally funded big data projects for the realization of precision medicine. In 2011, he became a Honorary Professor at the Ludwig Maximilian University of Munich, Germany, where he teaches an annual course on machine learning.

\end{IEEEbiography}

\begin{IEEEbiography}[{\includegraphics[width=1in,height=1.25in,clip,keepaspectratio]{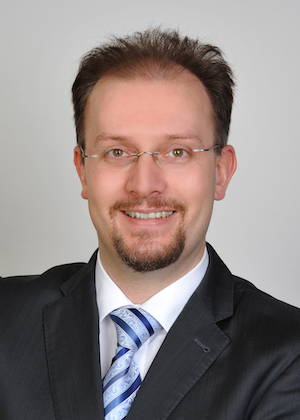}}]%
{Jens Lehmann}
Prof. Dr. Jens Lehmann leads the "Smart Data  Analytics" research group at the University of Bonn and Fraunhofer IAIS  with 40 researchers.
His research interests involve knowledge graphs, machine learning, question answering, distributed computing and knowledge representation. 
He is particularly excited about the combination of data- and knowledge-driven AI methods. 
Prof.~Lehmann won more than 10 international awards for his research work. 
He is founder, leader or contributor of several community research projects, including SANSA, DL-Learner, DBpedia and LinkedGeoData.
Previously, he completed his PhD with "summa cum laude" at the University of Leipzig with visits to the University of Oxford. He studied Computer Science at the Technical University of Dresden.
\end{IEEEbiography}

\clearpage
\pagenumbering{arabic}%
\renewcommand*{\thepage}{A\arabic{page}}
\clearpage

\begin{table}[t]
\centering
\caption{Investigated interaction models~\cite{Nickel2016} and their required number of parameters. $k$ corresponds to the number of neurons in the hidden layer, $n_f$ to the number of convolutional kernels, $k_r$ and $k_c$ to the height and width of the convolutional kernels.}\label{tab:supported_kge_models}
\begin{threeparttable}
\begin{tabular}{ll}
\toprule
Model & Parameters\\
\midrule
ComplEx\tnote{a}       & $|\mathcal{E}| 2d + |\mathcal{R}| 2d$ \\
ConvE\tnote{b}          & \begin{tabular}[c]{@{}l@{}l@{}l@{}}$|\mathcal{E}|d + |\mathcal{R}|d + d + n_f k_r k_c + 2 + 2n_f + 2d$ \\ $+ (h-k_r+1) (w-k_c+1)n_f d  +|\mathcal{E}|$\end{tabular} \\
ConvKB      & $|\mathcal{E}| d + |\mathcal{R}| d + n_f (d + 4) + 1$ \\
DistMult          & $|\mathcal{E}|  d  + |\mathcal{R}| d$ \\
ER-MLP             & $|\mathcal{E}| d + |\mathcal{R}| d + k(3d + 2) + 1$ \\
HolE             & $|\mathcal{E}|  d  + |\mathcal{R}| d$ \\
KG2E         & $|\mathcal{E}|2d + 2 |\mathcal{R}|d$ \\
MuRE        & $|\mathcal{E}| (d + 2) + 3 |\mathcal{R}|d$ \\
NTN     & $|\mathcal{E}| d + |\mathcal{R}|k(d^2 + 2 d + 2)$ \\
ProjE          & $|\mathcal{E}| d + |\mathcal{R}| d + 3d + 1$ \\
QuatE\tnote{c}           & $|\mathcal{E}|  4d  + |\mathcal{R}|  4d$ \\
RESCAL           & $|\mathcal{E}|  d  + |\mathcal{R}|  d^2$ \\
RotatE\tnote{a}    & $|\mathcal{E}| 2d + |\mathcal{R}| d$ \\
\ac{se}      & $|\mathcal{E}| d + 2 |\mathcal{R}| d^2$ \\
SimplE     & $|\mathcal{E}| 2d + 2|\mathcal{R}|d$ \\
TransE  & $|\mathcal{E}|  d  + |\mathcal{R}|  d$ \\
TransH    & $|\mathcal{E}|  d  +  2 |\mathcal{R}| d$ \\
TransR     & $|\mathcal{E}|  d_e +  |\mathcal{R}| d_r + d_e d_r$ \\
\ac{um}     & $|\mathcal{E}|d$ \\
TuckER & $|\mathcal{E}| d_e  + |\mathcal{R}| d_r + d_e^2 d_r + 4d_e$ \\
\bottomrule
\end{tabular}
\begin{tablenotes}
        \item[a] $2 d$, because of complex valued vectors, i.e.\ imaginary and real part of a number.
        \item[b] $w$ and $h$ correspond to the height and weight of the reshaped input.
        \item[c] $4 d$, because of hyper-complex valued (quaternion) vectors, i.e.\ a real part and three imaginary parts of a quaternion.
    \end{tablenotes}
\end{threeparttable}
\end{table}

\begin{table}[t]
\caption{Denotes for each proposed model whether results have been reported for FB15K, WN18, or their alterations.
Furthermore, it indicates whether an official implementation exists where P corresponds to a PyTorch based implementation, T to a TensorFlow based implementation, and O to other implementations.
A green background indicates that the full experimental setup was available. The models highlighted with * where included in the reproducibility study.}
\label{fig:overview_model_dataset}
\begin{threeparttable}
\begin{tabular}{lccccc}
\toprule
Model & Code & FB15K & FB15K-237 & WN18 & WN18RR\\
\midrule
ComplEx* & O &\cellcolor[HTML]{ADF7AD}\checkmark &    &  \cellcolor[HTML]{ADF7AD}\checkmark    &   \\
ConvE* &  P & \checkmark & \checkmark & \checkmark  &  \checkmark \\
ConvKB* & T & & \checkmark &     & \checkmark  \\
DistMult* &- & \cellcolor[HTML]{ADF7AD}\checkmark &   & \cellcolor[HTML]{ADF7AD}\checkmark &   \\
ER-MLP &- & & & &      \\
HolE*  & O &\checkmark& &\checkmark&     \\
KG2E* &- &\cellcolor[HTML]{ADF7AD}\checkmark& &\cellcolor[HTML]{ADF7AD}\checkmark&     \\
MuRE* &P &  & \cellcolor[HTML]{ADF7AD}\checkmark &  &  \cellcolor[HTML]{ADF7AD}\checkmark \\
\ac{ntn} &- & & & &       \\
ProjE &T &\cellcolor[HTML]{ADF7AD}\checkmark& & \cellcolor[HTML]{ADF7AD}\checkmark &       \\
QuatE*\tnote{a}  & P & \cellcolor[HTML]{ADF7AD}\checkmark & \cellcolor[HTML]{ADF7AD}\checkmark & \cellcolor[HTML]{ADF7AD}\checkmark  &  \cellcolor[HTML]{ADF7AD}\checkmark \\
\ac{um}  &- & & & &   \\
RESCAL & O & & & &      \\
RotatE* & P & \checkmark & \checkmark & \checkmark & \checkmark\\
\ac{se} & O & & & &       \\
SimplE*  &T, P &\cellcolor[HTML]{ADF7AD}\checkmark& &\cellcolor[HTML]{ADF7AD}\checkmark&     \\
TransD* & - &\cellcolor[HTML]{ADF7AD}\checkmark& &\cellcolor[HTML]{ADF7AD}\checkmark&    \\
TransE* & O &\checkmark& &\checkmark& \\
TransH* & - &\cellcolor[HTML]{ADF7AD}\checkmark& &\cellcolor[HTML]{ADF7AD}\checkmark&  \\
TransR* &O &\cellcolor[HTML]{ADF7AD}\checkmark& &\cellcolor[HTML]{ADF7AD}\checkmark&   \\
TuckER*  & P & \cellcolor[HTML]{ADF7AD}\checkmark & \cellcolor[HTML]{ADF7AD}\checkmark & \cellcolor[HTML]{ADF7AD}\checkmark  &  \cellcolor[HTML]{ADF7AD}\checkmark \\
\ac{um}  &- & & & &   \\
\bottomrule
\end{tabular}%
\begin{tablenotes}
        \item[a] Code is based on the framework OpenKE~\url{https://github.com/thunlp/OpenKE}.
    \end{tablenotes}
\end{threeparttable}
\end{table}

\begin{table}[t]
    \centering
    \caption{Hyper-Parameter Ranges for Ablation Experiments}
    \label{tab:hpo_params}
    \begin{threeparttable}
    \begin{tabular}{lll}
        \toprule
        & Hyper-Parameter     & Range  \\
        \midrule
        \multirow{7}{*}{\rotatebox[origin=c]{90}{Shared}}
        & Embedding-Dimension & \{64,128,256\} \\
        & Initialization & \{Xavier\} \\
        & Optimizers\tnote{a} & \{Adam, Adadelta\} \\
        & Learning Rate (log scale) & [0.001, 0.1) \\
        & Batch Size\tnote{b} & \{128, 256, 512\} \\
        & Model inverse relations & \{Yes, No\} \\
        & Epochs & 1,000 \\
        \midrule
        \multirow{5}{*}{\rotatebox[origin=c]{90}{\ac{slcwa}}}
        & Loss & \{\ac{bcel}, \ac{mrl}, \ac{nssal}, \ac{spl}\} \\
        & Margin for \ac{mrl} & \{0.5, 1.5, ... , 9.5\} \\
        & Margin for \ac{nssal} & \{1, 3, 5, ... , 29\} \\
        & ADVT for \ac{nssal} & \{0.1, 0.2, ... , 1.0\} \\
        & Number of Negatives\tnote{c} & \{1, 2, ... , 100\} \\
         \midrule
         \multirow{3}{*}{\rotatebox{90}{\ac{lcwa}}}
         & Loss & \{\ac{bcel}, \ac{cel}, \ac{spl}\} \\
         & Label Smoothing (log scale) & [0.001, 1.0)\\
         &&\\
        \bottomrule
    \end{tabular}
    \begin{tablenotes}
        \item[a] For Kinships, we evaluated Adam and Adadelta, and for the remaining datasets we sticked to Adam since it performed almost in every experiment at least equally good as Adadelta and in many experiments significantly better.
        \item[b] For YAGO3-10, the batch-size has been sampled from the set \{1024, 2048, 2096, 8192\}.
        \item[c] For YAGO3-10, the number of negative triples per each each positive has been sampled from the set \{1, 2, ..., 50\}.
    \end{tablenotes}
\end{threeparttable}
\end{table}{}

\begin{table}
    \centering
    \renewcommand{\arraystretch}{1.2}
    \caption{Evaluation statistics}
    \begin{tabular}{lr}
        \toprule 
        Metric & Value \\
        \midrule 
         Datasets & 4 \\
         Interaction Models & 21 \\
         Training approaches & 2 \\
         Loss Functions & 5 \\
         Negative Samplers & 1 \\
         Optimizers & 2 \\
         Ablation Studies & 1,207 \\
         Number of Experiments & 73,683 \\
         Compute Time (hours) & 24,804 \\
         \bottomrule
    \end{tabular}
    \label{tab:eval_stats}
\end{table} \clearpage
\onecolumn
\clearpage

 \clearpage
\section*{Additional Results From Reproducibility Study}

\begin{table}[H]
    \centering
    \caption{Reproduction of Studies on FB15K where \textbf{pub} refers to published results, \textbf{R} to results based on the realistic ranking, \textbf{O} to results based on the optimistic ranking, and \textbf{P} to results based on the pessimistic ranking. 
    For published results, there are two additional rank types, \textbf{U} for undefined due to missing official implementation and \textbf{ND} for non-deterministic.
    We only show the results of the optimistic and pessimistic ranking in case they differ from the realistic ranking.}
    \label{tab:rps_fb15k}
\begin{tabular}{llrrrrrrr}
\toprule
       &   &       MRR (\%) &   Hits@1 (\%) &    Hits@3 (\%) &    Hits@5 (\%) &   Hits@10 (\%) &                MR &      AMR (\%) \\
\textbf{model} & {} &                &               &                &                &                &                   &               \\
\midrule
\textbf{\textcolor{RedOrange}{ComplEx}} & \textbf{pub (O)} &     69.20 \phantom{$\pm$} $\phantom{5}$\phantom{0.00} &     59.90 \phantom{$\pm$} \phantom{0.00} &     75.90 \phantom{$\pm$} $\phantom{5}$\phantom{0.00} &                &     84.00 \phantom{$\pm$} $\phantom{5}$\phantom{0.00} &                   &               \\
       & \textbf{R} &  21.94 $\pm$ $\phantom{5}$0.71 &  12.73 $\pm$ 0.74 &  24.18 $\pm$ $\phantom{5}$0.65 &  30.67 $\pm$ $\phantom{5}$0.60 &  40.61 $\pm$ $\phantom{5}$0.74 &  $\phantom{5}$$\phantom{5}$170.56 $\pm$ 17.18 &  $\phantom{5}$2.31 $\pm$ 0.23 \\
\textbf{ConvE} & \textbf{pub (ND)} &     65.70 \phantom{$\pm$} $\phantom{5}$\phantom{0.00} &     55.80 \phantom{$\pm$} \phantom{0.00} &     72.30 \phantom{$\pm$} $\phantom{5}$\phantom{0.00} &                &     83.10 \phantom{$\pm$} $\phantom{5}$\phantom{0.00} &     $\phantom{5}$$\phantom{5}$$\phantom{5}$51.00 \phantom{$\pm$} $\phantom{5}$\phantom{0.00} &               \\
       & \textbf{R} &  75.45 $\pm$ $\phantom{5}$0.17 &  68.26 $\pm$ 0.27 &  80.47 $\pm$ $\phantom{5}$0.08 &  83.94 $\pm$ $\phantom{5}$0.01 &  87.68 $\pm$ $\phantom{5}$0.04 &  $\phantom{5}$$\phantom{5}$$\phantom{5}$43.97 $\pm$ $\phantom{5}$0.60 &  $\phantom{5}$0.60 $\pm$ 0.01 \\
\textbf{\textcolor{RedOrange}{DistMult}} & \textbf{pub (U)} &     35.00 \phantom{$\pm$} $\phantom{5}$\phantom{0.00} &               &                &                &     57.70 \phantom{$\pm$} $\phantom{5}$\phantom{0.00} &                   &               \\
       & \textbf{R} &  28.47 $\pm$ $\phantom{5}$0.23 &  18.59 $\pm$ 0.19 &  31.77 $\pm$ $\phantom{5}$0.29 &  38.24 $\pm$ $\phantom{5}$0.38 &  47.81 $\pm$ $\phantom{5}$0.36 &  $\phantom{5}$$\phantom{5}$127.16 $\pm$ $\phantom{5}$0.85 &  $\phantom{5}$1.72 $\pm$ 0.01 \\
\textbf{\textcolor{RedOrange}{HolE}} & \textbf{pub (ND)} &     52.40 \phantom{$\pm$} $\phantom{5}$\phantom{0.00} &     40.20 \phantom{$\pm$} \phantom{0.00} &     61.30 \phantom{$\pm$} $\phantom{5}$\phantom{0.00} &                &     73.90 \phantom{$\pm$} $\phantom{5}$\phantom{0.00} &                   &               \\
       & \textbf{R} &  39.72 $\pm$ $\phantom{5}$0.32 &  27.15 $\pm$ 0.33 &  46.13 $\pm$ $\phantom{5}$0.40 &  54.05 $\pm$ $\phantom{5}$0.36 &  64.02 $\pm$ $\phantom{5}$0.27 &  $\phantom{5}$$\phantom{5}$186.22 $\pm$ $\phantom{5}$6.21 &  $\phantom{5}$2.52 $\pm$ 0.08 \\
\textbf{\textcolor{RedOrange}{KG2E}} & \textbf{pub (U)} &                &               &                &                &     71.50 \phantom{$\pm$} $\phantom{5}$\phantom{0.00} &     $\phantom{5}$$\phantom{5}$$\phantom{5}$59.00 \phantom{$\pm$} $\phantom{5}$\phantom{0.00} &               \\
       & \textbf{R} &  $\phantom{5}$0.63 $\pm$ $\phantom{5}$0.08 &  $\phantom{5}$0.15 $\pm$ 0.04 &  $\phantom{5}$0.41 $\pm$ $\phantom{5}$0.11 &  $\phantom{5}$0.66 $\pm$ $\phantom{5}$0.17 &  $\phantom{5}$1.25 $\pm$ $\phantom{5}$0.21 &  $\phantom{5}$5784.42 $\pm$ 22.26 &  78.31 $\pm$ 0.30 \\
\textbf{\textcolor{RedOrange}{QuatE$^{1}$}} & \textbf{pub (O)} &     77.00 \phantom{$\pm$} $\phantom{5}$\phantom{0.00} &     70.00 \phantom{$\pm$} \phantom{0.00} &     82.10 \phantom{$\pm$} $\phantom{5}$\phantom{0.00} &                &     87.80 \phantom{$\pm$} $\phantom{5}$\phantom{0.00} &     $\phantom{5}$$\phantom{5}$$\phantom{5}$41.00 \phantom{$\pm$} $\phantom{5}$\phantom{0.00} &               \\
       & \textbf{R} &  22.19 $\pm$ $\phantom{5}$0.17 &  14.65 $\pm$ 0.16 &  23.76 $\pm$ $\phantom{5}$0.26 &  29.37 $\pm$ $\phantom{5}$0.40 &  37.42 $\pm$ $\phantom{5}$0.39 &  $\phantom{5}$$\phantom{5}$229.99 $\pm$ $\phantom{5}$1.57 &  $\phantom{5}$3.11 $\pm$ 0.02 \\
\textbf{\textcolor{NavyBlue}{RotatE}} & \textbf{pub (ND)} &     79.70 \phantom{$\pm$} $\phantom{5}$\phantom{0.00} &     74.60 \phantom{$\pm$} \phantom{0.00} &     83.00 \phantom{$\pm$} $\phantom{5}$\phantom{0.00} &                &     88.40 \phantom{$\pm$} $\phantom{5}$\phantom{0.00} &     $\phantom{5}$$\phantom{5}$$\phantom{5}$40.00 \phantom{$\pm$} $\phantom{5}$\phantom{0.00} &               \\
       & \textbf{R} &  64.94 $\pm$ $\phantom{5}$0.03 &  53.05 $\pm$ 0.05 &  73.31 $\pm$ $\phantom{5}$0.06 &  78.74 $\pm$ $\phantom{5}$0.06 &  84.85 $\pm$ $\phantom{5}$0.03 &  $\phantom{5}$$\phantom{5}$$\phantom{5}$35.66 $\pm$ $\phantom{5}$0.06 &  $\phantom{5}$0.48 $\pm$ 0.00 \\
\textbf{\textcolor{RedOrange}{SimplE}} & \textbf{pub (O)} &     72.70 \phantom{$\pm$} $\phantom{5}$\phantom{0.00} &     66.00 \phantom{$\pm$} \phantom{0.00} &     77.30 \phantom{$\pm$} $\phantom{5}$\phantom{0.00} &                &     83.80 \phantom{$\pm$} $\phantom{5}$\phantom{0.00} &                   &               \\
       & \textbf{R} &  $\phantom{5}$0.04 $\pm$ $\phantom{5}$0.00 &  $\phantom{5}$0.01 $\pm$ 0.00 &  $\phantom{5}$0.03 $\pm$ $\phantom{5}$0.01 &  $\phantom{5}$0.03 $\pm$ $\phantom{5}$0.01 &  $\phantom{5}$0.05 $\pm$ $\phantom{5}$0.00 &  $\phantom{5}$7386.02 $\pm$ $\phantom{5}$2.11 &  99.99 $\pm$ 0.03 \\
       & \textbf{O} &  23.62 $\pm$ 12.90 &  11.67 $\pm$ 8.68 &  24.65 $\pm$ 16.33 &  34.28 $\pm$ 20.19 &  51.91 $\pm$ 24.57 &  $\phantom{5}$$\phantom{5}$148.27 $\pm$ 89.28 &               \\
       & \textbf{P} &  $\phantom{5}$0.03 $\pm$ $\phantom{5}$0.00 &  $\phantom{5}$0.01 $\pm$ 0.00 &  $\phantom{5}$0.03 $\pm$ $\phantom{5}$0.01 &  $\phantom{5}$0.03 $\pm$ $\phantom{5}$0.01 &  $\phantom{5}$0.05 $\pm$ $\phantom{5}$0.00 &  14623.77 $\pm$ 91.95 &               \\
\textbf{\textcolor{RedOrange}{TransD}} & \textbf{pub (U)} &                &               &                &                &     77.30 \phantom{$\pm$} $\phantom{5}$\phantom{0.00} &     $\phantom{5}$$\phantom{5}$$\phantom{5}$91.00 \phantom{$\pm$} $\phantom{5}$\phantom{0.00} &               \\
       & \textbf{R} &  37.30 $\pm$ $\phantom{5}$0.05 &  24.45 $\pm$ 0.08 &  44.22 $\pm$ $\phantom{5}$0.09 &  51.78 $\pm$ $\phantom{5}$0.09 &  61.31 $\pm$ $\phantom{5}$0.07 &  $\phantom{5}$$\phantom{5}$146.55 $\pm$ $\phantom{5}$3.10 &  $\phantom{5}$1.98 $\pm$ 0.04 \\
\textbf{TransE} & \textbf{pub (U)} &                &               &                &                &     47.10 \phantom{$\pm$} $\phantom{5}$\phantom{0.00} &     $\phantom{5}$$\phantom{5}$125.00 \phantom{$\pm$} $\phantom{5}$\phantom{0.00} &               \\
       & \textbf{R} &  29.11 $\pm$ $\phantom{5}$0.20 &  17.99 $\pm$ 0.27 &  33.53 $\pm$ $\phantom{5}$0.18 &  40.76 $\pm$ $\phantom{5}$0.21 &  50.84 $\pm$ $\phantom{5}$0.28 &  $\phantom{5}$$\phantom{5}$122.01 $\pm$ $\phantom{5}$1.09 &  $\phantom{5}$1.65 $\pm$ 0.01 \\
\textbf{\textcolor{RedOrange}{TransH}} & \textbf{pub (U)} &                &               &                &                &     64.40 \phantom{$\pm$} $\phantom{5}$\phantom{0.00} &     $\phantom{5}$$\phantom{5}$$\phantom{5}$87.00 \phantom{$\pm$} $\phantom{5}$\phantom{0.00} &               \\
       & \textbf{R} &  $\phantom{5}$2.59 $\pm$ $\phantom{5}$0.27 &  $\phantom{5}$1.89 $\pm$ 0.35 &  $\phantom{5}$2.87 $\pm$ $\phantom{5}$0.23 &  $\phantom{5}$3.16 $\pm$ $\phantom{5}$0.11 &  $\phantom{5}$3.46 $\pm$ $\phantom{5}$0.15 &  $\phantom{5}$6318.90 $\pm$ 18.86 &  85.54 $\pm$ 0.26 \\
\textbf{\textcolor{RedOrange}{TransR}} & \textbf{pub (ND)} &                &               &                &                &     68.70 \phantom{$\pm$} $\phantom{5}$\phantom{0.00} &     $\phantom{5}$$\phantom{5}$$\phantom{5}$77.00 \phantom{$\pm$} $\phantom{5}$\phantom{0.00} &               \\
       & \textbf{R} &  $\phantom{5}$1.23 $\pm$ $\phantom{5}$0.04 &  $\phantom{5}$0.38 $\pm$ 0.00 &  $\phantom{5}$1.34 $\pm$ $\phantom{5}$0.10 &  $\phantom{5}$1.93 $\pm$ $\phantom{5}$0.12 &  $\phantom{5}$2.79 $\pm$ $\phantom{5}$0.09 &  $\phantom{5}$6130.41 $\pm$ $\phantom{5}$9.59 &  82.99 $\pm$ 0.13 \\
\textbf{TuckER} & \textbf{pub (ND)} &     79.50 \phantom{$\pm$} $\phantom{5}$\phantom{0.00} &     74.10 \phantom{$\pm$} \phantom{0.00} &     83.30 \phantom{$\pm$} $\phantom{5}$\phantom{0.00} &                &     89.20 \phantom{$\pm$} $\phantom{5}$\phantom{0.00} &                   &               \\
       & \textbf{R} &  79.02 $\pm$ $\phantom{5}$0.12 &  73.10 $\pm$ 0.11 &  83.05 $\pm$ $\phantom{5}$0.13 &  85.93 $\pm$ $\phantom{5}$0.16 &  89.10 $\pm$ $\phantom{5}$0.10 &  $\phantom{5}$$\phantom{5}$$\phantom{5}$40.35 $\pm$ $\phantom{5}$0.83 &  $\phantom{5}$0.55 $\pm$ 0.01 \\
\bottomrule
\label{tab:fb15k_table_without_std}
\end{tabular}

 \end{table}

\begin{table}[H]
    \centering
    \caption{Reproduction of Studies on FB15K-237 where \textbf{pub} refers to published results, \textbf{R} to results based on the realistic ranking, \textbf{O} to results based on the optimistic ranking, and \textbf{P} to results based on the pessimistic ranking. 
    For published results, there are two additional rank types, \textbf{U} for undefined due to missing official implementation and \textbf{ND} for non-deterministic.
    We only show the results of the optimistic and pessimistic ranking in case they differ from the realistic ranking.}
    \label{tab:rps_fb15k237}
\begin{tabular}{llrrrrrrr}
\toprule
       &   &      MRR (\%) &   Hits@1 (\%) &   Hits@3 (\%) &   Hits@5 (\%) &  Hits@10 (\%) &               MR &      AMR (\%) \\
\textbf{model} & {} &               &               &               &               &               &                  &               \\
\midrule
\textbf{\textcolor{NavyBlue}{ConvE}} & \textbf{pub (ND)} &     32.50 \phantom{$\pm$} \phantom{0.00} &     23.70 \phantom{$\pm$} \phantom{0.00} &     35.60 \phantom{$\pm$} \phantom{0.00} &               &     50.10 \phantom{$\pm$} \phantom{0.00} &     $\phantom{5}$244.00 \phantom{$\pm$} $\phantom{5}$\phantom{0.00} &               \\
       & \textbf{R} &  29.69 $\pm$ 0.19 &  21.13 $\pm$ 0.21 &  32.32 $\pm$ 0.19 &  38.57 $\pm$ 0.12 &  47.19 $\pm$ 0.08 &  $\phantom{5}$245.83 $\pm$ $\phantom{5}$4.97 &  $\phantom{5}$3.45 $\pm$ 0.07 \\
\textbf{\textcolor{RedOrange}{ConvKB}} & \textbf{pub (O)} &     39.60 \phantom{$\pm$} \phantom{0.00} &               &               &               &     51.70 \phantom{$\pm$} \phantom{0.00} &     $\phantom{5}$257.00 \phantom{$\pm$} $\phantom{5}$\phantom{0.00} &               \\
       & \textbf{R} &  $\phantom{5}$4.22 $\pm$ 0.18 &  $\phantom{5}$2.75 $\pm$ 0.27 &  $\phantom{5}$3.65 $\pm$ 0.19 &  $\phantom{5}$4.44 $\pm$ 0.19 &  $\phantom{5}$7.18 $\pm$ 0.71 &  4314.45 $\pm$ 27.24 &  60.46 $\pm$ 0.38 \\
\textbf{\textcolor{RedOrange}{MuRE}} & \textbf{pub (R)} &     33.60 \phantom{$\pm$} \phantom{0.00} &     24.50 \phantom{$\pm$} \phantom{0.00} &     37.00 \phantom{$\pm$} \phantom{0.00} &               &     52.10 \phantom{$\pm$} \phantom{0.00} &                  &               \\
       & \textbf{R} &  25.16 $\pm$ 0.20 &  16.12 $\pm$ 0.30 &  27.67 $\pm$ 0.21 &  34.21 $\pm$ 0.32 &  43.78 $\pm$ 0.13 &  $\phantom{5}$190.61 $\pm$ $\phantom{5}$0.58 &  $\phantom{5}$2.67 $\pm$ 0.01 \\
\textbf{\textcolor{RedOrange}{QuatE$^{1}$}} & \textbf{pub (O)} &     31.10 \phantom{$\pm$} \phantom{0.00} &     22.10 \phantom{$\pm$} \phantom{0.00} &     34.20 \phantom{$\pm$} \phantom{0.00} &               &     49.50 \phantom{$\pm$} \phantom{0.00} &     $\phantom{5}$176.00 \phantom{$\pm$} $\phantom{5}$\phantom{0.00} &               \\
       & \textbf{R} &  $\phantom{5}$0.26 $\pm$ 0.02 &  $\phantom{5}$0.18 $\pm$ 0.03 &  $\phantom{5}$0.23 $\pm$ 0.02 &  $\phantom{5}$0.25 $\pm$ 0.02 &  $\phantom{5}$0.30 $\pm$ 0.01 &  7119.76 $\pm$ 36.06 &  99.78 $\pm$ 0.51 \\
\textbf{\textcolor{RedOrange}{RotatE}} & \textbf{pub (ND)} &     33.80 \phantom{$\pm$} \phantom{0.00} &     24.10 \phantom{$\pm$} \phantom{0.00} &     37.50 \phantom{$\pm$} \phantom{0.00} &               &     53.30 \phantom{$\pm$} \phantom{0.00} &     $\phantom{5}$177.00 \phantom{$\pm$} $\phantom{5}$\phantom{0.00} &               \\
       & \textbf{R} &  28.79 $\pm$ 0.07 &  19.74 $\pm$ 0.08 &  31.67 $\pm$ 0.05 &  37.89 $\pm$ 0.07 &  47.13 $\pm$ 0.07 &  $\phantom{5}$176.70 $\pm$ $\phantom{5}$0.48 &  $\phantom{5}$2.48 $\pm$ 0.01 \\
\textbf{TuckER} & \textbf{pub (ND)} &     35.80 \phantom{$\pm$} \phantom{0.00} &     26.60 \phantom{$\pm$} \phantom{0.00} &     39.40 \phantom{$\pm$} \phantom{0.00} &               &     54.40 \phantom{$\pm$} \phantom{0.00} &                  &               \\
       & \textbf{R} &  35.51 $\pm$ 0.08 &  26.20 $\pm$ 0.15 &  39.05 $\pm$ 0.10 &  45.59 $\pm$ 0.12 &  54.11 $\pm$ 0.04 &  $\phantom{5}$152.46 $\pm$ $\phantom{5}$2.32 &  $\phantom{5}$2.14 $\pm$ 0.03 \\
\bottomrule
\end{tabular}

 \end{table}

\begin{table}[H]
    \centering
    \caption{Reproduction of Studies on WN18 where \textbf{pub} refers to published results, \textbf{R} to results based on the realistic ranking, \textbf{O} to results based on the optimistic ranking, and \textbf{P} to results based on the pessimistic ranking.
    For published results, there are two additional rank types, \textbf{U} for undefined due to missing official implementation and \textbf{ND} for non-deterministic.
    We only show the results of the optimistic and pessimistic ranking in case they differ from the realistic ranking.\tnote{a}}
    \label{tab:rps_wn18}
\begin{tabular}{llrrrrrrr}
\toprule
       &   &      MRR (\%) &   Hits@1 (\%) &   Hits@3 (\%) &   Hits@5 (\%) &   Hits@10 (\%) &                 MR &      AMR (\%) \\
\textbf{model} & {} &               &               &               &               &                &                    &               \\
\midrule
\textbf{\textcolor{RedOrange}{ComplEx}} & \textbf{pub (O)} &     94.10 \phantom{$\pm$} \phantom{0.00} &     93.60 \phantom{$\pm$} \phantom{0.00} &     94.50 \phantom{$\pm$} \phantom{0.00} &               &     94.70 \phantom{$\pm$} $\phantom{5}$\phantom{0.00} &                    &               \\
       & \textbf{R} &  18.28 $\pm$ 2.10 &  11.65 $\pm$ 1.32 &  19.04 $\pm$ 2.44 &  23.38 $\pm$ 3.06 &  30.70 $\pm$ $\phantom{5}$3.90 &  $\phantom{5}$$\phantom{5}$442.51 $\pm$ $\phantom{5}$47.32 &  $\phantom{5}$2.16 $\pm$ 0.23 \\
\textbf{ConvE} & \textbf{pub (ND)} &     94.30 \phantom{$\pm$} \phantom{0.00} &     93.50 \phantom{$\pm$} \phantom{0.00} &     94.60 \phantom{$\pm$} \phantom{0.00} &               &     95.60 \phantom{$\pm$} $\phantom{5}$\phantom{0.00} &     $\phantom{5}$$\phantom{5}$374.00 \phantom{$\pm$} $\phantom{5}$$\phantom{5}$\phantom{0.00} &               \\
       & \textbf{R} &  94.23 $\pm$ 0.08 &  93.54 $\pm$ 0.16 &  94.68 $\pm$ 0.04 &  95.03 $\pm$ 0.02 &  95.39 $\pm$ $\phantom{5}$0.09 &  $\phantom{5}$$\phantom{5}$462.53 $\pm$ $\phantom{5}$32.15 &  $\phantom{5}$2.26 $\pm$ 0.16 \\
\textbf{DistMult} & \textbf{pub (U)} &     83.00 \phantom{$\pm$} \phantom{0.00} &               &               &               &     94.20 \phantom{$\pm$} $\phantom{5}$\phantom{0.00} &                    &               \\
       & \textbf{R} &  82.41 $\pm$ 0.24 &  74.74 $\pm$ 0.31 &  89.09 $\pm$ 0.19 &  91.36 $\pm$ 0.22 &  93.44 $\pm$ $\phantom{5}$0.15 &  $\phantom{5}$$\phantom{5}$454.41 $\pm$ $\phantom{5}$43.08 &  $\phantom{5}$2.22 $\pm$ 0.21 \\
\textbf{\textcolor{RedOrange}{HolE}} & \textbf{pub (ND)} &     93.80 \phantom{$\pm$} \phantom{0.00} &     93.00 \phantom{$\pm$} \phantom{0.00} &     94.50 \phantom{$\pm$} \phantom{0.00} &               &     94.90 \phantom{$\pm$} $\phantom{5}$\phantom{0.00} &                    &               \\
       & \textbf{R} &  73.43 $\pm$ 0.40 &  63.22 $\pm$ 0.57 &  81.80 $\pm$ 0.40 &  85.81 $\pm$ 0.20 &  89.30 $\pm$ $\phantom{5}$0.26 &  $\phantom{5}$$\phantom{5}$786.05 $\pm$ $\phantom{5}$33.16 &  $\phantom{5}$3.84 $\pm$ 0.16 \\
\textbf{\textcolor{RedOrange}{KG2E}} & \textbf{pub (U)} &               &               &               &               &     92.80 \phantom{$\pm$} $\phantom{5}$\phantom{0.00} &     $\phantom{5}$$\phantom{5}$331.00 \phantom{$\pm$} $\phantom{5}$$\phantom{5}$\phantom{0.00} &               \\
       & \textbf{R} &  $\phantom{5}$3.73 $\pm$ 0.22 &  $\phantom{5}$1.46 $\pm$ 0.19 &  $\phantom{5}$3.27 $\pm$ 0.26 &  $\phantom{5}$4.77 $\pm$ 0.32 &  $\phantom{5}$7.39 $\pm$ $\phantom{5}$0.33 &  $\phantom{5}$2732.49 $\pm$ $\phantom{5}$57.69 &  13.35 $\pm$ 0.28 \\
       & \textbf{O} &  $\phantom{5}$3.74 $\pm$ 0.22 &  $\phantom{5}$1.46 $\pm$ 0.19 &  $\phantom{5}$3.27 $\pm$ 0.26 &  $\phantom{5}$4.77 $\pm$ 0.32 &  $\phantom{5}$7.39 $\pm$ $\phantom{5}$0.33 &  $\phantom{5}$2732.49 $\pm$ $\phantom{5}$57.69 &               \\
\textbf{\textcolor{RedOrange}{QuatE$^{1}$}} & \textbf{pub (O)} &     94.90 \phantom{$\pm$} \phantom{0.00} &     94.10 \phantom{$\pm$} \phantom{0.00} &     95.40 \phantom{$\pm$} \phantom{0.00} &               &     96.00 \phantom{$\pm$} $\phantom{5}$\phantom{0.00} &     $\phantom{5}$$\phantom{5}$388.00 \phantom{$\pm$} $\phantom{5}$$\phantom{5}$\phantom{0.00} &               \\
       & \textbf{R} &  67.28 $\pm$ 0.70 &  58.38 $\pm$ 0.88 &  73.05 $\pm$ 0.61 &  77.86 $\pm$ 0.53 &  83.25 $\pm$ $\phantom{5}$0.38 &  $\phantom{5}$$\phantom{5}$327.12 $\pm$ $\phantom{5}$12.44 &  $\phantom{5}$1.60 $\pm$ 0.06 \\
\textbf{RotatE} & \textbf{pub (ND)} &     94.90 \phantom{$\pm$} \phantom{0.00} &     94.40 \phantom{$\pm$} \phantom{0.00} &     95.20 \phantom{$\pm$} \phantom{0.00} &               &     95.90 \phantom{$\pm$} $\phantom{5}$\phantom{0.00} &     $\phantom{5}$$\phantom{5}$309.00 \phantom{$\pm$} $\phantom{5}$$\phantom{5}$\phantom{0.00} &               \\
       & \textbf{R} &  93.71 $\pm$ 0.03 &  92.27 $\pm$ 0.03 &  94.87 $\pm$ 0.06 &  95.34 $\pm$ 0.04 &  95.83 $\pm$ $\phantom{5}$0.05 &  $\phantom{5}$$\phantom{5}$270.22 $\pm$ $\phantom{5}$$\phantom{5}$7.24 &  $\phantom{5}$1.32 $\pm$ 0.04 \\
\textbf{\textcolor{RedOrange}{SimplE}} & \textbf{pub (O)} &     94.20 \phantom{$\pm$} \phantom{0.00} &     93.90 \phantom{$\pm$} \phantom{0.00} &     94.40 \phantom{$\pm$} \phantom{0.00} &               &     94.70 \phantom{$\pm$} $\phantom{5}$\phantom{0.00} &                    &               \\
       & \textbf{R} &  $\phantom{5}$0.04 $\pm$ 0.02 &  $\phantom{5}$0.01 $\pm$ 0.01 &  $\phantom{5}$0.03 $\pm$ 0.02 &  $\phantom{5}$0.04 $\pm$ 0.03 &  $\phantom{5}$0.06 $\pm$ $\phantom{5}$0.03 &  20355.98 $\pm$ $\phantom{5}$19.42 &  99.48 $\pm$ 0.09 \\
       & \textbf{O} &  32.95 $\pm$ 8.10 &  28.19 $\pm$ 6.94 &  33.94 $\pm$ 8.84 &  37.28 $\pm$ 9.69 &  42.40 $\pm$ 10.53 &  $\phantom{5}$$\phantom{5}$469.49 $\pm$ 161.36 &               \\
       & \textbf{P} &  $\phantom{5}$0.03 $\pm$ 0.01 &  $\phantom{5}$0.01 $\pm$ 0.01 &  $\phantom{5}$0.03 $\pm$ 0.02 &  $\phantom{5}$0.04 $\pm$ 0.03 &  $\phantom{5}$0.06 $\pm$ $\phantom{5}$0.03 &  40242.47 $\pm$ 195.66 &               \\
\textbf{\textcolor{NavyBlue}{TransD}} & \textbf{pub (U)} &               &               &               &               &     92.20 \phantom{$\pm$} $\phantom{5}$\phantom{0.00} &     $\phantom{5}$$\phantom{5}$212.00 \phantom{$\pm$} $\phantom{5}$$\phantom{5}$\phantom{0.00} &               \\
       & \textbf{R} &  37.33 $\pm$ 0.52 &  $\phantom{5}$4.31 $\pm$ 0.42 &  67.90 $\pm$ 0.93 &  81.01 $\pm$ 0.30 &  87.80 $\pm$ $\phantom{5}$0.33 &  $\phantom{5}$$\phantom{5}$460.00 $\pm$ $\phantom{5}$$\phantom{5}$7.40 &  $\phantom{5}$2.25 $\pm$ 0.04 \\
\textbf{\textcolor{RedOrange}{TransE}} & \textbf{pub (U)} &               &               &               &               &     89.20 \phantom{$\pm$} $\phantom{5}$\phantom{0.00} &     $\phantom{5}$$\phantom{5}$251.00 \phantom{$\pm$} $\phantom{5}$$\phantom{5}$\phantom{0.00} &               \\
       & \textbf{R} &  37.04 $\pm$ 1.37 &  $\phantom{5}$9.29 $\pm$ 1.83 &  60.28 $\pm$ 1.25 &  72.02 $\pm$ 0.75 &  81.51 $\pm$ $\phantom{5}$0.52 &  $\phantom{5}$$\phantom{5}$489.84 $\pm$ $\phantom{5}$42.13 &  $\phantom{5}$2.39 $\pm$ 0.21 \\
\textbf{\textcolor{RedOrange}{TransH}} & \textbf{pub (U)} &               &               &               &               &     82.30 \phantom{$\pm$} $\phantom{5}$\phantom{0.00} &     $\phantom{5}$$\phantom{5}$388.00 \phantom{$\pm$} $\phantom{5}$$\phantom{5}$\phantom{0.00} &               \\
       & \textbf{R} &  $\phantom{5}$0.17 $\pm$ 0.17 &  $\phantom{5}$0.08 $\pm$ 0.12 &  $\phantom{5}$0.17 $\pm$ 0.20 &  $\phantom{5}$0.21 $\pm$ 0.24 &  $\phantom{5}$0.31 $\pm$ $\phantom{5}$0.29 &  19551.68 $\pm$ 166.54 &  95.55 $\pm$ 0.81 \\
\textbf{\textcolor{RedOrange}{TransR}} & \textbf{pub (ND)} &               &               &               &               &     92.00 \phantom{$\pm$} $\phantom{5}$\phantom{0.00} &     $\phantom{5}$$\phantom{5}$225.00 \phantom{$\pm$} $\phantom{5}$$\phantom{5}$\phantom{0.00} &               \\
       & \textbf{R} &  $\phantom{5}$0.24 $\pm$ 0.03 &  $\phantom{5}$0.00 $\pm$ 0.01 &  $\phantom{5}$0.22 $\pm$ 0.05 &  $\phantom{5}$0.38 $\pm$ 0.05 &  $\phantom{5}$0.63 $\pm$ $\phantom{5}$0.07 &  18882.20 $\pm$ 240.51 &  92.27 $\pm$ 1.18 \\
\textbf{TuckER} & \textbf{pub (ND)} &     95.30 \phantom{$\pm$} \phantom{0.00} &     94.90 \phantom{$\pm$} \phantom{0.00} &     95.50 \phantom{$\pm$} \phantom{0.00} &               &     95.80 \phantom{$\pm$} $\phantom{5}$\phantom{0.00} &                    &               \\
       & \textbf{R} &  94.89 $\pm$ 0.05 &  94.52 $\pm$ 0.05 &  95.17 $\pm$ 0.07 &  95.30 $\pm$ 0.07 &  95.50 $\pm$ $\phantom{5}$0.06 &  $\phantom{5}$$\phantom{5}$532.05 $\pm$ $\phantom{5}$45.91 &  $\phantom{5}$2.60 $\pm$ 0.22 \\
\bottomrule
\end{tabular}

 \end{table}

\begin{table}[H]
    \centering
    \caption{Reproduction of Studies on WN18RR where \textbf{pub} refers to published results, \textbf{R} to results based on the realistic ranking, \textbf{O} to results based on the optimistic ranking, and \textbf{P} to results based on the pessimistic ranking.
    For published results, there are two additional rank types, \textbf{U} for undefined due to missing official implementation and \textbf{ND} for non-deterministic.
    We only show the results of the optimistic and pessimistic ranking in case they differ from the realistic ranking.}
    \label{tab:rps_wn18rr}
    \begin{threeparttable}
\begin{tabular}{llrrrrrrr}
\toprule
       &   &      MRR (\%) &   Hits@1 (\%) &   Hits@3 (\%) &   Hits@5 (\%) &  Hits@10 (\%) &                 MR &       AMR (\%) \\
\textbf{model} & {} &               &               &               &               &               &                    &                \\
\midrule
\textbf{ConvE} & \textbf{pub (ND)} &     43.00 \phantom{$\pm$} \phantom{0.00} &     40.00 \phantom{$\pm$} \phantom{0.00} &     44.00 \phantom{$\pm$} \phantom{0.00} &               &     52.00 \phantom{$\pm$} \phantom{0.00} &     $\phantom{5}$4187.00 \phantom{$\pm$} $\phantom{5}$$\phantom{5}$\phantom{0.00} &                \\
       & \textbf{R} &  45.28 $\pm$ 0.13 &  41.93 $\pm$ 0.19 &  46.64 $\pm$ 0.25 &  49.07 $\pm$ 0.22 &  51.98 $\pm$ 0.24 &  $\phantom{5}$5203.77 $\pm$ 129.07 &  $\phantom{5}$25.67 $\pm$ 0.64 \\
\textbf{\textcolor{RedOrange}{ConvKB}} & \textbf{pub (O)} &     24.80 \phantom{$\pm$} \phantom{0.00} &               &               &               &     52.50 \phantom{$\pm$} \phantom{0.00} &     $\phantom{5}$2554.00 \phantom{$\pm$} $\phantom{5}$$\phantom{5}$\phantom{0.00} &                \\
       & \textbf{R} &  $\phantom{5}$0.34 $\pm$ 0.05 &  $\phantom{5}$0.11 $\pm$ 0.04 &  $\phantom{5}$0.27 $\pm$ 0.02 &  $\phantom{5}$0.43 $\pm$ 0.06 &  $\phantom{5}$0.63 $\pm$ 0.08 &  13905.99 $\pm$ 962.71 &  $\phantom{5}$68.60 $\pm$ 4.75 \\
\textbf{\textcolor{RedOrange}{QuatE$^{1}$}} & \textbf{pub (O)} &     48.10 \phantom{$\pm$} \phantom{0.00} &     43.60 \phantom{$\pm$} \phantom{0.00} &     50.00 \phantom{$\pm$} \phantom{0.00} &               &     56.40 \phantom{$\pm$} \phantom{0.00} &     $\phantom{5}$3472.00 \phantom{$\pm$} $\phantom{5}$$\phantom{5}$\phantom{0.00} &                \\
       & \textbf{R} &  $\phantom{5}$0.58 $\pm$ 0.05 &  $\phantom{5}$0.38 $\pm$ 0.06 &  $\phantom{5}$0.56 $\pm$ 0.08 &  $\phantom{5}$0.66 $\pm$ 0.06 &  $\phantom{5}$0.88 $\pm$ 0.09 &  20404.47 $\pm$ 196.81 &  100.65 $\pm$ 0.97 \\
\textbf{RotatE} & \textbf{pub (ND)} &     47.60 \phantom{$\pm$} \phantom{0.00} &     42.80 \phantom{$\pm$} \phantom{0.00} &     49.20 \phantom{$\pm$} \phantom{0.00} &               &     57.10 \phantom{$\pm$} \phantom{0.00} &     $\phantom{5}$3340.00 \phantom{$\pm$} $\phantom{5}$$\phantom{5}$\phantom{0.00} &                \\
       & \textbf{R} &  49.39 $\pm$ 0.06 &  45.49 $\pm$ 0.12 &  51.03 $\pm$ 0.10 &  53.36 $\pm$ 0.15 &  57.05 $\pm$ 0.14 &  $\phantom{5}$4046.79 $\pm$ $\phantom{5}$89.15 &  $\phantom{5}$19.96 $\pm$ 0.44 \\
\textbf{TuckER} & \textbf{pub (ND)} &     47.00 \phantom{$\pm$} \phantom{0.00} &     44.30 \phantom{$\pm$} \phantom{0.00} &     48.20 \phantom{$\pm$} \phantom{0.00} &               &     52.60 \phantom{$\pm$} \phantom{0.00} &                    &                \\
       & \textbf{R} &  47.62 $\pm$ 0.58 &  44.91 $\pm$ 0.62 &  48.81 $\pm$ 0.59 &  50.40 $\pm$ 0.58 &  52.80 $\pm$ 0.45 &  $\phantom{5}$5646.84 $\pm$ 146.30 &  $\phantom{5}$27.85 $\pm$ 0.72 \\
\bottomrule
\label{tab:wn18rr_table_without_std}
\end{tabular}

     \begin{tablenotes}
        \item[a] For MuRE, we obtained non-finite loss values while training on WN18RR with the setting defined in~\cite{DBLP:conf/nips/BalazevicAH19}. This might be explained by the fact that the specified learning rate of 50 is comparably large.
        In our benchmarking study, we show that we can outperform the published results with a different setting (Section~\ref{benchmarking_results_wn18rr}).
    \end{tablenotes}
\end{threeparttable}
\end{table}

\begin{table}[H]
    \centering
    \caption{%
    Model sizes in bytes for the best reported configurations studied for the the reproducibility study.%
    }
    \label{tab:repro_size}
\begin{tabular}{lrrrr}
\toprule
Dataset &     FB15K &  FB15K-237 &      WN18 &    WN18RR \\
Model    &           &           &           &           \\
\midrule
ComplEx  &   26.1 MB &         - &   49.2 MB &         - \\
ConvE    &   22.5 MB &   20.3 MB &   41.2 MB &   40.9 MB \\
ConvKB   &         - &    5.9 MB &         - &    8.2 MB \\
DistMult &    6.5 MB &         - &   16.4 MB &         - \\
HolE     &    9.8 MB &         - &   24.6 MB &         - \\
KG2E     &    6.5 MB &         - &   16.4 MB &         - \\
RotatE   &  130.4 MB &  117.9 MB &  163.8 MB &  162.3 MB \\
SimplE   &   26.1 MB &         - &   65.5 MB &         - \\
TransD   &    6.5 MB &         - &   16.4 MB &         - \\
TransE   &    3.3 MB &         - &    3.3 MB &         - \\
TransH   &    7.1 MB &         - &    8.2 MB &         - \\
TransR   &   16.7 MB &         - &    8.4 MB &         - \\
TuckER   &   46.1 MB &         - &   37.6 MB &         - \\
\bottomrule
\end{tabular}
\end{table} \clearpage

\section*{Additional Results From Benchmarking Study}
\begin{table}[H]
\centering
\caption{Pareto-optimal models for FB15k237 regarding Model Bytes and Hits@10}
\label{tab:skyline_fb15k237_model_bytes}
\begin{tabular}{llllrr}
\toprule
   Model & Loss & Training Approach & Inverse Relations & Model Bytes &  Hits@10 (\%) \\
\midrule
  TuckER & BCEL &              LCWA &               yes &     8.0 MiB &       52.857 \\
DistMult &  CEL &              LCWA &               yes &     3.7 MiB &       47.387 \\
  TransE &  SPL &              LCWA &                no &     3.6 MiB &       45.318 \\
      UM &  MRL &             sLCWA &                no &     3.5 MiB &        3.432 \\
      UM &  MRL &             sLCWA &               yes &     3.5 MiB &        3.305 \\
\bottomrule
\end{tabular}
\end{table}

\begin{table}[H]
\centering
\caption{Pareto-optimal models for Kinships regarding Model Bytes and Hits@10}
\label{tab:skyline_kinships_model_bytes}
\begin{tabular}{llllrr}
\toprule
 Model & Loss & Training Approach & Inverse Relations & Model Bytes &  Hits@10 (\%) \\
\midrule
TuckER &  SPL &              LCWA &               yes &     1.0 MiB &       98.603 \\
RotatE &  MRL &             sLCWA &               yes &   154.0 KiB &       98.557 \\
RotatE &  MRL &             sLCWA &                no &   129.0 KiB &       98.324 \\
SimplE & BCEL &              LCWA &               yes &    77.0 KiB &       97.765 \\
 ProjE &  SPL &             sLCWA &               yes &    39.3 KiB &       96.648 \\
 ProjE &  SPL &             sLCWA &                no &    33.0 KiB &       94.600 \\
  HolE &  CEL &              LCWA &                no &    32.2 KiB &       88.873 \\
    UM &  SPL &              LCWA &               yes &    26.0 KiB &       11.313 \\
    UM &  SPL &             sLCWA &               yes &    26.0 KiB &        6.844 \\
\bottomrule
\end{tabular}
\end{table}

\begin{table}[H]
\centering
\caption{Pareto-optimal models for WN18RR regarding Model Bytes and Hits@10}
\label{tab:skyline_wn18rr_model_bytes}
\begin{tabular}{llllrr}
\toprule
 Model & Loss & Training Approach & Inverse Relations & Model Bytes &  Hits@10 (\%) \\
\midrule
RotatE & BCEL &              LCWA &               yes &    79.3 MiB &       60.089 \\
RotatE &  SPL &              LCWA &               yes &    19.8 MiB &       58.328 \\
TuckER &  CEL &              LCWA &               yes &    11.9 MiB &       56.088 \\
  MuRE &  SPL &              LCWA &                no &    10.2 MiB &       55.489 \\
TransH &  MRL &             sLCWA &                no &     9.9 MiB &       48.170 \\
    UM &  SPL &              LCWA &               yes &     9.9 MiB &       44.682 \\
    UM &  SPL &             sLCWA &               yes &     9.9 MiB &       39.022 \\
\bottomrule
\end{tabular}
\end{table}

\begin{table}[H]
\centering
\caption{Pareto-optimal models for YAGO310 regarding Model Bytes and Hits@10}
\label{tab:skyline_yago310_model_bytes}
\begin{tabular}{llllrr}
\toprule
   Model &  Loss & Training Approach & Inverse Relations & Model Bytes &  Hits@10 (\%) \\
\midrule
    MuRE &   SPL &             sLCWA &               yes &    61.1 MiB &       66.851 \\
  ConvKB & NSSAL &             sLCWA &                no &    30.1 MiB &       52.921 \\
DistMult &   SPL &             sLCWA &               yes &    30.1 MiB &       50.562 \\
  TransE &  BCEL &             sLCWA &                no &    30.1 MiB &       14.663 \\
\bottomrule
\end{tabular}
\end{table}

\begin{table}
\centering
\caption{Best configuration for each model in FB15k237}
\label{best_models_fb15k237}
\begin{tabular}{llllr}
\toprule
   Model &  Loss & Training Approach & Inverse Relations &  Hits@10 (\%) \\
\midrule
 ComplEx &   CEL &              LCWA &              True &       44.838 \\
   ConvE &  BCEL &              LCWA &              True &       49.212 \\
  ConvKB &   SPL &             sLCWA &             False &       32.261 \\
DistMult &   CEL &              LCWA &              True &       47.387 \\
   ERMLP &  BCEL &              LCWA &              True &       45.100 \\
    HolE &   CEL &              LCWA &              True &       42.225 \\
    KG2E &   SPL &              LCWA &              True &       45.501 \\
    MuRE &  BCEL &              LCWA &              True &       47.199 \\
     NTN &   SPL &             sLCWA &             False &       20.342 \\
   ProjE &  BCEL &              LCWA &              True &       41.616 \\
   QuatE &   CEL &              LCWA &              True &       46.166 \\
  RESCAL &   CEL &              LCWA &              True &       46.460 \\
  RotatE & NSSAL &             sLCWA &             False &       49.750 \\
      SE & NSSAL &             sLCWA &              True &       39.427 \\
  SimplE &   CEL &              LCWA &              True &       40.307 \\
  TransD &   MRL &             sLCWA &              True &       41.856 \\
  TransE &   MRL &             sLCWA &             False &       46.423 \\
  TransH &   MRL &             sLCWA &             False &       35.295 \\
  TransR &   CEL &              LCWA &              True &       39.187 \\
  TuckER &  BCEL &              LCWA &              True &       52.857 \\
      UM &   CEL &              LCWA &             False &        8.024 \\
\bottomrule
\end{tabular}
\end{table}

\begin{table}
\centering
\caption{Best configuration for each model in Kinships}
\label{best_models_kinships}
\begin{tabular}{llllr}
\toprule
   Model &  Loss & Training Approach & Inverse Relations &  Hits@10 (\%) \\
\midrule
 ComplEx &   CEL &              LCWA &              True &       98.371 \\
   ConvE & NSSAL &             sLCWA &              True &       98.557 \\
  ConvKB & NSSAL &             sLCWA &              True &       97.067 \\
DistMult &   CEL &              LCWA &              True &       93.529 \\
   ERMLP &   SPL &             sLCWA &              True &       97.486 \\
    HolE &   CEL &              LCWA &              True &       93.715 \\
    KG2E &   MRL &             sLCWA &              True &       91.853 \\
    MuRE &   SPL &              LCWA &              True &       95.019 \\
     NTN &  BCEL &             sLCWA &              True &       93.622 \\
   ProjE &   SPL &             sLCWA &              True &       96.648 \\
   QuatE &   CEL &              LCWA &              True &       98.184 \\
  RESCAL &   SPL &             sLCWA &              True &       97.719 \\
  RotatE & NSSAL &             sLCWA &             False &       98.557 \\
      SE & NSSAL &             sLCWA &              True &       98.324 \\
  SimplE &  BCEL &             sLCWA &             False &       98.277 \\
  TransD &   CEL &              LCWA &              True &       45.205 \\
  TransE &   CEL &              LCWA &              True &       92.877 \\
  TransH &   CEL &              LCWA &              True &       52.048 \\
  TransR &   MRL &             sLCWA &             False &       73.324 \\
  TuckER &   SPL &              LCWA &              True &       98.603 \\
      UM &   SPL &              LCWA &              True &       11.313 \\
\bottomrule
\end{tabular}
\end{table}

\begin{table}
\centering
\caption{Best configuration for each model in WN18RR}
\label{best_models_wn18rr}
\begin{tabular}{llllr}
\toprule
   Model &  Loss & Training Approach & Inverse Relations &  Hits@10 (\%) \\
\midrule
 ComplEx &   CEL &              LCWA &             False &       53.745 \\
   ConvE &   CEL &              LCWA &              True &       56.327 \\
  ConvKB & NSSAL &             sLCWA &              True &       42.083 \\
DistMult &   CEL &              LCWA &              True &       52.616 \\
   ERMLP &   SPL &             sLCWA &              True &       47.657 \\
    HolE &   CEL &              LCWA &             False &       50.017 \\
    KG2E &   SPL &              LCWA &             False &       52.035 \\
    MuRE &   SPL &              LCWA &              True &       57.900 \\
     NTN &   MRL &             sLCWA &             False &       31.857 \\
   ProjE &   CEL &              LCWA &              True &       51.727 \\
   QuatE &   CEL &              LCWA &             False &       55.010 \\
  RESCAL &   CEL &              LCWA &             False &       53.916 \\
  RotatE &  BCEL &              LCWA &              True &       60.089 \\
      SE &   SPL &             sLCWA &             False &       45.486 \\
  SimplE &   CEL &              LCWA &              True &       50.889 \\
  TransD &   MRL &             sLCWA &             False &       46.546 \\
  TransE &   SPL &              LCWA &             False &       56.977 \\
  TransH &   MRL &             sLCWA &             False &       48.170 \\
  TransR &   MRL &             sLCWA &             False &       42.510 \\
  TuckER &   CEL &              LCWA &              True &       56.088 \\
      UM &   SPL &              LCWA &             False &       44.887 \\
\bottomrule
\end{tabular}
\end{table}

\begin{table}
\centering
\caption{Best configuration for each model in YAGO310}
\label{best_models_yago310}
\begin{tabular}{llllr}
\toprule
   Model &  Loss & Training Approach & Inverse Relations &  Hits@10 (\%) \\
\midrule
 ComplEx &  BCEL &             sLCWA &              True &       62.575 \\
  ConvKB &   SPL &             sLCWA &              True &       58.149 \\
DistMult &  BCEL &             sLCWA &             False &       55.580 \\
   ERMLP &  BCEL &             sLCWA &              True &       58.531 \\
    HolE &  BCEL &             sLCWA &             False &       60.177 \\
    MuRE &   SPL &             sLCWA &              True &       66.851 \\
   QuatE &   SPL &             sLCWA &              True &       60.709 \\
  RESCAL &   SPL &             sLCWA &              True &       54.045 \\
  RotatE & NSSAL &             sLCWA &              True &       63.077 \\
      SE & NSSAL &             sLCWA &              True &       29.757 \\
  TransD &   MRL &             sLCWA &             False &       35.397 \\
  TransE &   MRL &             sLCWA &              True &       49.217 \\
\bottomrule
\end{tabular}
\end{table}

  \clearpage

\begin{figure}[H]
    \centering
    \includegraphics[width=.8\textwidth]{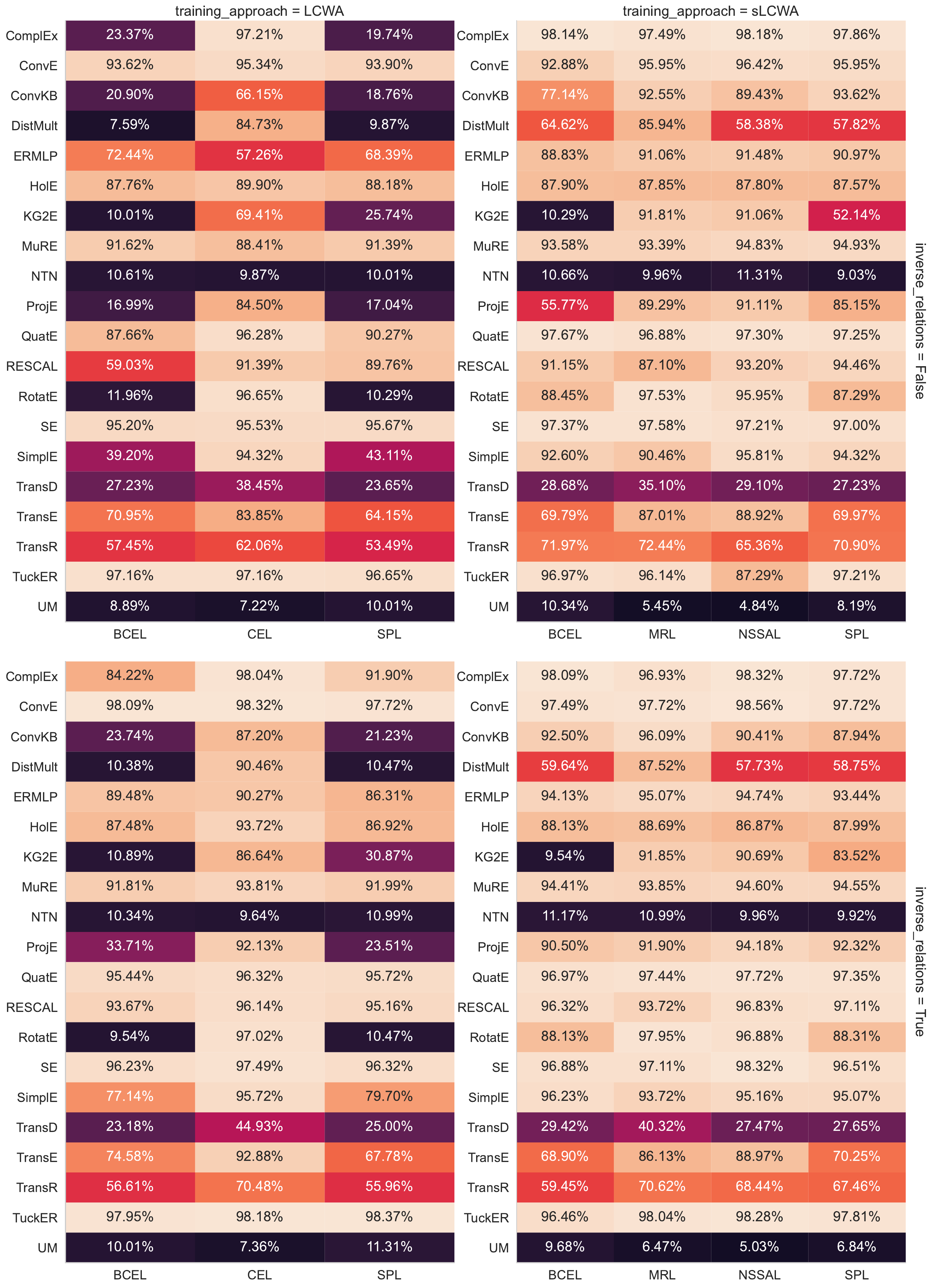}
    \caption{Results for all configurations on Kinships based on Adadelta. \textbf{\Ac{bcel}} refers to the binary cross entropy loss, \textbf{\ac{cel}} to the cross entropy loss, \textbf{\ac{mrl}} to the margin ranking loss, \textbf{\ac{nssal}} refers to the negative sampling self-adversarial loss, \textbf{\ac{spl}} to the softplus loss, \textbf{\ac{lcwa}} to the local closed world assumption training approach and \textbf{\ac{slcwa}} to the stochastic local closed world assumption training approach.}
    \label{fig:all_configs_kinhsips_adadelta}
\end{figure}

\begin{figure}[H]
    \centering
    \includegraphics[width=.8\textwidth]{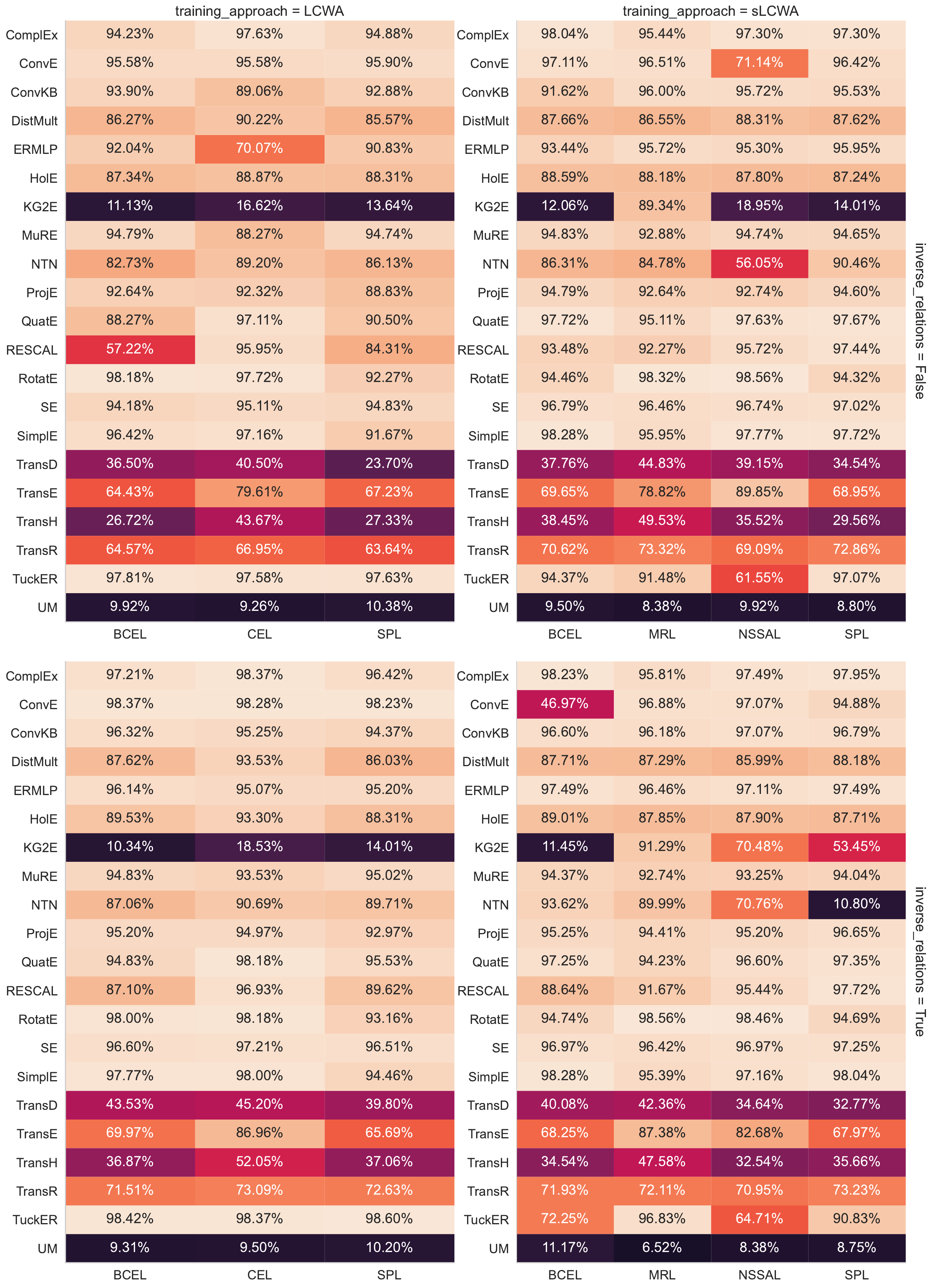}
    \caption{Results for all configurations on Kinships based on Adam. \textbf{\Ac{bcel}} refers to the binary cross entropy loss, \textbf{\ac{cel}} to the cross entropy loss, \textbf{\ac{mrl}} to the margin ranking loss, \textbf{\ac{nssal}} refers to the negative sampling self-adversarial loss, \textbf{\ac{spl}} to the softplus loss, \textbf{\ac{lcwa}} to the local closed world assumption training approach and \textbf{\ac{slcwa}} to the stochastic local closed world assumption training approach.}
    \label{fig:all_configs_kinhsips_adam}
\end{figure}

\begin{figure}[H]
    \centering
    \includegraphics[width=.8\textwidth]{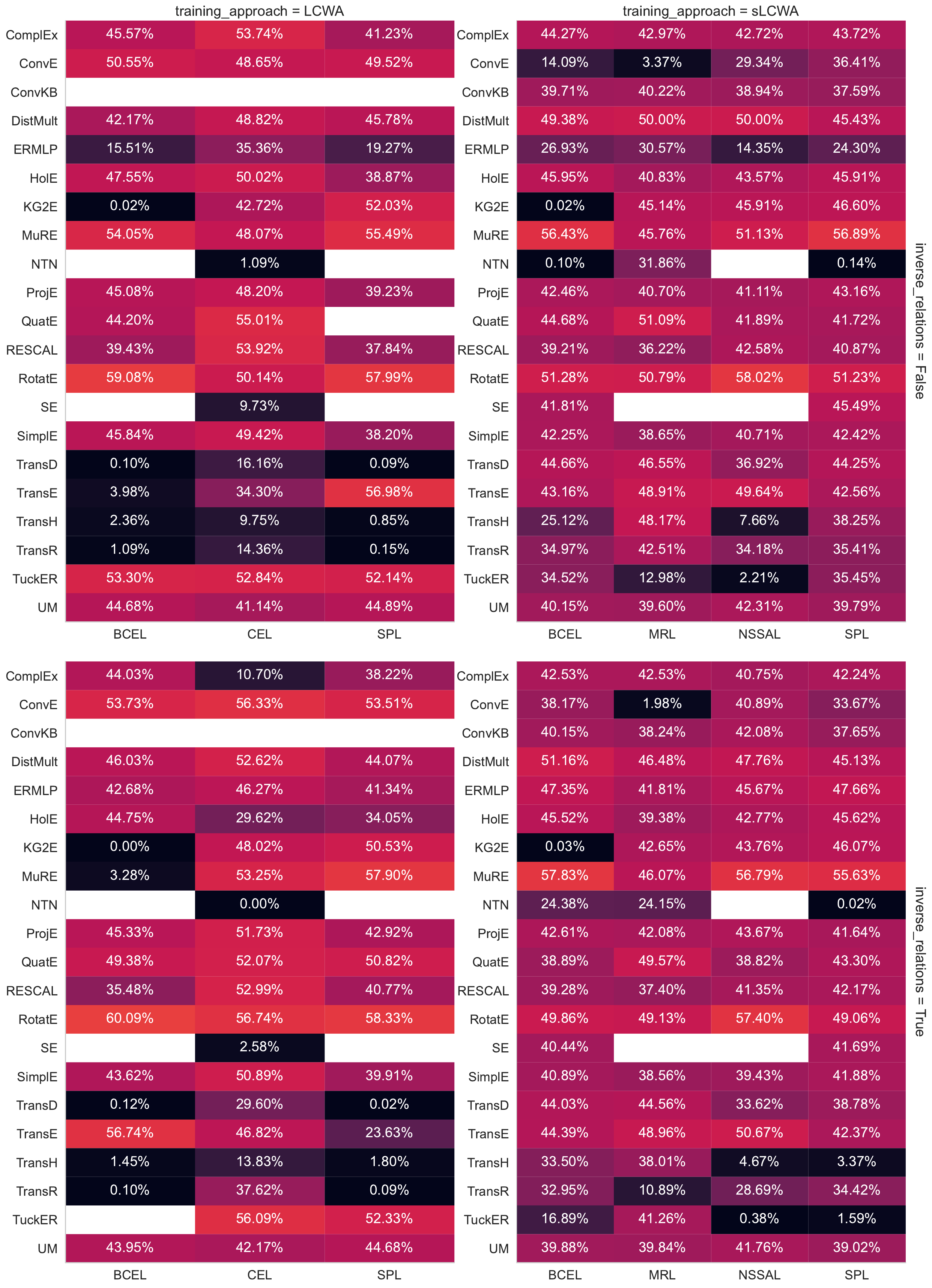}
    \caption{Results for all configurations on WN18RR based on Adam. \textbf{\Ac{bcel}} refers to the binary cross entropy loss, \textbf{\ac{cel}} to the cross entropy loss, \textbf{\ac{mrl}} to the margin ranking loss, \textbf{\ac{nssal}} refers to the negative sampling self-adversarial loss, \textbf{\ac{spl}} to the softplus loss, \textbf{\ac{lcwa}} to the local closed world assumption training approach and \textbf{\ac{slcwa}} to the stochastic local closed world assumption training approach.}
    \label{fig:all_configs_wn18rr_adam}
\end{figure}

\begin{figure}[H]
    \centering
    \includegraphics[width=.8\textwidth]{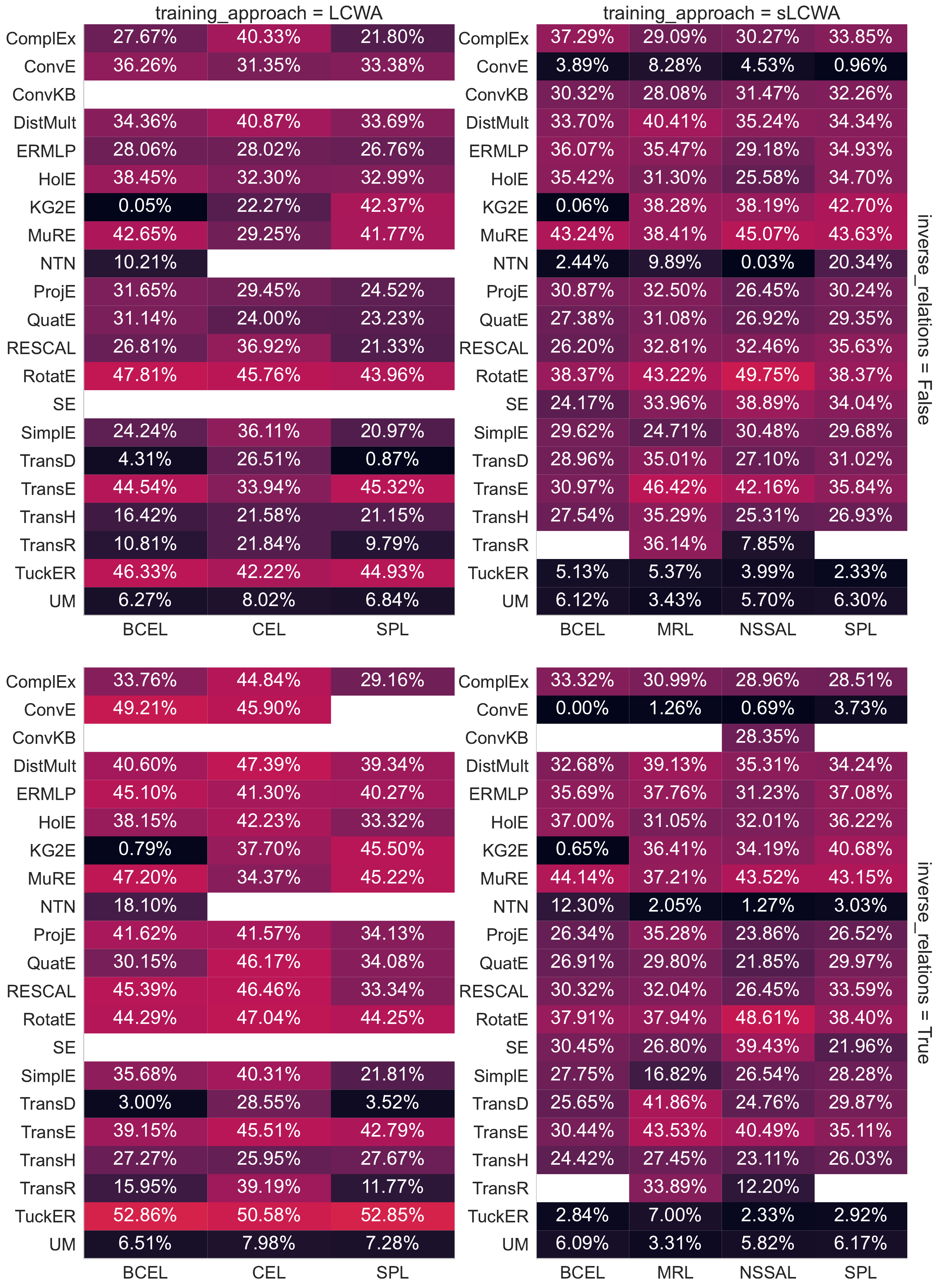}
    \caption{Results for all configurations on FB15K-237 based on Adam. \textbf{\Ac{bcel}} refers to the binary cross entropy loss, \textbf{\ac{cel}} to the cross entropy loss, \textbf{\ac{mrl}} to the margin ranking loss, \textbf{\ac{nssal}} refers to the negative sampling self-adversarial loss, \textbf{\ac{spl}} to the softplus loss, \textbf{\ac{lcwa}} to the local closed world assumption training approach and \textbf{\ac{slcwa}} to the stochastic local closed world assumption training approach.}
    \label{fig:all_configs_fb15k237_adam}
\end{figure}

\begin{figure}[H]
    \centering
    \includegraphics[width=.4\textwidth]{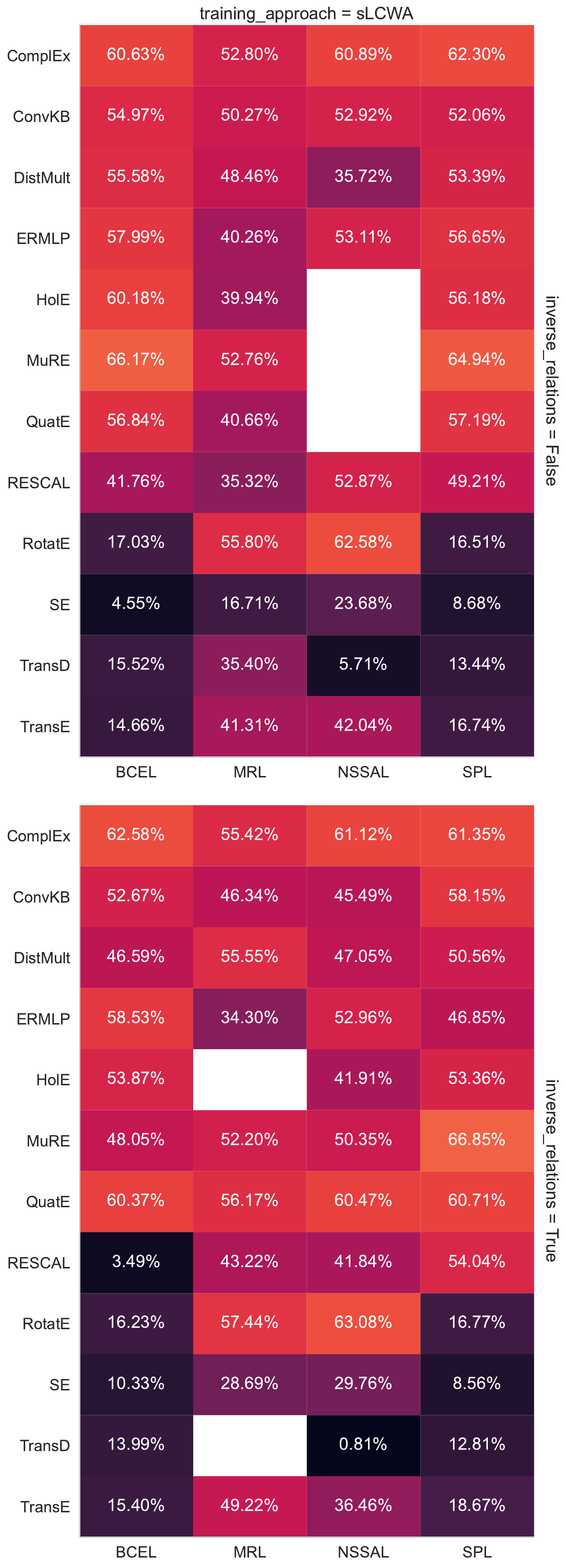}
    \caption{Results for all configurations on YAGO3-10 based on Adam. \textbf{\Ac{bcel}} refers to the binary cross entropy loss, \textbf{\ac{cel}} to the cross entropy loss, \textbf{\ac{mrl}} to the margin ranking loss, \textbf{\ac{nssal}} refers to the negative sampling self-adversarial loss, \textbf{\ac{spl}} to the softplus loss, \textbf{\ac{lcwa}} to the local closed world assumption training approach and \textbf{\ac{slcwa}} to the stochastic local closed world assumption training approach.}
    \label{fig:all_configs_yago310_adam}
\end{figure}

\begin{figure*}[!t]
\centering
\includegraphics[width=0.8\textwidth]{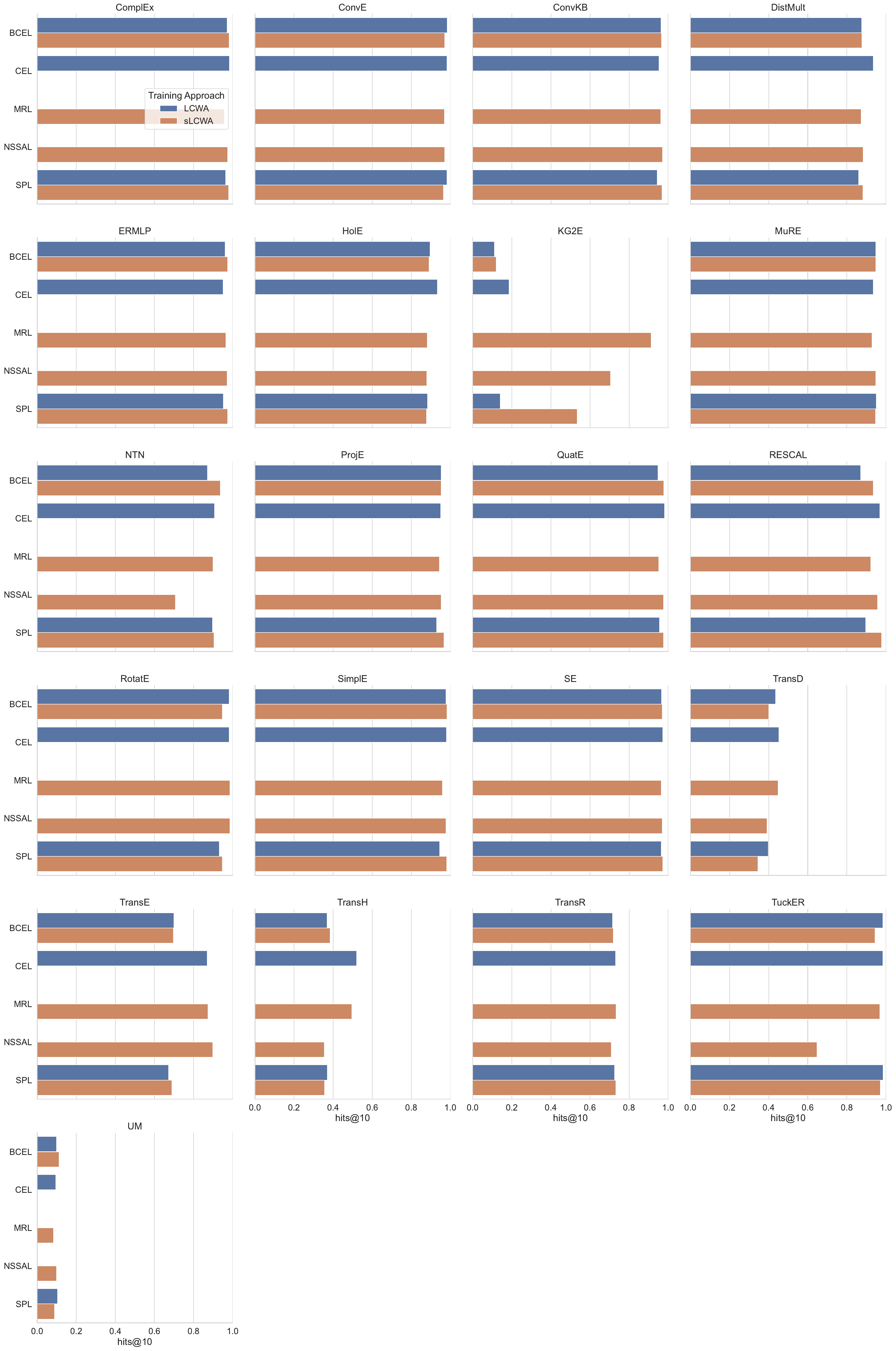}
\caption{Impact of the training approach on the performance for a fixed interaction model and loss function for the Kinships dataset (results represent for each setting the best-performing configuration). \textbf{\ac{bcel}} refers to the binary cross entropy loss, \textbf{\ac{cel}} to the cross entropy loss, \textbf{\ac{mrl}} to the margin ranking loss, \textbf{\ac{nssal}} refers to the negative sampling self-adversarial loss, \textbf{\ac{spl}} to the softplus loss, \textbf{\ac{lcwa}} to the local closed world assumption training approach and \textbf{\ac{slcwa}} to the stochastic local closed world assumption training approach.}
\label{fig:kinships_loss_fct_each_model}
\end{figure*}

\end{document}